\documentclass[letterpaper, 10 pt, conference]{ieeeconf}  

\IEEEoverridecommandlockouts                              

\usepackage{amsfonts}
\usepackage{amsmath}
\usepackage{caption}
\usepackage{subcaption}
\usepackage{graphicx}
\usepackage{algorithm2e}
\usepackage{wrapfig}
\usepackage{lipsum}
\usepackage{multirow}
\usepackage{multicol}
\usepackage{cite}
\usepackage{dblfloatfix}
\usepackage{float}
\usepackage{todonotes}
\usepackage{hyperref}

\SetKwInput{KwData}{Input}
\SetKwInput{KwResult}{Output}

\newtheorem{definition}{Definition}
\newtheorem{theorem}{Theorem}
\newtheorem{proposition}{Proposition}
\DeclareMathOperator*{\argmin}{arg\,min}

\title{\LARGE \bf
Learning Complex Motion Plans using Neural ODEs with Safety and Stability Guarantees 
}

\author{Farhad Nawaz$^{1*}$, Tianyu Li$^{1}$, Nikolai Matni$^{1}$ and Nadia Figueroa$^{1}$
\thanks{$^{1}$Farhad Nawaz, Tianyu Li, Nikolai Matni and Nadia Figueroa are with the GRASP Lab, University of Pennsylvania, PA 19104, USA.}
\thanks{$^{*}$Corresponding author: {\tt farhadn@seas.upenn.edu}}
}

\begin{document}

\maketitle
\thispagestyle{empty}
\pagestyle{empty}

\begin{abstract}
We propose a Dynamical System (DS) approach to learn complex, possibly periodic motion plans from kinesthetic demonstrations using Neural Ordinary Differential Equations (NODE). To ensure reactivity and robustness to disturbances, we propose a novel approach that selects a target point at each time step for the robot to follow, by combining tools from control theory and the target trajectory generated by the learned NODE. 
A correction term to the NODE model is computed online by solving a quadratic program that guarantees stability and safety using control Lyapunov functions and control barrier functions, respectively. Our approach outperforms baseline DS learning techniques on the LASA handwriting dataset and complex periodic trajectories. It is also validated on the Franka Emika robot arm to produce stable motions for wiping and stirring tasks that do not have a single attractor, while being robust to perturbations and safe around humans and obstacles. Project web-page: ~\url{https://sites.google.com/view/lfd-neural-ode/home}.
\end{abstract}
\vspace{-0.15em}

\section{Introduction}
\label{sec:Intro}

We are witnessing an increased use of robotic manipulators for everyday tasks in homes, offices, and factories, all of which involve human interaction. In such cases, robots should be be able to perform not just simple pick-and-place tasks~\cite{pick_place}, but also complex motions such as wiping a surface, stirring a pan, or peeling vegetables~\cite{periodic}. The major challenges in these settings are: (a) learning complex tasks from few observations (b) reactive and compliant motion planning in a dynamic environment and (c) remaining safe and robust to unexpected disturbances.

Learning from Demonstrations~(LfD) is a widely used framework that enables transfer of skills to robots from demonstrations of desired tasks~\cite{SEDS, DMP_related, periodic}. Typically, the observations are robot trajectories obtained through kinesthetic teaching, wherein humans passively guide the robot through the nominal motion. This approach has been shown to avoid the correspondence problem and reduce the simulation to reality gap~\cite{kinesthetic}. In this setting, demonstrations are costly to collect: hence, it is essential to learn motion plans from as few demonstrations as possible, while still being robust, safe, and reactive in a dynamic environment. 

There are multiple approaches for learning from demonstrations. Inverse Reinforcement Learning (IRL) and Behavior Cloning (BC) are popular methodologies to imitate motion~\cite{abbeel2004apprenticeship, IOC, BC} by optimizing an underlying task dependent objective function. IRL and BC require the demonstrator to explore the task space for learning an accurate motion policy that generalizes well: collecting the large amounts of data needed for these approaches is not feasible when the demonstrator is a human. Popular imitation algorithms such as DAgger~\cite{DAGGER} and deep imitation learning~\cite{deep_imit} are also data-hungry (typically requiring hundreds or thousands of demonstrations), and further provide no stability or safety guarantees. Another imitation method based on Gaussian Processes~(GPs) is proposed in~\cite{LfD_GP} that relies on time inputs and way-points to roll out desired trajectories. Such a framework cannot accommodate delays in accomplishing a task, e.g., as caused by a human moving the robot, and also cannot guarantee safety or stability. 

We therefore root our methods in Dynamical System (DS) based motion planning approaches~\cite{billard2022learning, figueroa2018physically, multi_arm, SO3_1} that learn a global vector field instead of discrete way-points, and have shown to generate adaptable motion plans with minimal demonstrations.
Prior work on LfD using DS~\cite{SEDS, figueroa2018physically, figueroa2022locally} have demonstrated success in providing a powerful framework that can learn stable robotic motions while being compliant and reactive to human interactions online. Stable Estimator of Dynamical Systems (SEDS)~\cite{SEDS} is a LfD method that learns globally stable dynamical systems with respect to a goal point using Gaussian mixture models and quadratic Lyapunov functions. An important limitation of SEDS is that it can only model trajectories whose distance to a single target point decreases monotonically in time. Another method based on SEDS is presented in~\cite{figueroa2018physically} via a Linear Parameter Varying (LPV) re-formulation that learns more complex hand-drawn trajectories than SEDS, but is limited to motions that converge to a \textit{single attractor (target)}. 

In contrast, for robots to perform a variety of everyday tasks, we must be able to model complex periodic motions such as wiping a surface or stirring a pan. We propose to learn DS based motion plans that converge to a \textit{target trajectory} which encodes the desired complex periodic motions, rather than to a single target point.  Stability is then guaranteed at task execution with respect to the error dynamics between the robot and the target trajectory, while remaining safe to unforeseen obstacles. To the best of our knowledge, no prior work except for the approach proposed in~\cite{imit_norm} learns stable periodic motions for robotic tasks which uses normalizing flows. However, the approach described in~\cite{imit_norm} requires prior knowledge of whether the demonstrations depict single attractor dynamics or limit cycle dynamics. Our approach relies on no prior knowledge about the demonstrations, and captures the invariant features of complex target trajectories using the rich model class of Neural Ordinary Differential Equations (NODEs). We then guarantee stability (in the error dynamics) and safety at task execution by augmenting the nominal motion plan with corrective inputs computed using Control Lyapunov Functions (CLFs) and Control Barrier Functions (CBFs), respectively~\cite{ames2019control}.

\noindent \textbf{Contributions} In summary, our contributions are given below.
\begin{enumerate}
    \item We employ a NODE based offline learning method that captures the invariant features of complex nonlinear and periodic motions, such as wiping and stirring, using only a few demonstrations.
    \item We propose a modular motion planner which generates a DS-based reactive motion policy by solving a Quadratic Program (QP) online at high frequency (1~KHz) that augments the nominal NODE-based plan with corrective terms based on CLFs and CBFs to guarantee stability and safety, respectively. 
    \item We define a novel look-ahead strategy that chooses a target point at every time step for the robot to follow a target trajectory instead of a single target point.
    \item We show significant performance improvements over existing methods on the LASA handwriting data set \cite{SEDS}, and on periodic 2D and 3D trajectories.  We further validate our approach on complex nonlinear and periodic motions with the Franka Emika robot arm.
\end{enumerate}
\vspace{-5pt}
\section{Problem Formulation}
\label{sec:probl}
\begin{figure}[!tbp]
\centering
        \begin{subfigure}[b]{0.49\linewidth}
         \centering         
         \includegraphics[width=\textwidth]{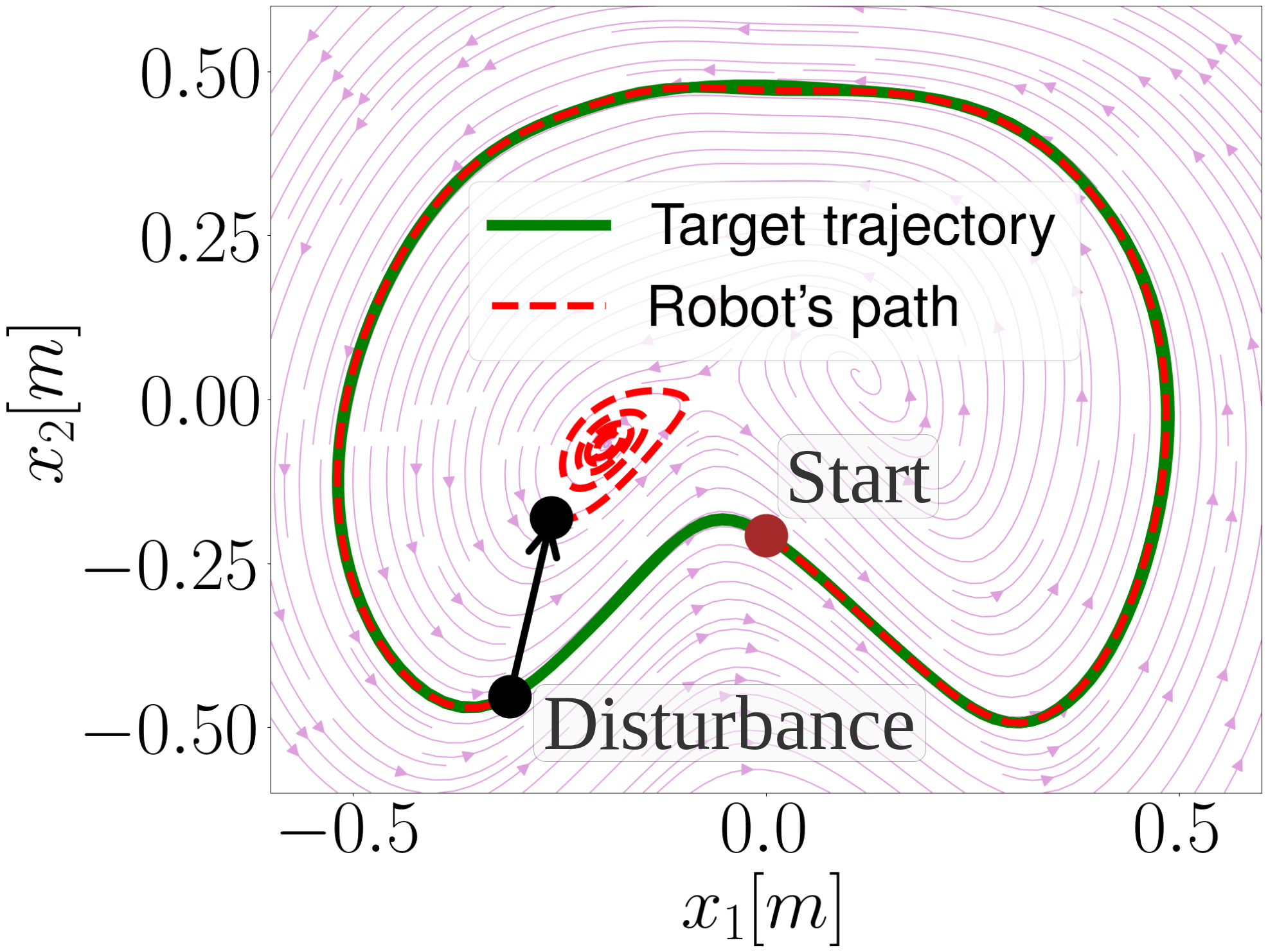}
         \caption{Using standard NODE}
         \label{fig:Spur_att}
        \end{subfigure}
        \begin{subfigure}[b]{0.49\linewidth}         
        \centering        
        \includegraphics[width=\textwidth]{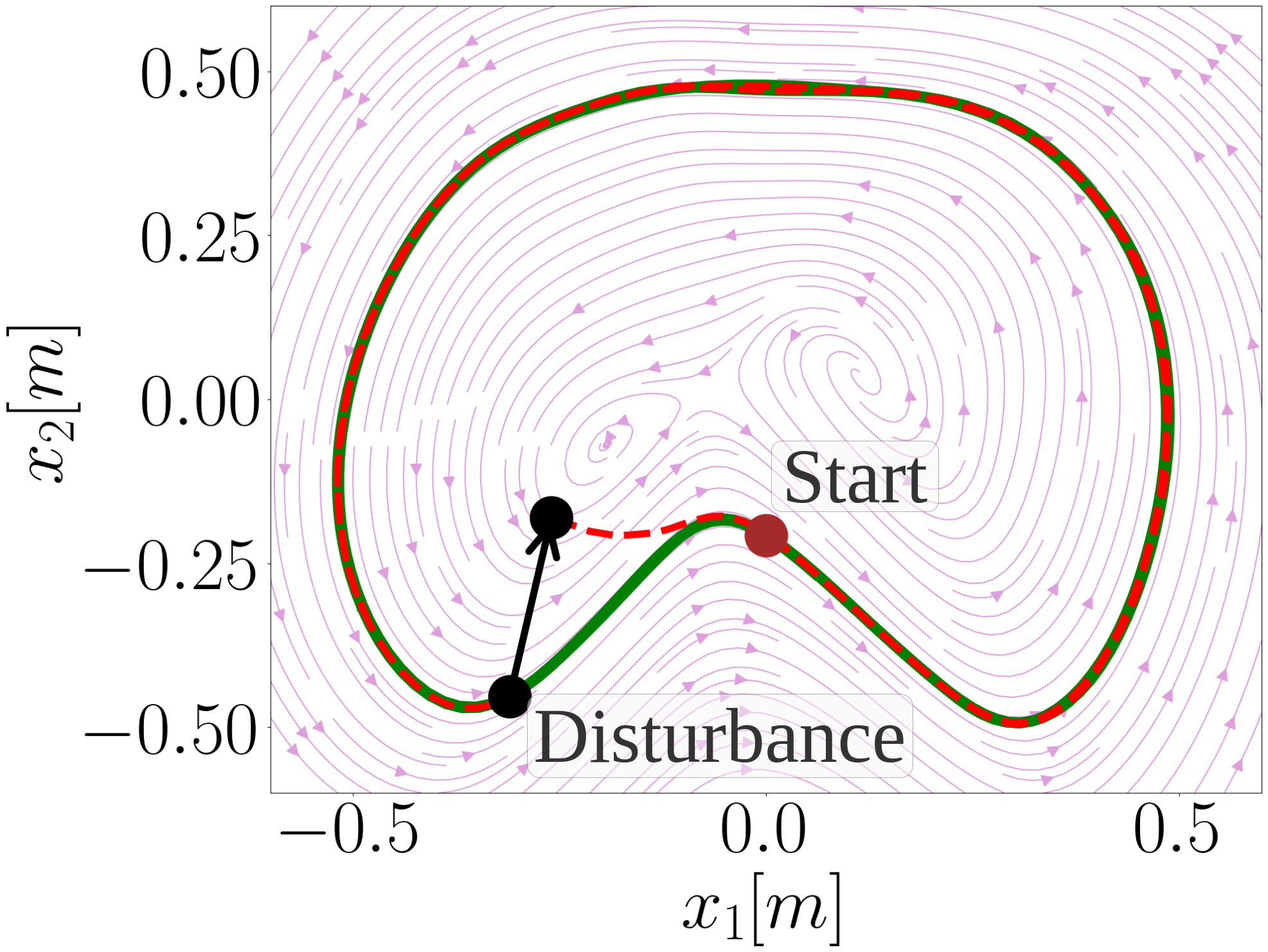}
         \caption{Using CLF-NODE (ours)}
         \label{fig:Spur_att_CLF}
        \end{subfigure}
        \caption{\small Illustrative example of a spurious attractor when the robot's path is guided by a DS-based motion plan in the presence of a disturbance (a) using NODE to encode the motion plan and (b) using the corrected CLF-NODE.}
        \label{fig:Spur_att_all} 
\end{figure}

A DS based motion plan for a robotic manipulator is defined in terms of the state variable $x\in \mathbb{R}^d$, where $x$ is the robot’s end-effector Cartesian state relevant to the task.
The motion plan is formulated as an autonomous DS
\vspace{-3pt}
\begin{equation}
    \dot{x} = f(x),
\label{DS_form}
\vspace{-3pt}
\end{equation}
where $f(\cdot) : \mathbb{R}^d \to \mathbb{R}^d$ is a nonlinear continuous function. The demonstrations from kinesthetic teaching is given by $\mathcal{D} := \{x_i(t_1), x_i(t_2), \ldots, x_i(t_T)\}_{i=1}^{M}$, where, $x_i(t_k)$ is the state of the robot at time $t_k$, for the $i^{th}$ demonstration. We have a total of $M$ demonstrations. The discrete points in each demonstration are sampled at time~$\{t_1, t_2, \ldots, t_T\}$. We assume that the training data trajectories~$\mathcal{D}$ approximate an \textit{unknown nominal target trajectory} $z^*(t)$ that encodes the task of the robot such as wiping, stirring, scooping, etc. Our aim is to design a vector field~$f(x)$ using the demonstrations~$\mathcal{D}$ such that~$x(t)$ follows the target trajectory~$z^*(t)$.  
Previous work~\cite{SEDS, figueroa2018physically} in the DS-based motion planning framework has considered convergence only to a single target. We consider convergence to a trajectory $z^*(t)$ that can represent more complex, e.g., highly-nonlinear and periodic motions as shown in Fig.~\ref{fig:Spur_att_all}. 

Under nominal circumstances, i.e., in the absence of disturbances or obstacles, the target trajectory~$z^*(t)$ should be viewed as the reference for the low level controller to track. However, during deployment, the robot might not always follow the target trajectory because of disturbances, obstacles, time delays, etc. For example, consider the scenario in Fig.~\ref{fig:Spur_att}, where the target trajectory encodes a periodic wiping motion, and the vector field is an unconstrained DS model~\eqref{DS_form} learned from demonstrations~$\mathcal{D}$. As shown in Fig.~\ref{fig:Spur_att}, if the robot is perturbed by a disturbance during deployment to a region where there is no training data, the learned model commands the robot to a spurious attractor. However, the desired behaviour is to continue tracking the target trajectory so that the robot wipes the necessary space as shown in Fig.~\ref{fig:Spur_att_CLF}. Since we focus on stability of the error dynamics, the robot's path converge back to the target trajectory, instead of a single goal point~\cite{SEDS, figueroa2018physically}, which enables the robot to perform more complex and periodic motions. Ensuring robustness to spatial and temporal perturbations is critical for deploying robots in human-centric environments, as disturbances can arise due to obstacles unseen in demonstrations, intentional or adversarial disturbances caused by humans~\cite{billard2022learning,wang2022temporal}, and time delays in the controller~\cite{LfD_GP}. This leads to our formal problem statement: 

\textit{Given a set $\mathcal{D}$,
    design a vector field~$f(x)$ for the dynamical system~\eqref{DS_form}, such that it generates safe and stable motion plans at deployment for scenarios possibly not seen in the demonstrations, while ensuring that the robot's trajectory $x(t)$ converges to the target trajectory $z^*(t)$.}

\section{Proposed Approach}
\label{sec:approach}
We parameterize the vector field~\eqref{DS_form} of the motion plan as
\begin{equation}
\dot{x} = \hat{f}(x) + u(x),
    \label{ref_motion_app}
\end{equation}
where, $\hat{f}(x)$ is used to encode the nominal system behavior, and $u(x)$ is used to enforce safety and disturbance rejection properties. We learn the nominal system~$\hat{f}(x)$ from demonstrations~$\mathcal{D}$, and compute a correction term~$u(x)$ based on control theoretic tools so that the goals of stability and safety are met in a composable, modular, and data efficient way. 
Similar to our objectives, the method proposed in~\cite{figueroa2022locally} generates motion plans that not only converge to a global goal, but also has local stiff behaviours in regions around the target trajectory. Yet, they still lack in representing complex motions that are periodic, have high curvature and partial divergence. We use a NODE model to represent the complex nominal motion and a modular approach similar to~\cite{CLF_learning}, but, we define a CLF with respect to a target trajectory rather than a single goal point that is assumed in~\cite{CLF_learning}. CBFs~\cite{cbf_learning, ames2019control} are widely used in the low level controller to enforce safety of dynamical systems, and we adopt them to generate safe motion plans for robotic manipulation. 
\begin{figure}[!t]
         \centering         \includegraphics[width=0.5\textwidth]{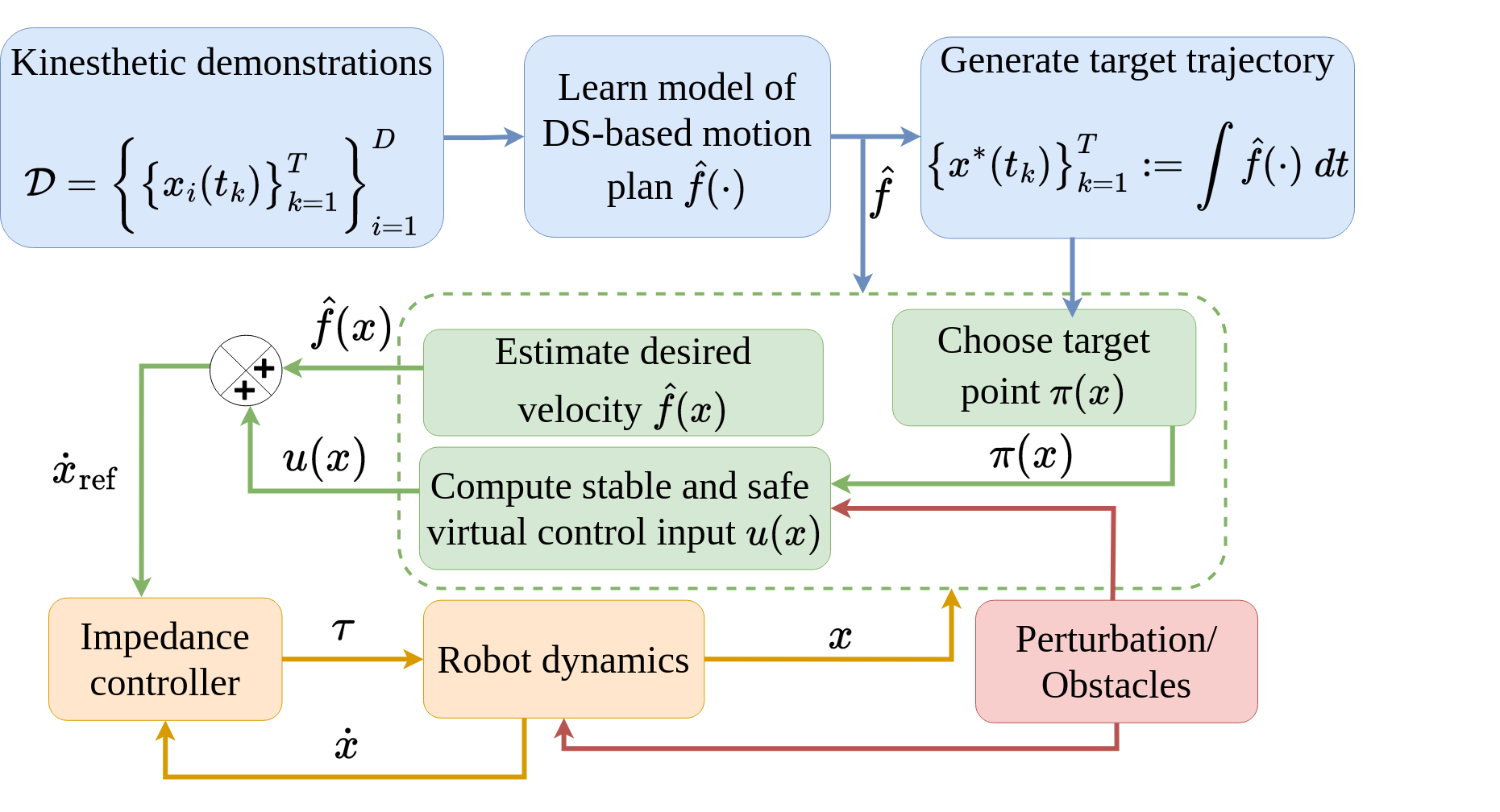}
      \caption{\small The control flow of our proposed pipeline. The state of the robot is $x$, the low level control input are the joint torques $\tau$, and the desired velocity in state space is $\dot{x}_\mathrm{ref}$.} 
        \label{fig:block_diag} 
        \vspace{-1.75em}
\end{figure}

A schematic of the proposed control flow is presented in Fig.~\ref{fig:block_diag}. The blue blocks represent the offline learning component and the green blocks are the online computation modules. We use a neural network parameterized model $\hat{f}$ that we learn from demonstrations~$\mathcal{D}$, but any other model class could be used within our proposed modular framework. Starting from an initial condition, we integrate $\hat{f}(x)$ for the same time span of the task given in demonstrations to generate a target trajectory $x^*(t)$ that approximates the unknown nominal target trajectory~$z^*(t)$. At deployment, given an observation of the current state $x$, our planner chooses the target point~$\pi(x)$ that the robot should follow using the pre-computed target trajectory~$x^*(t)$. The ability to select a target point without relying on time input, unlike prior BC and GP-based methods~\cite{DAGGER, LfD_GP}, enhances reactive motion planning by eliminating time delay issues in the control pipeline. We estimate the nominal desired velocity~$\hat{f}(x)$ using our learned model. However, as illustrated in Fig.~\ref{fig:Spur_att_all}, the generated motion plan from $\hat{f}$ is neither guaranteed to be stable nor safe. Hence, we compute a \textit{virtual control input}~$u(x)$ as an additive correction term that generates the reference motion plan~$\dot{x}$ using~\eqref{ref_motion_app} so that the trajectory generated by $\dot{x}$ converges to the target trajectory even in the presence of disturbances and unsafe regions such as obstacles. We denote the reference velocity for the low level controller as $\dot{x}_{\mathrm{ref}}$, which in general may be different from the real velocity $\dot{x}$ of the robot. The reference velocity $\dot{x}_{\mathrm{ref}}$ is given as input to the impedance controller~\cite{7358081} that computes the low level control input~$\tau$ (joint torques) for the physical robotic system.  We emphasize that the virtual control input~$u(x)$ is different from the low-level control inputs~$\tau$ given in Fig.~\ref{fig:block_diag}, and~$u(x)$ is a component of the motion planning DS~\eqref{ref_motion_app}.
\vspace{-5pt}
\section{Learning Nominal Motion Plans using Neural ODEs}
\label{sec:learning_method}
\vspace{-5pt}
We propose a neural network parameterized function class to learn the dynamics of the motion offline from demonstrations. Although existing work~\cite{figueroa2022locally} learns stable dynamics for motions that converge only to a single target, we aim to learn more complex trajectories that not necessarily converge to a single target. Since neural networks have demonstrated high capability for accurate time-series modeling~\cite{neural_event}, we base our approach on Neural ODEs~\cite{chen2019neural}, which are the continuous limit (in depth) of ResNets~\cite{haber2017stable}. We parameterize our models of nominal target trajectories as:
\begin{equation}
    \frac{d \hat{x}(t)}{dt} = f_{\theta}(\hat{x}(t)),
    \label{nODE_model}
\end{equation}
where $f_{\theta}(\cdot) : \mathbb{R}^d \to \mathbb{R}^d$ is a neural network with parameters $\theta$, and $\hat{x}(t) \in \mathbb{R}^d$ is the state variable predicted by $f_{\theta}(\cdot)$ at time $t$. In the forward pass, given the integration time interval $[a, b]$ and an initial point $\hat{x}(a)$, the model outputs the trajectory $\hat{x}(t)$ for $t \in [a, b]$. The trajectory is obtained by solving the differential equation in~\eqref{nODE_model} using a general-purpose differential equation solver based on fifth-order Runge-Kutta~\cite{rungekutta}. We set $f_{\theta}(\cdot)$ to be a Multi-Layer Perceptron~(MLP), where the inputs and outputs are in $\mathbb{R}^d$. We consider the supervised learning setup with training data~$\mathcal{D}$ and solve the empirical risk minimization problem
\vspace{-5pt}
\begin{equation}
        \min_{\theta} \frac{1}{MT}\sum_{i=1}^M\sum_{k=1}^T\Bigl\|x_{i}(t_k) - \hat{x}_i(t_k)\Bigr\|_2^2,
    \label{nODE_opt}
\end{equation}
to learn the parameters $\theta$,  where the predictions of the state $\hat{x}_{i}(t_k)$ are obtained by integrating~\eqref{nODE_model} with initial condition $\hat{x}_{i}(t_{1}) = x_{i}(t_{1}), \ \forall \ i \in \{1,2,\ldots,M\}$.\footnote{A binomial checkpoint method is used during training for solving~\eqref{nODE_opt} as implemented in~\cite{kidger2022neural}.}
In contrast to previous work~\cite{figueroa2022locally, CLF_learning} which learns a map $\hat{f}(\cdot)$ using labeled data $\{x(t), \dot{x}(t)\}$, we do not assume access to velocity measurements as they are often not easily collected and/or noisy~\cite{only_pos, vel_noisy}. Fitting a map to noisy measurements lead to aggressive trajectories at inference that are not desirable for the low-level controller. From our results presented in Fig.~\ref{fig:Spur_att_all} and Section~\ref{sec:experiments}, we observe that the NODE model generates smooth trajectories utilizing only state variables~$x(t)$ to learn $f_{\theta}$ and not their derivatives $\dot{x}(t)$.  While such a NODE-based vector field will behave reliably near the training data, unanticipated disturbances or obstacles during deployment might deviate the robot to regions of the state-space where the learned vector field is unreliable, as shown in Fig.~\ref{fig:Spur_att}. We present a method that computes a correction term to ensure robust and safe tracking of the learned target trajectory.  

\section{Enforcing Stability and Safety via Virtual Control Inputs}
\label{sec:correction}
\vspace{-2pt}
Safety and stability with respect to obstacles and changes in the environment, as illustrated in Fig.~\ref{fig:Spur_att_all}, are central for tasks in Human-Robot Interaction. Artificial Potential Fields~(APFs)~\cite{obst_compare, APF_slow} have been widely used for obstacle avoidance, but are prone to oscillations and are slow in real-time~\cite{obst_compare}. A DS based modulation strategy is presented in~\cite{mod_obst, bio_obst} to avoid obstacles that are limited to convex shapes. We use the CBF based approach that can handle dynamic and multiple non-convex obstacles without sacrificing real-time performance, and is a generalization of both the modulation strategy~\cite{mod_obst} and APFs~\cite{obst_compare}. Prior work on CLFs~\cite{CLF_learning} for motion generation of robotic manipulators focused only on point-to-point reaching motions, but we integrate them into our learned NODE model for complex periodic motions and present a unified framework that guarantees both stability and safety in real-time. 

\vspace{-3pt}
\subsection{Preliminaries on Safety and Stability}
\label{sec:CBF_def}
\vspace{-2pt}
We begin with a review of control theoretic tools that provide sufficient conditions for stability and safety of nonlinear control affine dynamical systems:
\begin{equation}
\dot{x} = g(x) + h(x)u,
    \label{nonlinear_DS}
\end{equation}
where, $x \in \mathcal{X} \subset \mathbb{R}^d$ and $u \in \mathcal{U} \subset \mathbb{R}^m$ are the set of allowable states and control inputs, respectively. The DS-based motion plan~\eqref{ref_motion_app} is a nonlinear control affine DS with $m = d, \ g(x) = \hat{f}(x)$ and $h(x) = I$.

We first define safety with respect to a safe set $\mathcal{C} \subseteq \mathcal{X}$ for the system~\eqref{nonlinear_DS}. The safe set $\mathcal{C}$ is defined as the super-level set of a function~$B(\cdot) : \mathbb{R}^d \to \mathbb{R}$:
\begin{equation}
\mathcal{C} = \{x \in \mathcal{X} : B(x) \geq 0\}.
    \label{superlevel}
\end{equation}
Our objective is to find a control input~$u$ such that the states $x$ that evolve according to the dynamics~\eqref{nonlinear_DS} always stay inside the safe set $\mathcal{C}$. Such an objective is formalized using \textit{forward invariance} of the safe set $\mathcal{C}$. Forward invariance and CBFs are defined as follows~\cite{ames2019control}. 
\begin{definition}
The safe set $\mathcal{C}$ is forward invariant if for every initial point $x(0) = x_0 \in \mathcal{C}$, the future states $x(t) \in \mathcal{C}$ for all $t \geq 0$.
    \label{def:invariance}
\end{definition}
\begin{definition}
Let $\mathcal{C}$ be the super-level set of 
a continuously differentiable function $B(\cdot): \mathbb{R}^d \to \mathbb{R}$ as given in~\eqref{superlevel}. Then, $B$ is a Control Barrier Function (CBF) for the dynamical system~\eqref{nonlinear_DS} and safe set~$\mathcal{C}$ if there exists an extended class $\mathcal{K}_{\infty}$ function $\gamma(\cdot)$ that satisfies
\begin{equation}
\sup_{u \in\mathcal{U}} \nabla_xB(x)^{\top} \left(g(x) + h(x)u\right) \geq -\gamma(B(x)), \ \forall \ x \in \mathcal{X}.
    \label{CBF_cond}
\end{equation}
    \label{def_CBF}
\end{definition}
\vspace{-7pt}
The set of all control inputs that satisfy the condition in~\eqref{CBF_cond} for each $x \in \mathcal{X}$ is   
\begin{equation}
K(x) := \{u \in \mathcal{U} : \nabla_xB(x)^{\top} \left(g(x) + h(x)u\right) \geq -\gamma(B(x))\}.
    \label{K_cbf}
    \vspace{-5pt}
\end{equation}
The formal result on safety follows from~\cite{ames2019control}.
\begin{theorem}
Let $B$ be a Control Barrier Function (CBF) for a safe set~$\mathcal{C}$ and a nonlinear control affine system~\eqref{nonlinear_DS}. Let $u(x) \in K(x)$ be a locally Lipschitz feedback control law. Then, the following holds: $x(0) \in \mathcal{C} \implies x(t) \in \mathcal{C}$ for all $t \in [0, t_{max})$. If the set $\mathcal{C}$ is compact, then, $\mathcal{C}$ is forward invariant, i.e., $t_{max} = \infty$, and $\mathcal{C}$ is asymptotically stable, i.e., $\lim_{t \to \infty}x(t) \in \mathcal{C}$ for all $x(0) \in \mathcal{X}$. 
    \label{thm:CBF}
\end{theorem}

A \textit{Control Lyapunov Function~(CLF)} $V(\cdot)$ is a special case of a CBF. 
 In particular, for $V(\cdot) : \mathbb{R}^d \to \mathbb{R}_{\geq 0}$ a positive definite function, if we set $B(\cdot) = -V(\cdot)$ and define the singleton safe set $\mathcal{C} = \{0\}$, then Theorem~\ref{thm:CBF} states that any control law $u(x)\in~K(x)$ with $B(\cdot) = -V(\cdot)$ and $\mathcal{C}=\{0\}$ will asymptotically stabilize the system~\eqref{nonlinear_DS} to the origin: see~\cite{ames2019control} for more details.

Since the inequality in~\eqref{K_cbf} is affine in $u$, they can be included in efficient optimization-based controllers for control affine systems. We present such an optimization-based planner in Section~\ref{sec:virtual_control} that has strong stability and safety guarantees as claimed in Theorem~\ref{thm:CBF}, but, with respect to a target trajectory as opposed to a single attractor.
\vspace{-2pt}
\subsection{Computing the Virtual Control Input}
\label{sec:virtual_control}
\vspace{-2pt}
We now show how to incorporate CLFs and CBFs into the DS-based motion plan~\eqref{ref_motion_app} to generate safe and stable motions that converge to a target trajectory, not just a single attractor that is presented in prior work~\cite{SEDS, figueroa2018physically}. In particular, we use the learned NODE $f_\theta$ to generate a nominal motion plan, and compute $u(x)$ using CLFs and CBFs to enforce stability and safety, resulting in a motion plan of the form:
\begin{equation}
\dot{x} = f_{\theta}(x) + u(x),
    \label{inference_DS}
\end{equation}
where $x$ is the state of the robot, and $u(x)$ is the virtual control input.
\subsubsection{Stability using Control Lyapunov Functions}
\label{sec:stability}
We utilize CLFs described in Section~\ref{sec:CBF_def} to generate a motion plan that always converge to the target trajectory $x^*(t)$ even in the presence of disturbances. We note that $x^*(t)$ is different from the unknown target trajectory $z^*(t)$ introduced in Section~\ref{sec:Intro}. Previous work~\cite{CLF_learning} have utilized CLFs only for convergence to a single target point. In contrast, we present a framework that integrates Neural ODEs for rich behaviors, CLFs to ensure convergence to a target trajectory $x^*(t)$, and CBFs for safety. To that end, we first define the error $e(t)$ between the robot state and the target trajectory: $e(t) = x(t) - x^*(t)$. For ease of notation, we drop the explicit dependence on time~$t$, and write $e$, $x$, and $x^*$ for the current error, state, and target point at time $t$, respectively. From~\eqref{inference_DS}, the error dynamics are given by
\begin{equation}
\dot{e} = \dot{x} - \dot{x}^* \Rightarrow \dot{e} =  f_{\theta}(x) - \dot{x}^* + u(x).
    \label{err_dynamics}
\end{equation}
The error dynamics~\eqref{err_dynamics} define a nonlinear control affine system~\eqref{nonlinear_DS}, where the state of the system is $e$, and $u(x)$ is the control input. Hence, by Theorem~\ref{thm:CBF}, if there exists a CLF~$V(\cdot) = -B(\cdot)$ for the error dynamics~\eqref{err_dynamics}, then, any feedback virtual control law $u(\cdot)$ that satisfies
\begin{equation}
\nabla_{e(t)}V(e)^{\top}\left(f_{\theta}(x) - \dot{x}^* + u(x)\right) \leq -\alpha(V(e)), \ \forall \ e \in \mathbb{R}^d
\label{virtual_control_stable}
\end{equation}
will drive the error asymptotically to zero, where~$\alpha(\cdot)$ is a class $\mathcal{K}_{\infty}$ function for CLF, which is introduced to differentiate from $\gamma(\cdot)$ in Definition~\ref{def_CBF} for the CBF. During online motion planning, given the current state of the robot $x$ and information about the target trajectory $x^*$, we compute the minimal control effort $u(x)$ that satisfies~\eqref{virtual_control_stable} by setting
\begin{equation}
    \begin{aligned}
        \quad & u(x) = \argmin_{v} \quad  \bigl\|v\bigr\|^2_2  \\  
        \textnormal{s.t.} \quad &
        \nabla_{e}V(e)^{\top}\left(f_{\theta}(x) - \dot{x}^*+ v \right) \leq -\alpha(V(e)), 
    \end{aligned}
    \label{virtual_control_opt}
\end{equation}
where $\alpha(\cdot)$ defines how aggressively the robot tracks the target trajectory. We describe how we choose $x^*$ and $\dot{x}^*$ in detail in Section~\ref{sec:target_point}. The optimization problem~\eqref{virtual_control_opt} is a quadratic program with a single affine inequality and has a closed form solution~\cite{CLF_learning}. The Lyapunov function we use is $V(e) = \|e\|_2^2$, but, note that any positive definite function is a valid CLF. The presence of the virtual actuation term $v$ makes the optimization problem~\eqref{virtual_control_opt} always feasible. We refer the reader to Fig.~\ref{fig:Spur_att_all} to differentiate between the paths generated by only $f_{\theta}(\cdot)$, and by~\eqref{inference_DS} using the correction term~$u(\cdot)$. We refer to this approach as the CLF-NODE.

\subsubsection{Safety using Control Barrier Functions}
\label{sec:obstacle_avoid}
We build on the framework in Section~\ref{sec:CBF_def} by integrating CBFs into the virtual control input computation to guarantee safety for the generated motion plan. We define safety with respect to a safe set~$\mathcal{C} \subseteq \mathcal{X}$ as described in Section~\ref{sec:CBF_def} for the system~\eqref{inference_DS}. From Theorem~\ref{thm:CBF}, if there exists a CBF~$B(\cdot)$ for the dynamics~\eqref{inference_DS}, then, any feedback control law $u(\cdot)$ that satisfies
\vspace{-5pt}
\begin{equation}
\nabla_{x}B(x)^{\top} \left(f_{\theta}(x) + u(x)\right) \geq -\gamma(B(x)), \ \forall \ x \in \mathcal{X}
    \label{virtual_control_safe}
\vspace{-5pt}
\end{equation}
will render the system~\eqref{inference_DS} safe, where, $\gamma(\cdot)$ is an extended class $\mathcal{K}_{\infty}$ function. At inference, the DS-based motion plan is still given by~\eqref{inference_DS}, but the virtual control input $u(x)$ is computed such that it satisfies the CBF condition in~\eqref{virtual_control_safe} for the dynamics $\dot{x}$ and a given CBF~$B(\cdot)$ for the safe set~$\mathcal{C}$. 

In cases where an obstacle obstructs the robot moving along the nominal trajectory, the robot should automatically avoid the obstacle, but converge back to complete the desired task when possible.  However, this may lead to a conflict between preserving safety and stability: during the obstacle avoidance phase, the CLF constraint in~\eqref{virtual_control_opt} may be violated as the robot takes safety preserving actions that increase tracking error. We prioritize safety and adapt the approach proposed in~\cite{ames2019control} for balancing these competing objectives to our setting, and solve an optimization problem with the CBF condition~\eqref{virtual_control_safe} as a hard constraint and the CLF condition~\eqref{virtual_control_stable} as a soft constraint. Given the current state of the robot~$x$ and the target point~$x^*$, the optimization problem that guarantees a safe motion plan is
\begin{equation}
    \begin{aligned}
\vspace{-10pt}
        \quad & (u(x), \_) = \argmin_{\{v, \epsilon\}} \quad \bigl\|v\bigr\|^2_2 + \lambda \epsilon^2 \\
        \textnormal{s.t.} \quad & \nabla_{x}B(x)^{\top}\left(f_{\theta}(x) + v\right) \geq -\gamma(B(x))\\
        \vspace{-10pt}
        \quad & \nabla_{e}V(e)^{\top}\left(f_{\theta}(x)  - \dot{x}^* + v\right) \leq -\alpha(V(e)) + \epsilon
    \end{aligned}
\vspace{-5pt}
    \label{safety_opt}
\end{equation}
where $\epsilon$ is a relaxation variable to ensure feasibility of~\eqref{safety_opt} and is penalized by $\lambda > 0$. The problem in~\eqref{safety_opt} is a parametric program, where the parameters of interest are $\{x, x^*, \dot{x}^*\}$. We abuse notation and denote the optimal virtual control input $u(x)$ for~\eqref{safety_opt} to be $u^*(x, x^*, \dot{x}^*)$ which will be used in the next section. Problem~\eqref{safety_opt} is a QP that we solve efficiently in real-time using OSQP solver~\cite{osqp}. Multiple CBFs and CLFs can be composed in a way analogous to problem~\eqref{safety_opt} to represent multiple obstacles of complex non-convex shapes as given in Section~\ref{sec:experiments}. We refer to this approach as the CLF-CBF-NODE.

\subsubsection{Choosing a Target Point}
\label{sec:target_point}

As shown in Fig.~\ref{fig:block_diag}, we first integrate the learnt model $f_{\theta}(\cdot)$ offline to generate the \textit{target array}~${\mathcal{T} := \{x^*(t_k)\}_{k=1}^T}$ from a given initial condition $x^*(t_1)$. The target array~$\mathcal{T}$ is exactly the target trajectory $x^*(t)$, but only at time steps~$\{{t_k}\}_{k=1}^T$. Given an observation of the current state of the robot~$x$ at time~$t$, and the target array~$\mathcal{T}$, we select the next target point~$x^*$ for the robot to follow using the map $\pi(x)$ defined in Algorithm~\ref{alg:target}. We remove the direct dependence of the target point $x^*$ on time~$t$, which leads to a more reactive motion plan that adapts to both the time delays that are often present during online deployment, and to unforeseen perturbations of the robot away from the nominal plan, e.g., due to human interaction or obstacle avoidance.  The look-ahead horizon length~$N$ is used to construct the array~$\mathcal{T}_N$, consisting of $N$ future points starting at the current target state~$x^*(t_m)$. We choose the target point $\pi(x)$ from $\mathcal{T}_N$ that results in the smallest norm of virtual control input when solving~\eqref{safety_opt} for all~$y \in\mathcal{T}_N$. We use a forward looking horizon~$N$ to ensure the robot moves forward along the target trajectory, which is illustrated in the video submission. To the best of our knowledge, this is the first time that the norm of the correction input~$u(\cdot)$ is used as a metric for choosing an appropriate nearest neighbor point in motion planning. We use $\dot{x}^* := f_{\theta}(\pi(x))$, since $\pi(x) \in \mathcal{T}$ and we obtained the target array~$\mathcal{T}$ by integrating $f_{\theta}(\cdot)$.

\RestyleAlgo{ruled}
\setlength{\textfloatsep}{0pt}
\begin{algorithm}[!t]
\footnotesize
\caption{Choose target point}
\label{alg:target}
\KwData{$\mathcal{T}:=\{x^*(t_k)\}_{k=1}^T, f_{\theta}(\cdot), x, N$}
\KwResult{$\pi(x)$}
$m \gets \argmin_{k} \big\|x - x^*(t_k)\big\|_2$\;
${\mathcal{T}_N :=\{x^*(t_m), x^*(t_{m+1}), \ldots, x^*(t_{m+N-1})\}}$\;
Solve~\eqref{safety_opt} with parameters $\{x, y, f_{\theta}(y)\}$ for~each~$y \in \mathcal{T}_N$\;
$\pi(x) \gets \argmin_{y \in \mathcal{T}_N} \big\|u^*(x, y, f_{\theta}(y))\big\|_2^2$\;
  \end{algorithm}

\vspace{-5pt}
\section{Experimental Validation and Results}
\label{sec:experiments}
\vspace{-2pt}

\subsection{LASA handwriting dataset}
\label{sec:LASA_exp}

\begin{figure}[!t]
\centering
     \begin{subfigure}[t]{0.4\linewidth}\centering        \includegraphics[width=\textwidth]{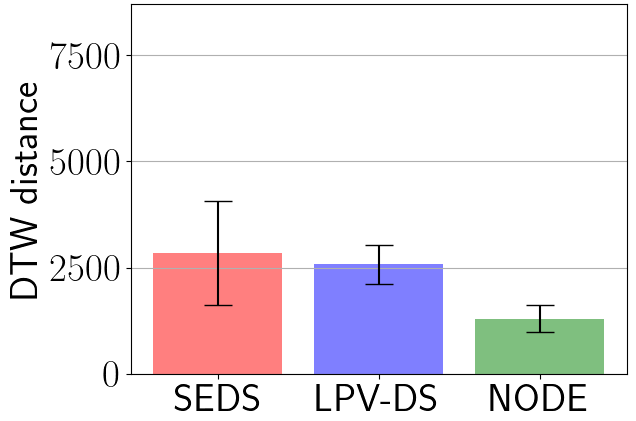}      
     \caption{Train data set}
     \label{fig:dtw_dist_train}
     \end{subfigure}
     \begin{subfigure}[t]{0.4\linewidth}  
     \centering        \includegraphics[width=\textwidth]{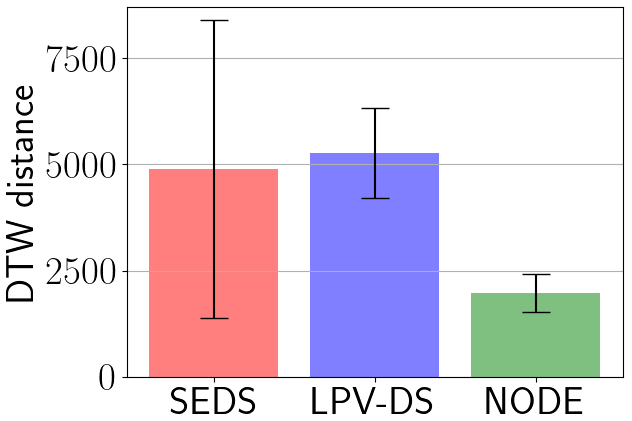}
     \caption{Test data set}\label{fig:dtw_dist_test}
     \end{subfigure}
     \caption{Comparison of DTW distance on the LASA data set.}
     \label{fig:LASA_dtwd}
     \end{figure}

\begin{figure}[!t]
 \vspace{-10 pt}
 \centering
          \begin{subfigure}[t]{0.49\linewidth}  
     \centering        \includegraphics[width=\textwidth]{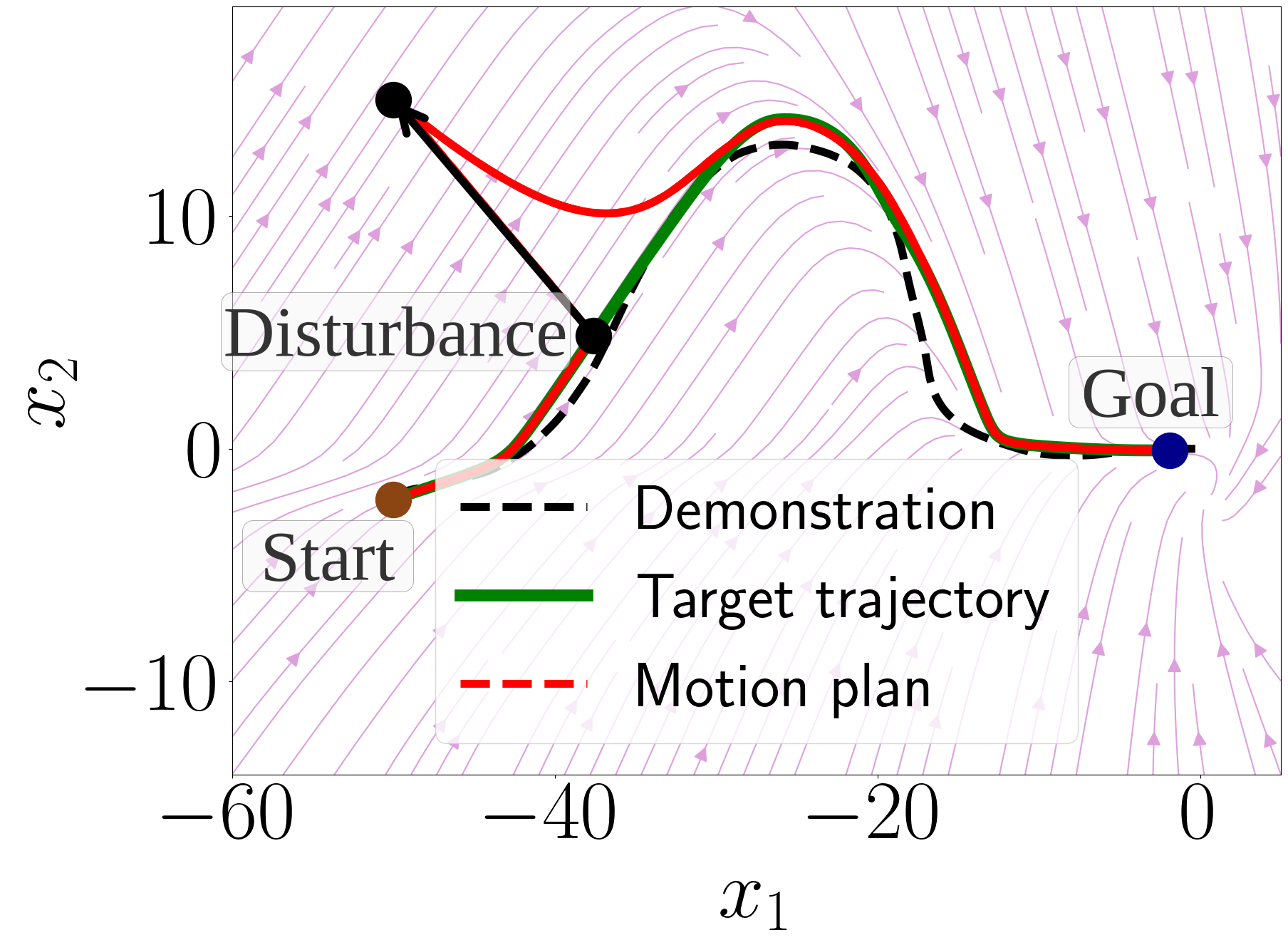}
         \caption{\textit{Worm} with disturbance}
         \label{fig:worm_dist}
     \end{subfigure}
          \begin{subfigure}[t]{0.49\linewidth}  
     \centering          \includegraphics[width=\textwidth]{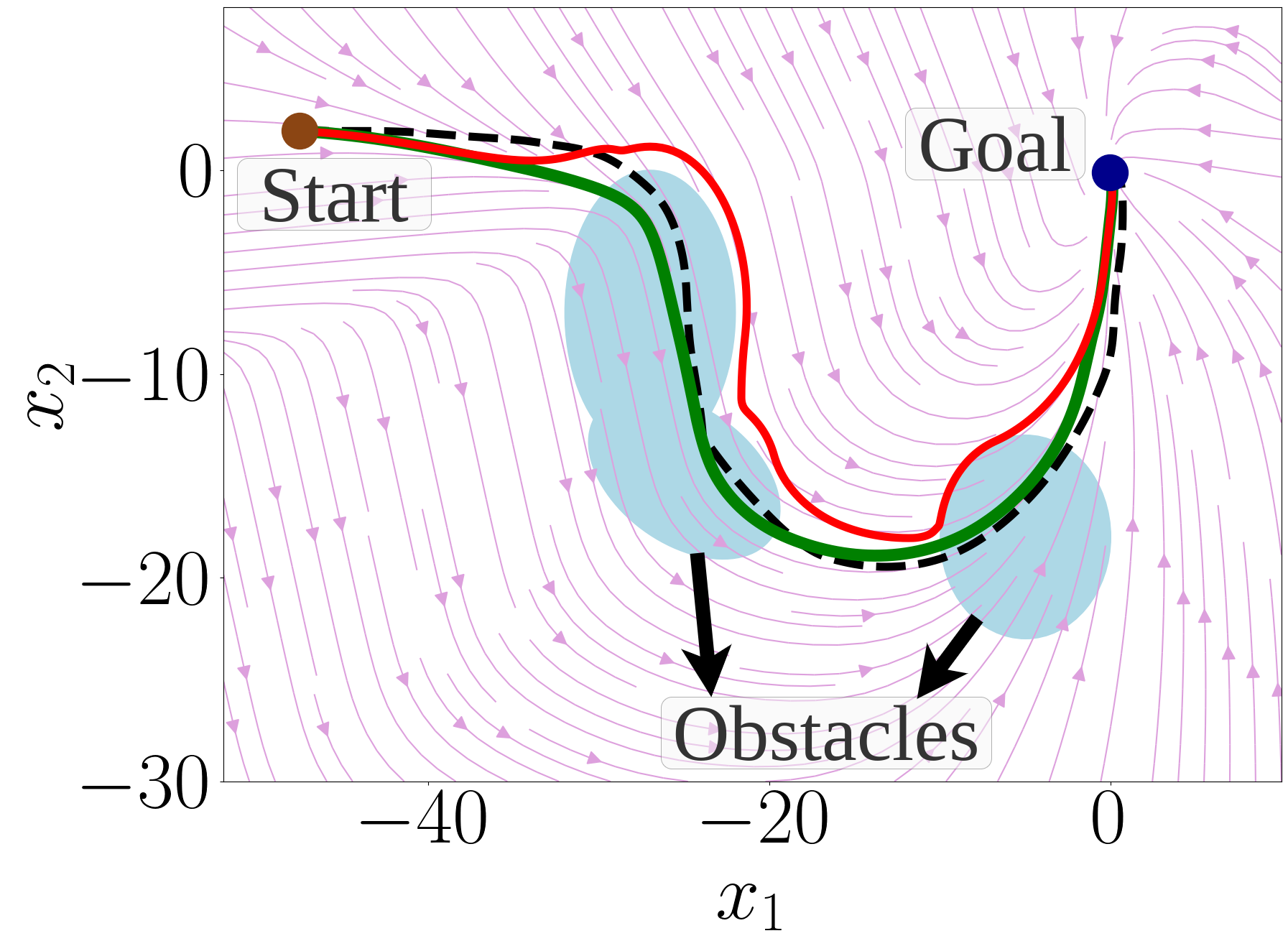}
         \caption{\textit{Spoon} with obstacles}
         \label{fig:trapezoid_obst}
     \end{subfigure}
        \caption{Illustration of (a) disturbance rejection using CLF-NODE and (b) obstacle avoidance using CLF-CBF-NODE.}
        \label{fig:LASA_stable_safe}
\end{figure}

We validate our approach on the LASA handwriting data set~\cite{SEDS} that contains 30 2D nonlinear motions. Each motion set has~$7$ demonstrations: we use 4 as the training set, and the remaining 3 as the test set. In Fig.~\ref{fig:LASA_dtwd}, we compare the performance of our NODE model with two existing DS-based learning approaches: SEDS~\cite{SEDS} and LPV-DS~\cite{figueroa2018physically} using the mean and variance of Dynamic Time Warping~(DTW) distance~\cite{fastDTW}. DTW distance measures the dissimilarity between the demonstrations and the corresponding reproductions starting from the same initial condition. We note that although SEDS and LPV-DS use velocity data for regression, which our approach does not have access to, the mean of the DTW distance for our NODE approach is approximately half of the existing methods~\cite{SEDS, figueroa2018physically}. We illustrate disturbance rejection in Fig.~\ref{fig:worm_dist} using CLF-NODE and obstacle avoidance in Fig.~\ref{fig:trapezoid_obst} using CLF-CBF-NODE with a circular obstacle and a non-convex obstacle. The barrier function~$B(\cdot)$ is defined to be less than zero inside the obstacles, and we choose linear $\alpha(\cdot)$ and  $\gamma(\cdot)$ functions.
\vspace{-4pt}
\subsection{Periodic trajectories}
\vspace{-1pt}

\begin{figure}[!t]
         \vspace{-10pt}
\centering
     \begin{subfigure}[b]{0.3\linewidth}  
     \centering        \includegraphics[width=\textwidth]{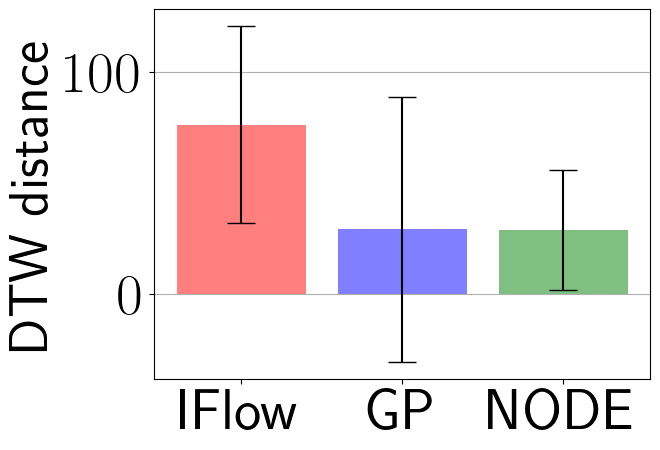}      
     \caption{DTW distance}
     \label{fig:dtw_dist_IFlow}
     \end{subfigure}     
          \begin{subfigure}[b]{0.3\linewidth} 
     \centering        \includegraphics[width=\textwidth]{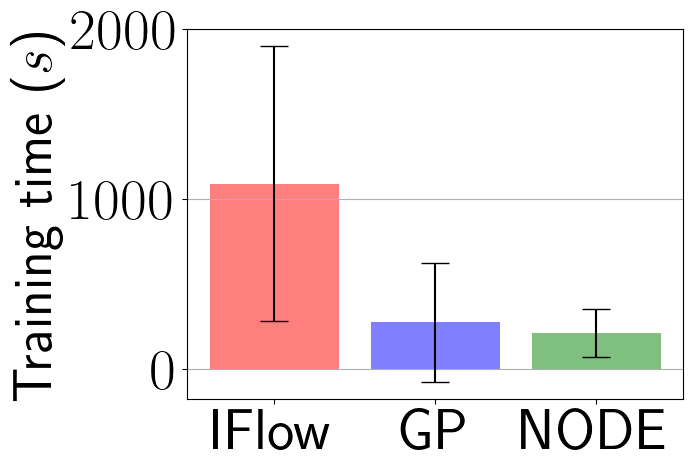}
         \caption{Training time}
         \label{fig:train_time_IFlow}
     \end{subfigure}
          \begin{subfigure}[b]{0.35\linewidth} 
     \centering          \includegraphics[width=\textwidth]{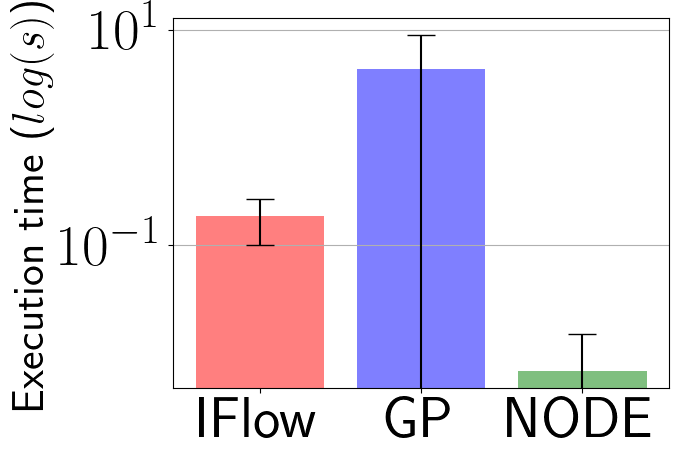}
         \caption{Execution time}
         \label{fig:exec_time_IFlow}
     \end{subfigure}
     \begin{subfigure}[b]{0.4\linewidth}  
     \centering        \includegraphics[width=\textwidth]{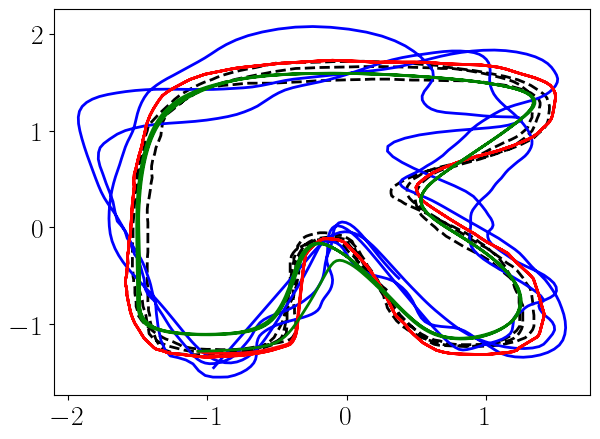}
     \caption{R shape}\label{fig:RShape}
     \end{subfigure}
          \begin{subfigure}[b]{0.48\linewidth}  
     \centering          \includegraphics[width=\textwidth]{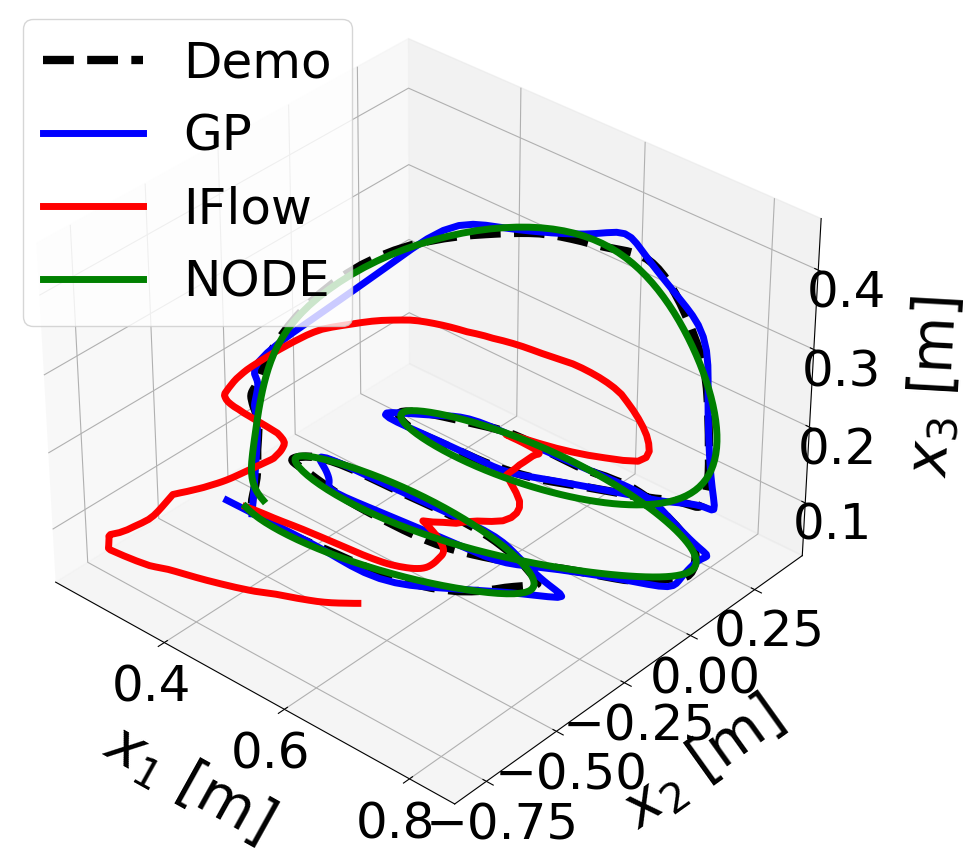}
         \caption{Spiral wiping}
         \label{fig:IFlow_Spiral}
     \end{subfigure}
     \caption{Comparison of performance metrics and trajectory reproductions between IFlow, GP and our approach (NODE) on periodic trajectories.}
     \label{fig:periodic_metrics}
     \end{figure}

We validate our approach on handwritten 2D periodic motions of the letters \textbf{I, R, O} and \textbf{S} given in~\cite{imit_norm} and 3D periodic trajectories that encode three wiping tasks as given in Figs.~\ref{fig:IFlow_Spiral},~\ref{fig:wiping_randy}, and~\ref{fig:wiping_white_board}. We compare our method with Imitation Flow (IFlow)~\cite{imit_norm} and a Gaussian Process~(GP)~\cite{LfD_GP} based approach, since SEDS and LPV-DS compared in Section~\ref{sec:LASA_exp} model trajectories that converge only to a single attractor. IFlow is based on normalizing flows that learns stable motions using prior knowledge on whether the underlying dynamics of the demonstrations has a single attractor or a limit cycle, but our approach requires no prior knowledge. We present the DTW distance in Fig.~\ref{fig:dtw_dist_IFlow}, training time in Fig.~\ref{fig:train_time_IFlow} and execution time in Fig.~\ref{fig:exec_time_IFlow}. The execution time is the computation time for a single forward pass of the model to generate the entire trajectory from a given initial point and the time span of the demonstration. We also compare the trajectory reproductions for the \textbf{R}~shape in Fig.~\ref{fig:RShape}, and for the spiral wiping task by the Franka robot arm in Fig.~\ref{fig:IFlow_Spiral}. The IFlow approach is not able to learn the complex motion for spiral wiping, but our NODE approach learns with high accuracy and lesser computation time. The execution time comparison in Fig.~\ref{fig:exec_time_IFlow} is plotted in $\log$ scale and we note that our approach (NODE) has much lesser execution time, which is important for real-time robot experiments. Although the GP based method~\cite{LfD_GP} learns complex trajectories with comparable accuracy and training time to NODE, the execution time is much longer and they rely on time inputs for desired roll outs with no capability to generate safe and stable motion plans. All computations are performed on Google Colab. 

     \vspace{-5 pt}

\subsection{Robotic experiments}
\begin{figure}[!t]
\vspace{-5pt}
\centering
     \begin{subfigure}[b]{0.49\linewidth}         \centering        \includegraphics[width=\textwidth]{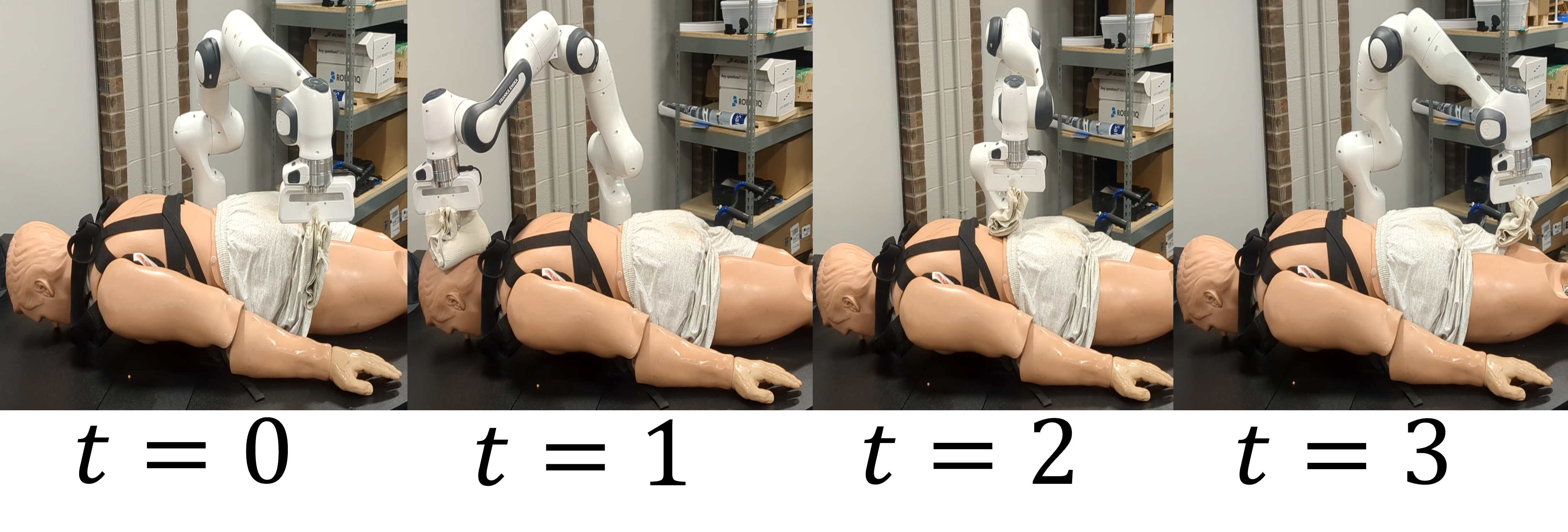}
     \end{subfigure}
     \begin{subfigure}[b]{0.49\linewidth}  
     \centering        \includegraphics[width=\textwidth]{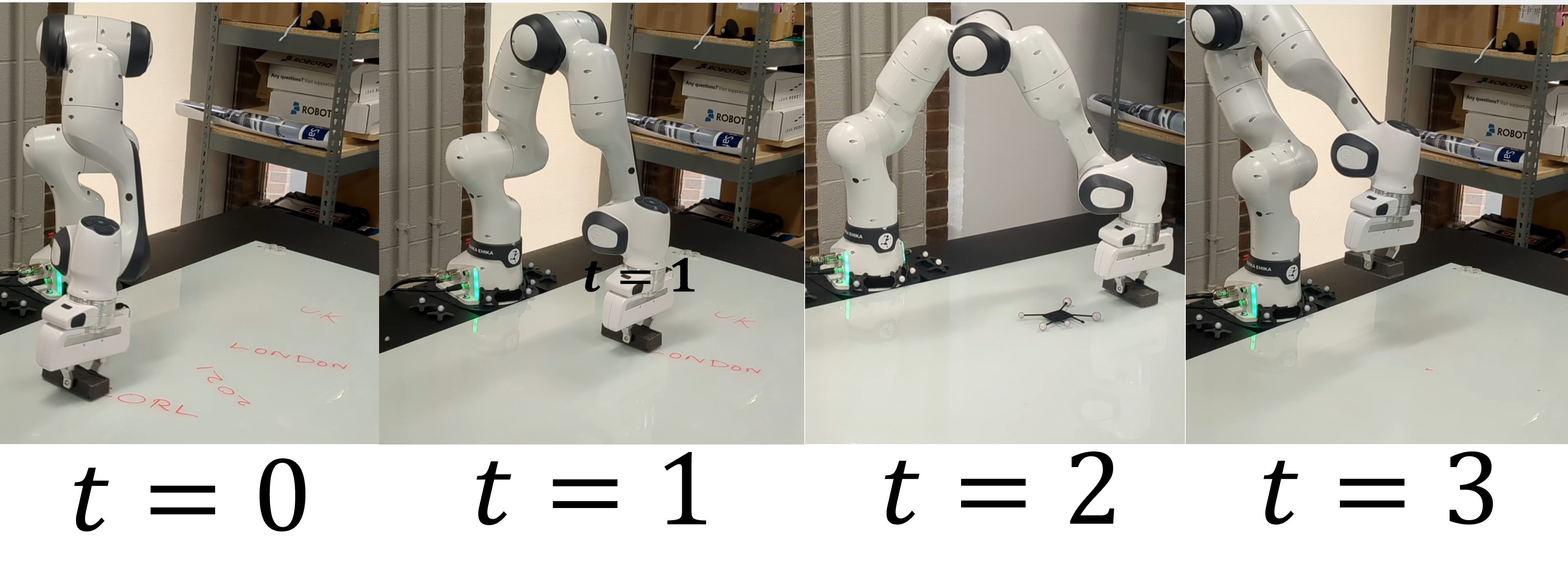}
     \end{subfigure}    
     \begin{subfigure}[b]{0.47\linewidth}         \centering        \includegraphics[width=\textwidth]{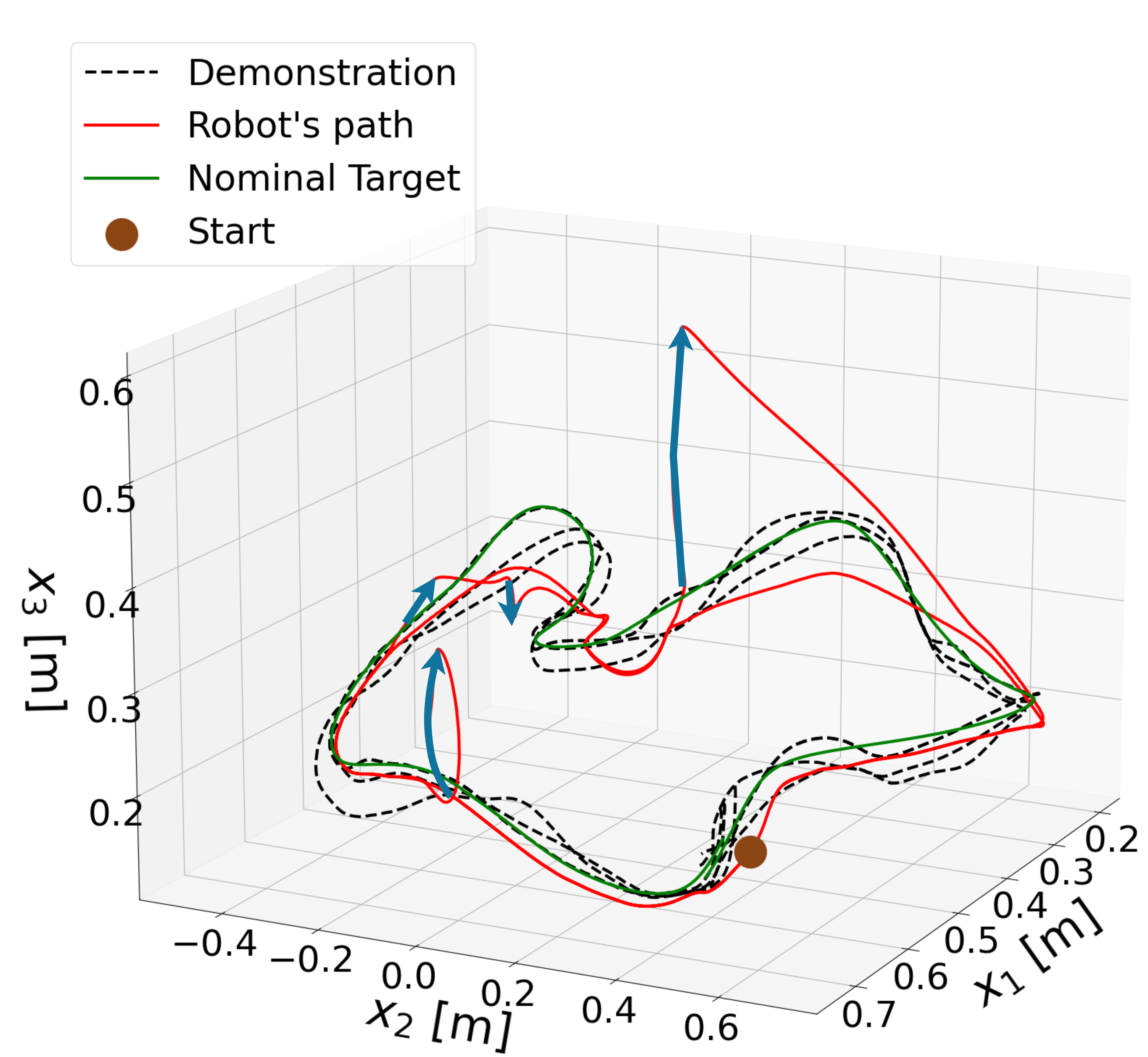} 
     \caption{Wiping a mannequin}
     \label{fig:wiping_randy}
     \end{subfigure} 
     \begin{subfigure}[b]{0.47\linewidth}  
     \centering        \includegraphics[width=\textwidth]{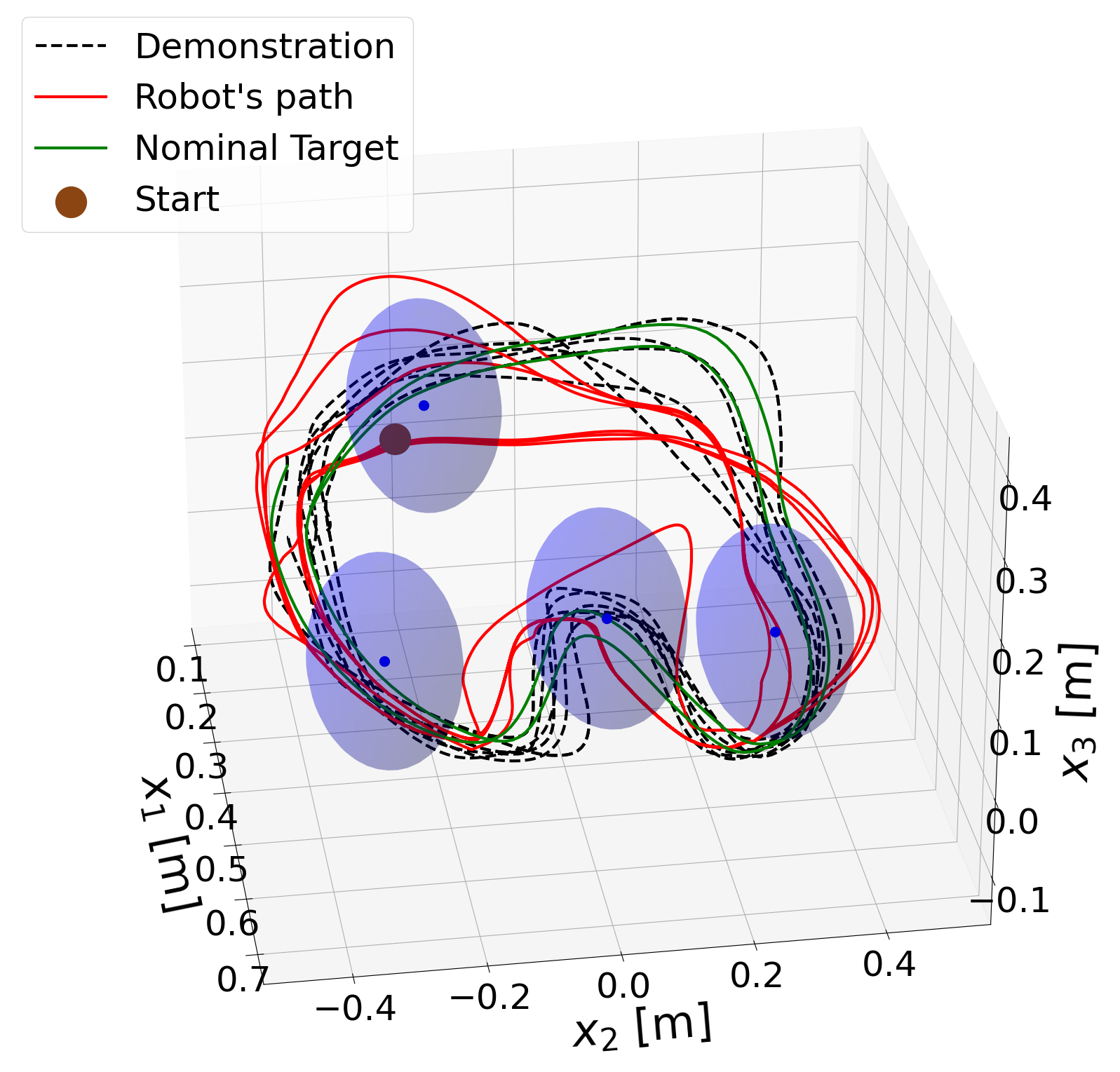}
     \caption{Wiping a white board}
     \label{fig:wiping_white_board}
     \end{subfigure}
             \caption{The robot performing different wiping tasks with complex periodic motions. The blue arrows denote the perturbations, and the purple spheres is the dynamic obstacle at different time.}      
     \label{fig:wiping}
     \end{figure}

We validate our approach on the Franka robot arm performing complex periodic motions: wiping a mannequin with a towel~(Fig.~\ref{fig:wiping_randy}) and wiping a white board with an eraser~(Fig.~\ref{fig:wiping_white_board}). We use the passive DS impedance controller~\cite{passive_DS}. We used $2$ demonstrations for the mannequin task, and $3$ demonstrations for the board wiping task. Each demonstration had between $300$ and $600$~data samples. The average training time (offline) is~$3-6$ minutes for each task on Google Colab. The obstacle shown in Fig.~\ref{fig:wiping_white_board} at $t=2$ has markers on it that are tracked in real-time by OptiTrack motion capture system. We observe that the robot tracks the desired nominal trajectories while remaining compliant to human interaction, robust to perturbations, and safe with respect to unforeseen dynamic obstacles. The online motion planning module runs at 1KHz that solves~\eqref{safety_opt} and implements Algorithm~\ref{alg:target}.  We include the stirring task, the effect of Algorithm~\ref{alg:target} and some preliminary results for full pose (position in~$\mathbb{R}^3$ and orientation in~$\mathcal{SO}(3)$) motion planning in~\url{https://sites.google.com/view/lfd-neural-ode/home} and the video submission.
\vspace{-5pt}
\section{Conclusion \& Future Work}    
\label{sec:limit_concl}
\vspace{-5pt}
We propose a modular DS-based motion planner using NODE with an additive CLF and CBF-based correction term that guarantees stability and safety on the error dynamics for convergence to a target trajectory, rather than a single attractor.  We validate our approach on complex non-linear and periodic handwritten motions and on the Franka robot arm for wiping and stirring tasks. 

Our robot experiments are limited to the Cartesian end-effector position: future work will address this by extending our approach to higher dimensions such as (a) orientation in~$\mathcal{SO}(3)$ and position in~$\mathbb{R}^3$ of the end-effector~\cite{SO3_1, SO3_2, SO3_3}, (b) full pose of end-effector in~$\mathcal{SE}(3)$~\cite{SE3_1}, and (c) joint space~\cite{Joint} that can include joint limits and avoid collision with arm links. Future work will also incorporate the CLF-CBF NODE approach with end-to-end learning frameworks using image observations~\cite{end_end, end_end_1}.



\clearpage


\bibliographystyle{IEEEtran}
\bibliography{IEEEabrv, root}

\begin{thebibliography}{10}
\providecommand{\url}[1]{#1}
\csname url@rmstyle\endcsname
\providecommand{\newblock}{\relax}
\providecommand{\bibinfo}[2]{#2}
\providecommand\BIBentrySTDinterwordspacing{\spaceskip=0pt\relax}
\providecommand\BIBentryALTinterwordstretchfactor{4}
\providecommand\BIBentryALTinterwordspacing{\spaceskip=\fontdimen2\font plus
\BIBentryALTinterwordstretchfactor\fontdimen3\font minus \fontdimen4\font\relax}
\providecommand\BIBforeignlanguage[2]{{%
\expandafter\ifx\csname l@#1\endcsname\relax
\typeout{** WARNING: IEEEtran.bst: No hyphenation pattern has been}%
\typeout{** loaded for the language `#1'. Using the pattern for}%
\typeout{** the default language instead.}%
\else
\language=\csname l@#1\endcsname
\fi
#2}}

\bibitem{pick_place}
R.~Ma, J.~Chen, and J.~Oyekan, ``A learning from demonstration framework for adaptive task and motion planning in varying package-to-order scenarios,'' \emph{Robotics and Computer-Integrated Manufacturing}, vol.~82, p. 102539, 2023.

\bibitem{periodic}
J.~Yang, J.~Zhang, C.~Settle, A.~Rai, R.~Antonova, and J.~Bohg, ``Learning periodic tasks from human demonstrations,'' in \emph{2022 International Conference on Robotics and Automation (ICRA)}.\hskip 1em plus 0.5em minus 0.4em\relax IEEE, 2022, pp. 8658--8665.

\bibitem{SEDS}
S.~M. Khansari-Zadeh and A.~Billard, ``Learning stable nonlinear dynamical systems with gaussian mixture models,'' \emph{IEEE Transactions on Robotics}, vol.~27, no.~5, pp. 943--957, 2011.

\bibitem{DMP_related}
A.~J. Ijspeert, J.~Nakanishi, H.~Hoffmann, P.~Pastor, and S.~Schaal, ``Dynamical movement primitives: learning attractor models for motor behaviors,'' \emph{Neural computation}, vol.~25, no.~2, pp. 328--373, 2013.

\bibitem{kinesthetic}
B.~Akgun and K.~Subramanian, ``Robot learning from demonstration : Kinesthetic teaching vs . teleoperation,'' 2011.

\bibitem{abbeel2004apprenticeship}
P.~Abbeel and A.~Y. Ng, ``Apprenticeship learning via inverse reinforcement learning,'' in \emph{Proceedings of the twenty-first international conference on Machine learning}, 2004, p.~1.

\bibitem{IOC}
M.~C. Priess, J.~Choi, and C.~Radcliffe, ``The inverse problem of continuous-time linear quadratic gaussian control with application to biological systems analysis,'' in \emph{Dynamic Systems and Control Conference}, vol. 46209.\hskip 1em plus 0.5em minus 0.4em\relax American Society of Mechanical Engineers, 2014, p. V003T42A004.

\bibitem{BC}
T.~Osa, J.~Pajarinen, G.~Neumann, J.~A. Bagnell, P.~Abbeel, J.~Peters, \emph{et~al.}, ``An algorithmic perspective on imitation learning,'' \emph{Foundations and Trends{\textregistered} in Robotics}, vol.~7, no. 1-2, pp. 1--179, 2018.

\bibitem{DAGGER}
S.~Ross, G.~Gordon, and D.~Bagnell, ``A reduction of imitation learning and structured prediction to no-regret online learning,'' in \emph{Proceedings of the fourteenth international conference on artificial intelligence and statistics}.\hskip 1em plus 0.5em minus 0.4em\relax JMLR Workshop and Conference Proceedings, 2011, pp. 627--635.

\bibitem{deep_imit}
T.~Zhang, Z.~McCarthy, O.~Jow, D.~Lee, X.~Chen, K.~Goldberg, and P.~Abbeel, ``Deep imitation learning for complex manipulation tasks from virtual reality teleoperation,'' in \emph{2018 IEEE International Conference on Robotics and Automation (ICRA)}, 2018, pp. 5628--5635.

\bibitem{LfD_GP}
N.~Jaquier, D.~Ginsbourger, and S.~Calinon, ``Learning from demonstration with model-based gaussian process. arxiv preprint arxiv: 191005005,'' 2019.

\bibitem{billard2022learning}
A.~Billard, S.~Mirrazavi, and N.~Figueroa, \emph{Learning for Adaptive and Reactive Robot Control: A Dynamical Systems Approach}.\hskip 1em plus 0.5em minus 0.4em\relax MIT Press, 2022.

\bibitem{figueroa2018physically}
N.~Figueroa and A.~Billard, ``A physically-consistent bayesian non-parametric mixture model for dynamical system learning,'' in \emph{Proceedings of The 2nd Conference on Robot Learning}, ser. Proceedings of Machine Learning Research, vol.~87.\hskip 1em plus 0.5em minus 0.4em\relax PMLR, 29--31 Oct 2018, pp. 927--946.

\bibitem{multi_arm}
S.~S. Mirrazavi~Salehian, N.~Figueroa, and A.~Billard, ``A unified framework for coordinated multi-arm motion planning,'' \emph{The International Journal of Robotics Research}, vol.~37, no.~10, pp. 1205--1232, 2018.

\bibitem{SO3_1}
N.~Figueroa, S.~Faraji, M.~Koptev, and A.~Billard, ``A dynamical system approach for adaptive grasping, navigation and co-manipulation with humanoid robots,'' in \emph{2020 IEEE International Conference on Robotics and Automation (ICRA)}, 2020, pp. 7676--7682.

\bibitem{figueroa2022locally}
N.~Figueroa and A.~Billard, ``Locally active globally stable dynamical systems: Theory, learning, and experiments,'' \emph{The International Journal of Robotics Research}, vol.~41, no.~3, pp. 312--347, 2022.

\bibitem{imit_norm}
J.~Urain, M.~Ginesi, D.~Tateo, and J.~Peters, ``Imitationflow: Learning deep stable stochastic dynamic systems by normalizing flows,'' in \emph{2020 IEEE/RSJ International Conference on Intelligent Robots and Systems (IROS)}.\hskip 1em plus 0.5em minus 0.4em\relax IEEE, 2020, pp. 5231--5237.

\bibitem{ames2019control}
A.~D. Ames, S.~Coogan, M.~Egerstedt, G.~Notomista, K.~Sreenath, and P.~Tabuada, ``Control barrier functions: Theory and applications,'' in \emph{2019 18th European control conference (ECC)}.\hskip 1em plus 0.5em minus 0.4em\relax IEEE, 2019, pp. 3420--3431.

\bibitem{wang2022temporal}
\BIBentryALTinterwordspacing
Y.~Wang, N.~Figueroa, S.~Li, A.~Shah, and J.~Shah, ``Temporal logic imitation: Learning plan-satisficing motion policies from demonstrations,'' in \emph{6th Annual Conference on Robot Learning}, 2022. [Online]. Available: \url{https://openreview.net/forum?id=ndYsaoyzCWv}
\BIBentrySTDinterwordspacing

\bibitem{CLF_learning}
S.~M. Khansari-Zadeh and A.~Billard, ``Learning control lyapunov function to ensure stability of dynamical system-based robot reaching motions,'' \emph{Robotics and Autonomous Systems}, vol.~62, no.~6, pp. 752--765, 2014.

\bibitem{cbf_learning}
A.~Robey, H.~Hu, L.~Lindemann, H.~Zhang, D.~V. Dimarogonas, S.~Tu, and N.~Matni, ``Learning control barrier functions from expert demonstrations,'' in \emph{2020 59th IEEE Conference on Decision and Control (CDC)}.\hskip 1em plus 0.5em minus 0.4em\relax IEEE, 2020, pp. 3717--3724.

\bibitem{7358081}
K.~Kronander and A.~Billard, ``Passive interaction control with dynamical systems,'' \emph{IEEE Robotics and Automation Letters}, vol.~1, no.~1, pp. 106--113, 2016.

\bibitem{neural_event}
R.~T. Chen, B.~Amos, and M.~Nickel, ``Learning neural event functions for ordinary differential equations,'' \emph{arXiv preprint arXiv:2011.03902}, 2020.

\bibitem{chen2019neural}
R.~T.~Q. Chen, Y.~Rubanova, J.~Bettencourt, and D.~Duvenaud, ``Neural ordinary differential equations,'' 2019.

\bibitem{haber2017stable}
E.~Haber and L.~Ruthotto, ``Stable architectures for deep neural networks,'' \emph{Inverse problems}, vol.~34, no.~1, p. 014004, 2017.

\bibitem{rungekutta}
J.~C. Butcher, ``A history of runge-kutta methods,'' \emph{Applied numerical mathematics}, vol.~20, no.~3, pp. 247--260, 1996.

\bibitem{kidger2022neural}
P.~Kidger, ``On neural differential equations,'' \emph{arXiv preprint arXiv:2202.02435}, 2022.

\bibitem{only_pos}
S.~Purwar, I.~N. Kar, and A.~N. Jha, ``Adaptive output feedback tracking control of robot manipulators using position measurements only,'' \emph{Expert systems with applications}, vol.~34, no.~4, pp. 2789--2798, 2008.

\bibitem{vel_noisy}
B.~Xiao, L.~Cao, S.~Xu, and L.~Liu, ``Robust tracking control of robot manipulators with actuator faults and joint velocity measurement uncertainty,'' \emph{IEEE/ASME Transactions on Mechatronics}, vol.~25, no.~3, pp. 1354--1365, 2020.

\bibitem{obst_compare}
A.~Singletary, K.~Klingebiel, J.~Bourne, A.~Browning, P.~Tokumaru, and A.~Ames, ``Comparative analysis of control barrier functions and artificial potential fields for obstacle avoidance,'' in \emph{2021 IEEE/RSJ International Conference on Intelligent Robots and Systems (IROS)}.\hskip 1em plus 0.5em minus 0.4em\relax IEEE, 2021, pp. 8129--8136.

\bibitem{APF_slow}
Y.~Chen, L.~Chen, J.~Ding, and Y.~Liu, ``Research on real-time obstacle avoidance motion planning of industrial robotic arm based on artificial potential field method in joint space,'' \emph{Applied Sciences}, vol.~13, no.~12, p. 6973, 2023.

\bibitem{mod_obst}
S.~M. Khansari-Zadeh and A.~Billard, ``A dynamical system approach to realtime obstacle avoidance,'' \emph{Autonomous Robots}, vol.~32, pp. 433--454, 2012.

\bibitem{bio_obst}
H.~Hoffmann, P.~Pastor, D.-H. Park, and S.~Schaal, ``Biologically-inspired dynamical systems for movement generation: Automatic real-time goal adaptation and obstacle avoidance,'' in \emph{2009 IEEE international conference on robotics and automation}.\hskip 1em plus 0.5em minus 0.4em\relax IEEE, 2009, pp. 2587--2592.

\bibitem{osqp}
\BIBentryALTinterwordspacing
B.~Stellato, G.~Banjac, P.~Goulart, A.~Bemporad, and S.~Boyd, ``{OSQP}: an operator splitting solver for quadratic programs,'' \emph{Mathematical Programming Computation}, vol.~12, no.~4, pp. 637--672, 2020. [Online]. Available: \url{https://doi.org/10.1007/s12532-020-00179-2}
\BIBentrySTDinterwordspacing

\bibitem{fastDTW}
S.~Salvador and P.~Chan, ``Toward accurate dynamic time warping in linear time and space,'' \emph{Intelligent Data Analysis}, vol.~11, no.~5, pp. 561--580, 2007.

\bibitem{passive_DS}
K.~Kronander and A.~Billard, ``Passive interaction control with dynamical systems,'' \emph{IEEE Robotics and Automation Letters}, vol.~1, no.~1, pp. 106--113, 2016.

\bibitem{SO3_2}
H.~C. Ravichandar and A.~Dani, ``Learning position and orientation dynamics from demonstrations via contraction analysis,'' \emph{Autonomous Robots}, vol.~43, no.~4, pp. 897--912, 2019.

\bibitem{SO3_3}
J.~Zhang, H.~B. Mohammadi, and L.~Rozo, ``Learning riemannian stable dynamical systems via diffeomorphisms,'' in \emph{6th Annual Conference on Robot Learning}, 2022.

\bibitem{SE3_1}
J.~Urain, D.~Tateo, and J.~Peters, ``Learning stable vector fields on lie groups,'' \emph{IEEE Robotics and Automation Letters}, vol.~7, no.~4, pp. 12\,569--12\,576, 2022.

\bibitem{Joint}
Y.~Shavit, N.~Figueroa, S.~S.~M. Salehian, and A.~Billard, ``Learning augmented joint-space task-oriented dynamical systems: A linear parameter varying and synergetic control approach,'' \emph{IEEE Robotics and Automation Letters}, vol.~3, no.~3, pp. 2718--2725, 2018.

\bibitem{end_end}
D.~Totsila, K.~Chatzilygeroudis, D.~Hadjivelichkov, V.~Modugno, I.~Hatzilygeroudis, and D.~Kanoulas, ``End-to-end stable imitation learning via autonomous neural dynamic policies,'' \emph{arXiv preprint arXiv:2305.12886}, 2023.

\bibitem{end_end_1}
S.~Bahl, M.~Mukadam, A.~Gupta, and D.~Pathak, ``Neural dynamic policies for end-to-end sensorimotor learning,'' \emph{Advances in Neural Information Processing Systems}, vol.~33, pp. 5058--5069, 2020.

\bibitem{micro_lie}
J.~Sola, J.~Deray, and D.~Atchuthan, ``A micro lie theory for state estimation in robotics,'' \emph{arXiv preprint arXiv:1812.01537}, 2018.

\bibitem{quat_dyn}
B.~Graf, ``Quaternions and dynamics,'' \emph{arXiv preprint arXiv:0811.2889}, 2008.

\bibitem{merge_pos_rot}
M.~Saveriano, F.~Franzel, and D.~Lee, ``Merging position and orientation motion primitives,'' in \emph{2019 International Conference on Robotics and Automation (ICRA)}.\hskip 1em plus 0.5em minus 0.4em\relax IEEE, 2019, pp. 7041--7047.

\bibitem{clfd_sNODE}
S.~Auddy, J.~Hollenstein, M.~Saveriano, A.~Rodr{\'\i}guez-S{\'a}nchez, and J.~Piater, ``Scalable and efficient continual learning from demonstration via a hypernetwork-generated stable dynamics model,'' \emph{arXiv preprint arXiv:2311.03600}, 2023.

\end{thebibliography}

\section*{APPENDIX}

\subsection{Full Pose Motions}

We extend the CLF-NODE approach presented in Sections~\ref{sec:learning_method} and~\ref{sec:stability} to learn full pose trajectories and generate stable motions in $\mathcal{SE}(3)$ space.

\subsubsection{Neural ODE model}

We extend the Neural ODE model~\eqref{nODE_model} to learn both translational and rotational motions, i.e., the state~$x(t)$ is the full-pose of the end effector. The state is represented as $x(t) = \begin{bmatrix} p(t), & q(t)\end{bmatrix}^{\top}$ where $p(t) \in \mathbb{R}^3$ is the Cartesian coordinates and $q(t) \in \mathcal{S}^3 \subset \mathbb{R}^4$ lives on the 3-sphere~$\mathcal{S}^3$ representing the rotation as a unit quaternion~\cite{micro_lie}. 

\textbf{Quaternion Multiplication:} We denote the operator~$\otimes$ to represent the product between two quaternions, which is a linear operation~\cite{quat_dyn} detailed as follows. Let $q = \begin{bmatrix}
    q_1 & q_2 & q_3 & q_4
\end{bmatrix}^{\top}, r= \begin{bmatrix}
    r_1 & r_2 & r_3 & r_4
\end{bmatrix}^{\top} \in \mathbb{R}^4$ be two quaternions where the first elements $q_1$ and $r_1$ are the scalar parts, while the other three elements in each quaternion are the vector parts. Then, the left multiplication of $r$ by $q$ according to quaternion algebra~\cite{quat_dyn} is $q \otimes r = L_q r$, where
\begin{equation}
L_q = \begin{bmatrix}
    q_1 & -q_2  &-q_3 & -q_4 \\
    q_2 & q_1 & -q_4  & q_3 \\ 
    q_3 & q_4 & q_1 & -q_2 \\
    q_4 & -q_3 & q_2 & q_1 
\end{bmatrix}
    \label{quat_mult}
\end{equation}
is the skew-symmetric matrix obtained from~$q$. From~\eqref{quat_mult}, quaternion multiplication $\otimes$ is a linear operation, but non-commutative, i.e., $q \otimes r \neq r \otimes q$.

The state differential equation for full pose parameterized by the Neural ODE model is
\begin{equation}
\frac{d \hat{x}(t)}{dt} = \begin{bmatrix} \frac{d \hat{p}(t)}{dt} \\ \frac{d \hat{q}(t)}{dt}\end{bmatrix} =  
\begin{bmatrix}
f_p(\hat{x}(t)) \\ 0.5\left(\begin{bmatrix} 0 \\ f_q(\hat{x}(t))\end{bmatrix} \otimes \hat{q}(t)\right)
\end{bmatrix},
    \label{NODE_SE3}
\end{equation}
where hat $\hat{(\cdot)}$ explicitly represents the states predicted by the Neural ODE~\eqref{NODE_SE3}. The function~$f_p : \mathbb{R}^3 \times \mathcal{S}^3 \to \mathbb{R}^3$ predicts the translational velocity~$\frac{d \hat{p}(t)}{dt}$, while $f_q : \mathbb{R}^3 \times \mathcal{S}^3 \to \mathbb{R}^3$ predicts the angular velocity. Note that angular velocity is not the same as the rate of change of quaternions~$\frac{d \hat{q}(t)}{dt}$. Given the demonstrations of full pose trajectories, we solve the empirical risk minimization problem~\eqref{nODE_opt} to learn the parameters $\theta$ of the neural network~$f_{\theta}(\hat{x}(t)) = \begin{bmatrix} f_p(\hat{x}(t)), & f_q(\hat{x}(t))\end{bmatrix}^{\top}$ where $f_{\theta}: \mathbb{R}^3 \times \mathcal{S}^3 \to \mathbb{R}^6$.

\textbf{Remark}: We use quaternions since they offer a compact, non-singular representation of orientation that avoids gimbal lock and allows for faster computation compared to angle-based representations or rotation matrices in $\mathcal{SO}(3)$. Moreover, the quaternion trajectory~$\hat{q}(t)$ remains a unit quaternion throughout the integration process, provided the initial condition is a unit quaternion and the numerical integration is sufficiently accurate.

\subsubsection{Virtual Control for Stable Motions}

We build on Section~\ref{sec:virtual_control} to generate stable full pose motions as
\begin{equation}
\dot{x} = \begin{bmatrix}
    f_p(x) + u_p(x) \\
0.5\left(\begin{bmatrix} 0 \\ f_q(x) 
+ u_q(x)\end{bmatrix} \otimes q\right)
\end{bmatrix},
    \label{virtual_SE3}
\end{equation}
where the additive \textit{virtual control} term for linear velocity is $u_p(x) \in \mathbb{R}^3$, and that for angular velocity is $u_q(x)\in \mathbb{R}^3$. In our framework, we ensure convergence of the trajectory~$x(t)$ generated by~\eqref{virtual_SE3} to a target trajectory~${x^*(t) = \begin{bmatrix} p^*(t), & q^*(t)\end{bmatrix}^{\top} \in \mathbb{R}^3 \times \mathcal{S}^3}$ by integrating Neural ODEs for complex motions and CLFs defined on the error between the current state and target. We begin by formulating the error dynamics as a control affine dynamical system~\eqref{nonlinear_DS}, where the control inputs are the additive linear and angular velocity terms. Subsequently, we employ CLFs, as introduced in Section~\ref{sec:stability}, to compute the \textit{virtual control} input by solving a Quadratic Program~(QP).

\textbf{Error Dynamics:}
Given a target trajectory~$x^* = \begin{bmatrix}
    p^*, & q^*
\end{bmatrix}^{\top}$, we define the error in position as $e_p = p-p^*$ and the error in quaternion as $e_q = q - q^*$, where we eliminate the explicit dependence on time~$t$ for brevity. Although there are multiple error functions to quantify the distance between quaternions~\cite{merge_pos_rot}, the simple euclidean distance metric performed well in practice. The explicit control affine form of the full pose error dynamics is given below. 
\begin{proposition}
    Given a full pose target trajectory~$x^*$, the error dynamics is a nonlinear control affine dynamical system
    \begin{equation}
    \dot{e} = \begin{bmatrix}
    \dot{e}_p \\ \dot{e}_q
\end{bmatrix} = \textbf{g} + \textbf{h}u(x),
\label{err_dyn_prop_aff_eqn}
\end{equation}
where the error is between the full pose current and target state:
    \begin{equation}
    e = x - x^* = \begin{bmatrix}
    e_p \\ e_q
\end{bmatrix} = \begin{bmatrix}
    p - p^* \\ q - q^*
\end{bmatrix},
\label{err_state}
\end{equation}
the control input is $u(x) = \begin{bmatrix} u_p(x), & u_q(x)\end{bmatrix}^{\top}$, and
\begin{equation}
\textbf{g} = \begin{bmatrix}
    f_p(x)  - \dot{p}^* \\
0.5\left(\begin{bmatrix} 0 \\ f_q(x) 
\end{bmatrix} \otimes q\right) - \dot{q}^*
\end{bmatrix} ,
        \label{err_dyn_prop_g}
    \end{equation}
    \begin{equation}
    \textbf{h} = \begin{bmatrix}
        \textbf{I}_3 & \textbf{0}_{3 \times 3} \\
        \textbf{0}_{4 \times 3} & 0.5\begin{bmatrix}1 & \textbf{0}_3^{\top} \\ \textbf{0}_3 & -\textbf{I}_3\end{bmatrix}L_{\overline{q}}\begin{bmatrix}\textbf{0}_3^{\top} \\ -\textbf{I}_3\end{bmatrix}
    \end{bmatrix},
        \label{err_dyn_prop_h}
    \end{equation}
    where $\overline{q}$ is the conjugate of the quaternion~$q$ with $L_{\overline{q}}$ computed using~\eqref{quat_mult}, $\textbf{I}_n$ is an identity matrix of size~$n$, $\textbf{0}_{m \times n}$ is a zero matrix of size $m \times n$, and $\textbf{0}_n$ is an n-vector of all zeros. 
    \label{err_dyn_prop}
\end{proposition}

\begin{proof}
From~\eqref{virtual_SE3} and defining the error in position as ${e_p = p-p^*}$ and the error in quaternion as $e_q = q - q^*$, the error dynamics is given by
\begin{equation}
\dot{e} = \begin{bmatrix}
    \dot{e}_p \\ \dot{e_q}
\end{bmatrix} = 
\begin{bmatrix}
    f_p(x) + u_p(x)  - \dot{p}^* \\
0.5\left(\begin{bmatrix} 0 \\ f_q(x) 
+ u_q(x)\end{bmatrix} \otimes q\right) - \dot{q}^*
\end{bmatrix}.
    \label{err_dynamics_SE3}
\end{equation}
Since quaternion multiplication is linear~\eqref{quat_mult}, we can rewrite~\eqref{err_dynamics_SE3} as follows.
\begin{equation}
\dot{e} = \begin{bmatrix}
    f_p(x)  - \dot{p}^* \\
0.5\left(\begin{bmatrix} 0 \\ f_q(x) 
\end{bmatrix} \otimes q\right) - \dot{q}^*
\end{bmatrix} + \begin{bmatrix}
        u_p(x) \\
          0.5\left(\begin{bmatrix} 0 \\ u_q(x) 
\end{bmatrix} \otimes q\right)
    \end{bmatrix}.
    \label{err_dynamics_SE3_affine}
\end{equation}
We introduce the quaternion~$y =  0.5\left(\begin{bmatrix} 0 \\ u_q(x) 
\end{bmatrix} \otimes q\right)$ and rewrite~$y$ using quaternion conjugates. 

\textbf{Quaternion conjugate}: Given a quaternion $q = \begin{bmatrix}q_1 & q_2 & q_3 & q_4
\end{bmatrix}^{\top}$, the complex conjugate of $q$ is 

\begin{equation}
\overline{q} = \begin{bmatrix}q_1 & -q_2 & -q_3 & -q_4
\end{bmatrix}^{\top} = \begin{bmatrix} 1 & \textbf{0}_3^{\top} \\ \textbf{0}_3 & -\textbf{I}_3 \end{bmatrix}q.
\label{quat_conj}
\end{equation}

Since taking conjugates reverses the order of quaternion multiplication,
\begin{align*}
    y &= \overline{\overline{y}} = \overline{\overline{0.5\left(\begin{bmatrix} 0 \\ u_q(x) 
\end{bmatrix} \otimes q\right)}} = \overline{0.5\left(\overline{q} \otimes \overline{\begin{bmatrix} 0 \\ u_q(x) 
\end{bmatrix}}\right)}.
\end{align*}
Using~\eqref{quat_conj},
\begin{align*}
y &= \overline{0.5\left(\overline{q} \otimes \begin{bmatrix} 1 & \textbf{0}_3^{\top} \\ \textbf{0}_3 & -\textbf{I}_3 \end{bmatrix}\begin{bmatrix} 0 \\ u_q(x) 
\end{bmatrix}\right)},
\end{align*}
and from~\eqref{quat_mult},
\begin{align*}
y &= \overline{0.5\left(L_{\overline{q}}\begin{bmatrix} 1 & \textbf{0}_3^{\top} \\ \textbf{0}_3 & -\textbf{I}_3 \end{bmatrix}\begin{bmatrix} 0 \\ u_q(x) 
\end{bmatrix}\right)},
\end{align*}
and again using~\eqref{quat_conj},
\begin{align}
y &= 0.5 \begin{bmatrix} 1 & \textbf{0}_3^{\top} \\ \textbf{0}_3 & -\textbf{I}_3 \end{bmatrix}L_{\overline{q}}\begin{bmatrix} 1 & \textbf{0}_3^{\top} \\ \textbf{0}_3 & -\textbf{I}_3 \end{bmatrix}\begin{bmatrix} 0 \\ u_q(x) 
\end{bmatrix}  \nonumber \\
\Rightarrow y  &=  0.5 \begin{bmatrix} 1 & \textbf{0}_3^{\top} \\ \textbf{0}_3 & -\textbf{I}_3 \end{bmatrix}L_{\overline{q}}\begin{bmatrix} \textbf{0}_3^{\top} \\ -\textbf{I}_3 \end{bmatrix}u_q(x).
\label{y_closed}
\end{align}
From~\eqref{y_closed} and~\eqref{err_dynamics_SE3_affine},
\begin{align*}
\dot{e} &= \begin{bmatrix}
    f_p(x)  - \dot{p}^* \\
0.5\left(\begin{bmatrix} 0 \\ f_q(x) 
\end{bmatrix} \otimes q\right) - \dot{q}^*
\end{bmatrix} + \\
&\begin{bmatrix}
        u_p(x) \\
          0.5 \begin{bmatrix} 1 & \textbf{0}_3^{\top} \\ \textbf{0}_3 & -\textbf{I}_3 \end{bmatrix}L_{\overline{q}}\begin{bmatrix} \textbf{0}_3^{\top} \\ -\textbf{I}_3 \end{bmatrix}u_q(x)
    \end{bmatrix}, \\
    \Rightarrow \dot{e} &= \begin{bmatrix}
    f_p(x)  - \dot{p}^* \\
0.5\left(\begin{bmatrix} 0 \\ f_q(x) 
\end{bmatrix} \otimes q\right) - \dot{q}^*
\end{bmatrix} + \\
& \begin{bmatrix}
        \textbf{I}_3 & \textbf{0}_{3 \times 3} \\
        \textbf{0}_{4 \times 3} & 0.5\begin{bmatrix}1 & \textbf{0}_3^{\top} \\ \textbf{0}_3 & -\textbf{I}_3\end{bmatrix}L_{\overline{q}}\begin{bmatrix}\textbf{0}_3^{\top} \\ -\textbf{I}_3\end{bmatrix}
    \end{bmatrix} \begin{bmatrix} u_p(x) \\ u_q(x)\end{bmatrix}.
    \label{e_dot_final}
    \end{align*}
\end{proof}

Based on the control-affine form of the error dynamics derived in Proposition~\ref{err_dyn_prop}, Theorem~\ref{thm:CBF} implies that if there exists a Control Lyapunov Function~(CLF)~$V(\cdot) = -B(\cdot)$ for the full pose error dynamics~\eqref{err_dyn_prop_aff_eqn}, then, any feedback virtual control law $u(\cdot)$ that satisfies
\begin{equation}
\nabla_{e}V(e)^{\top}\left(\textbf{g} + \textbf{h}u(x)\right) \leq -\alpha(V(e)), \ \forall \ e \in \mathbb{R}^7
\label{virtual_control_stable_full_pose}
\end{equation}
will drive the full pose error asymptotically to zero, where~$\alpha(\cdot)$ is a class $\mathcal{K}_{\infty}$ function for CLF. During online motion planning, given the current full pose state of the robot $x$ and information about the full pose target trajectory $x^*$, we compute the minimal control effort $u(x)$ that satisfies~\eqref{virtual_control_stable_full_pose} by setting
\begin{equation}
    \begin{aligned}
        \quad & u(x) = \argmin_{v} \quad  \bigl\|v\bigr\|^2_2  \\  
        \textnormal{s.t.} \quad &
        \nabla_{e}V(e)^{\top}\left(\begin{bmatrix}
    f_p(x)  - \dot{p}^* \\
0.5\left(\begin{bmatrix} 0 \\ f_q(x) 
\end{bmatrix} \otimes q\right) - \dot{q}^*
\end{bmatrix}  + \right. \\
 \quad & \left. \begin{bmatrix}
        \textbf{I}_3 & \textbf{0}_{3 \times 3} \\
        \textbf{0}_{4 \times 3} & 0.5\begin{bmatrix}1 & \textbf{0}_3^{\top} \\ \textbf{0}_3 & -\textbf{I}_3\end{bmatrix}L_{\overline{q}}\begin{bmatrix}\textbf{0}_3^{\top} \\ -\textbf{I}_3\end{bmatrix}
    \end{bmatrix} v \right) \leq -\alpha(V(e)), 
    \end{aligned}
    \label{virtual_control_opt_SE3}
\end{equation}
where $\alpha(\cdot)$ defines how aggressively the robot tracks the target trajectory. The Lyapunov function we use is $V(e) = \|e\|_2^2$, but, note that any positive definite function is valid. Similar to~\eqref{virtual_control_opt}, Problem~\eqref{virtual_control_opt_SE3} is also a Quadratic Program (QP), allowing for efficient computation in motion planning. This formulation stems from the control affine structure of the error dynamics derived in Proposition~\ref{err_dyn_prop}, which itself follows from the linearity of the quaternion multiplication operator~\eqref{quat_mult}. Solving~\eqref{virtual_control_opt_SE3} requires knowledge of the target point~$x^*$, as detailed in the explanation below.

\textbf{Choosing a Target Point:}
We use the same method described in Section~\ref{sec:target_point} to choose the target position $p^*$ and quaternion $q^*$, where the target trajectory~$x^*(t)$ is obtained by integrating the learned full pose Neural ODE model~\eqref{NODE_SE3}. In the closeness measure used to identify the nearest target point $m$ in Algorithm~\ref{alg:target}, we consider only the current position $p$, excluding the quaternion $q$. This is because, in our approach, the position and orientation are coupled through the full-pose Neural ODE model~\eqref{NODE_SE3}, such that the closest position target implicitly determines the appropriate quaternion target. This avoids the need for a separate proximity metric for quaternions, while leveraging the computational simplicity of the Euclidean distance in position space. Since the target point is obtained using the NODE model at each planning step, the target velocities are
\begin{equation}
    \dot{p}^* = f_p(x^*), \ \dot{q}^* = 0.5\left(\begin{bmatrix} 0 \\ f_q(x^*) 
\end{bmatrix} \otimes q^*\right),
\label{target_vel_SE3}
\end{equation}
which are used in solving~\eqref{virtual_control_opt_SE3}.

\begin{figure*}[!b]
\centering
\begin{subfigure}[h]{0.48\textwidth}
    \centering
    \includegraphics[width=\textwidth]{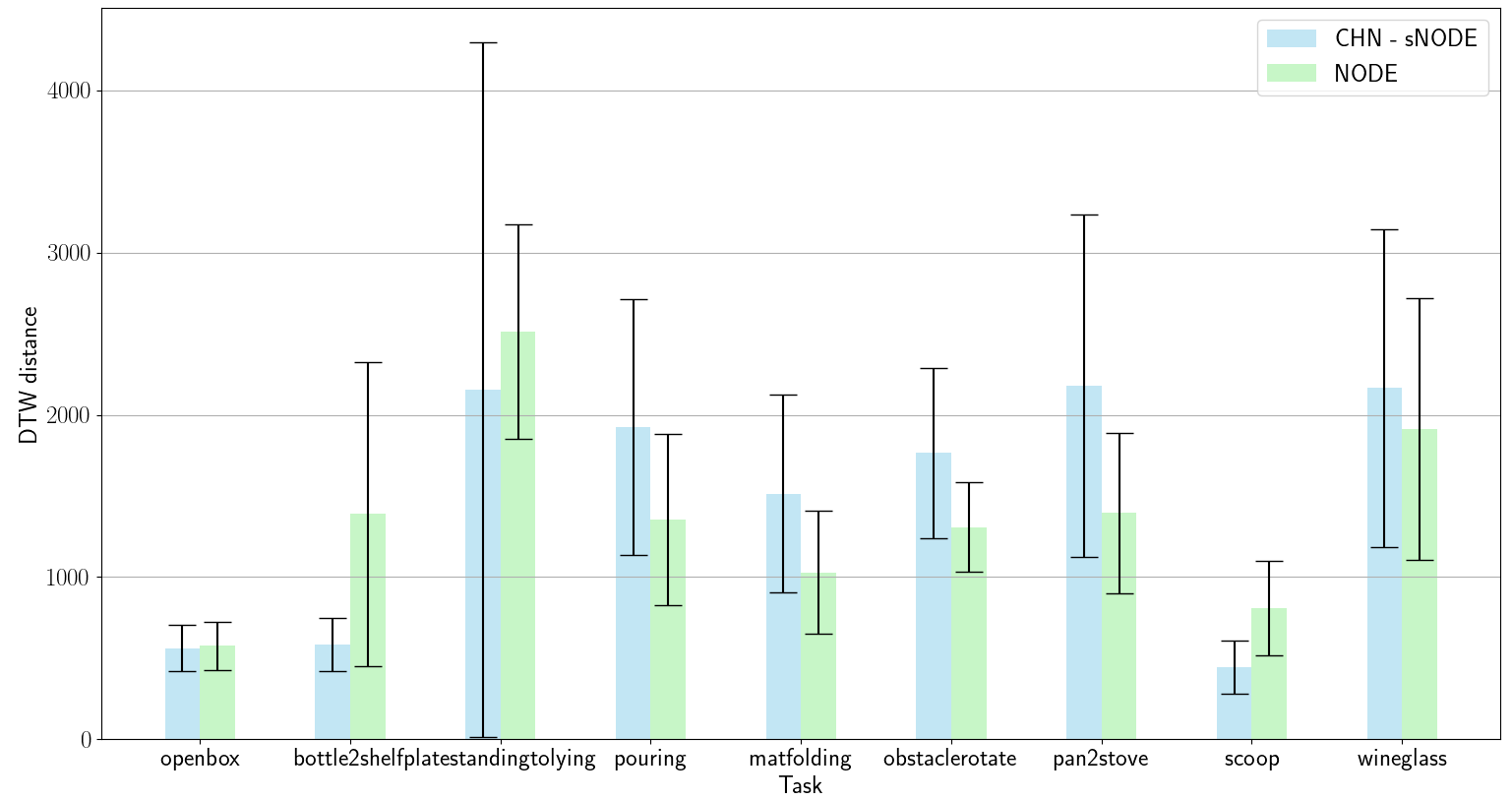}
    \caption{DTW distance}
    \label{fig:dtw_dist_rot}
\end{subfigure}
\hfill
\begin{subfigure}[h]{0.48\textwidth}
    \centering
    \includegraphics[width=\textwidth]{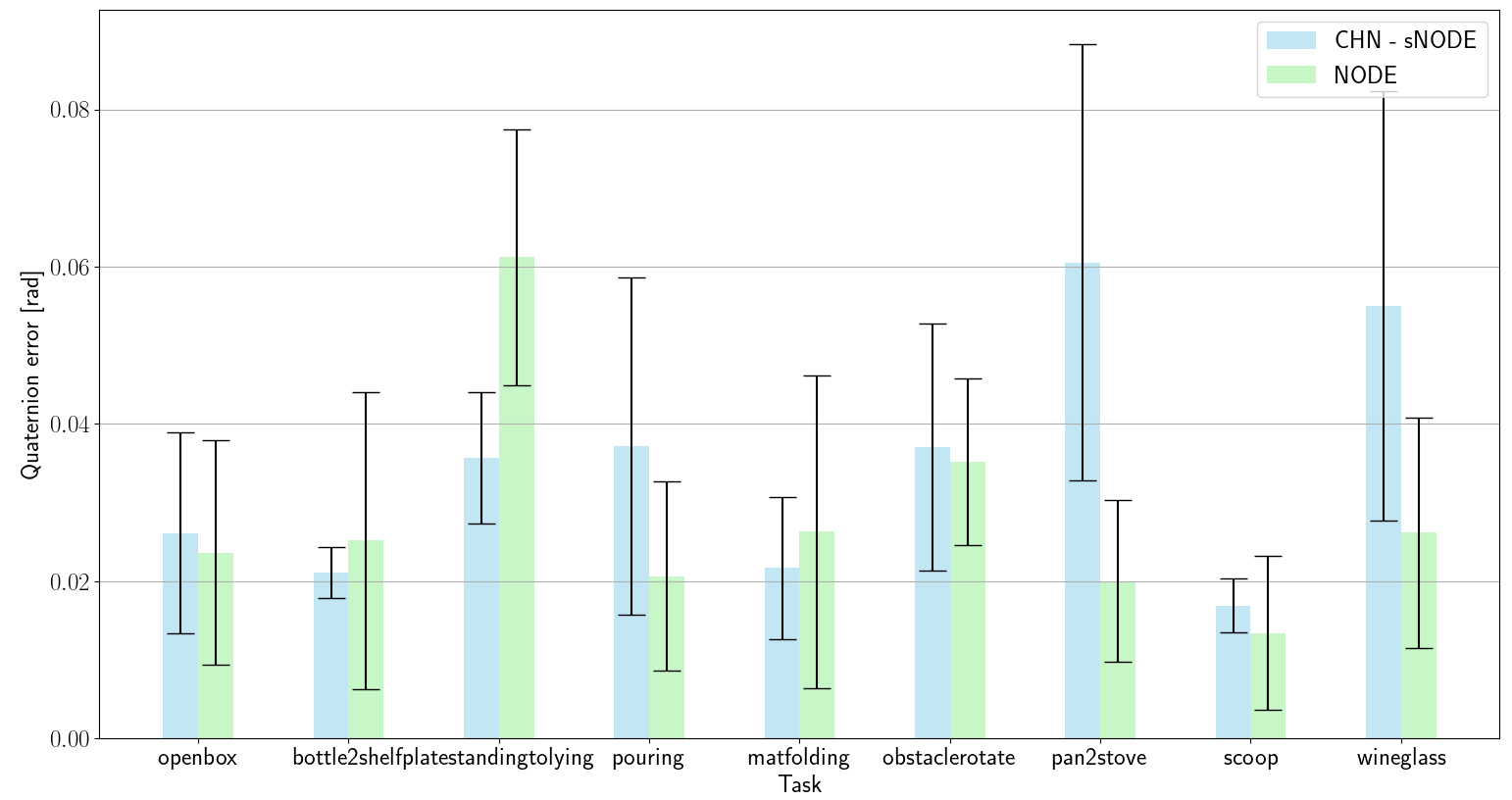}
    \caption{Quaternion error}
    \label{fig:quat_err_rot}
\end{subfigure}

\vspace{1em}

\begin{subfigure}[h]{0.48\textwidth}
    \centering
    \includegraphics[width=\textwidth]{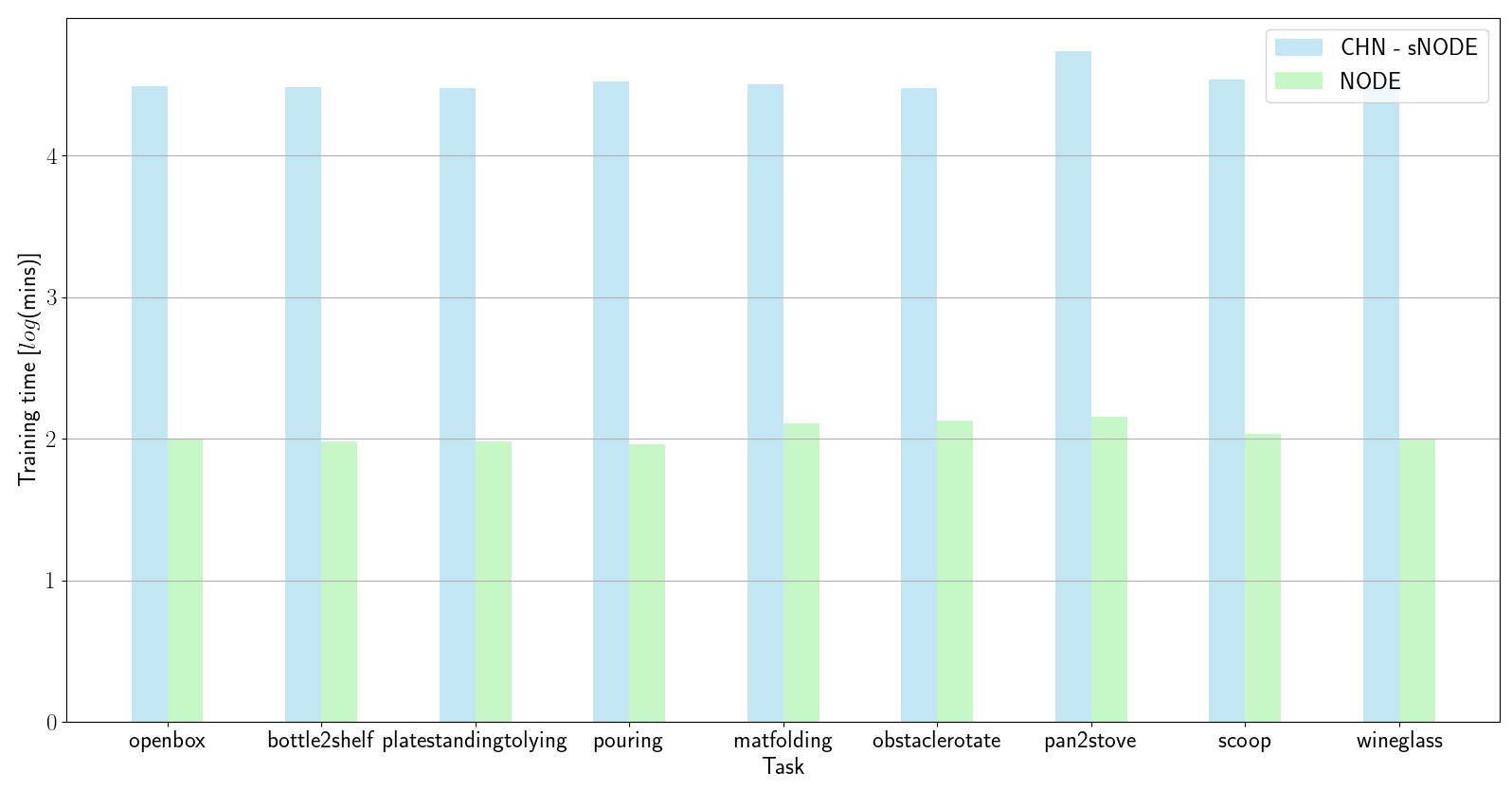}
    \caption{Training time}
    \label{fig:train_time_rot}
\end{subfigure}
\hfill
\begin{subfigure}[h]{0.48\textwidth}
    \centering
    \includegraphics[width=\textwidth]{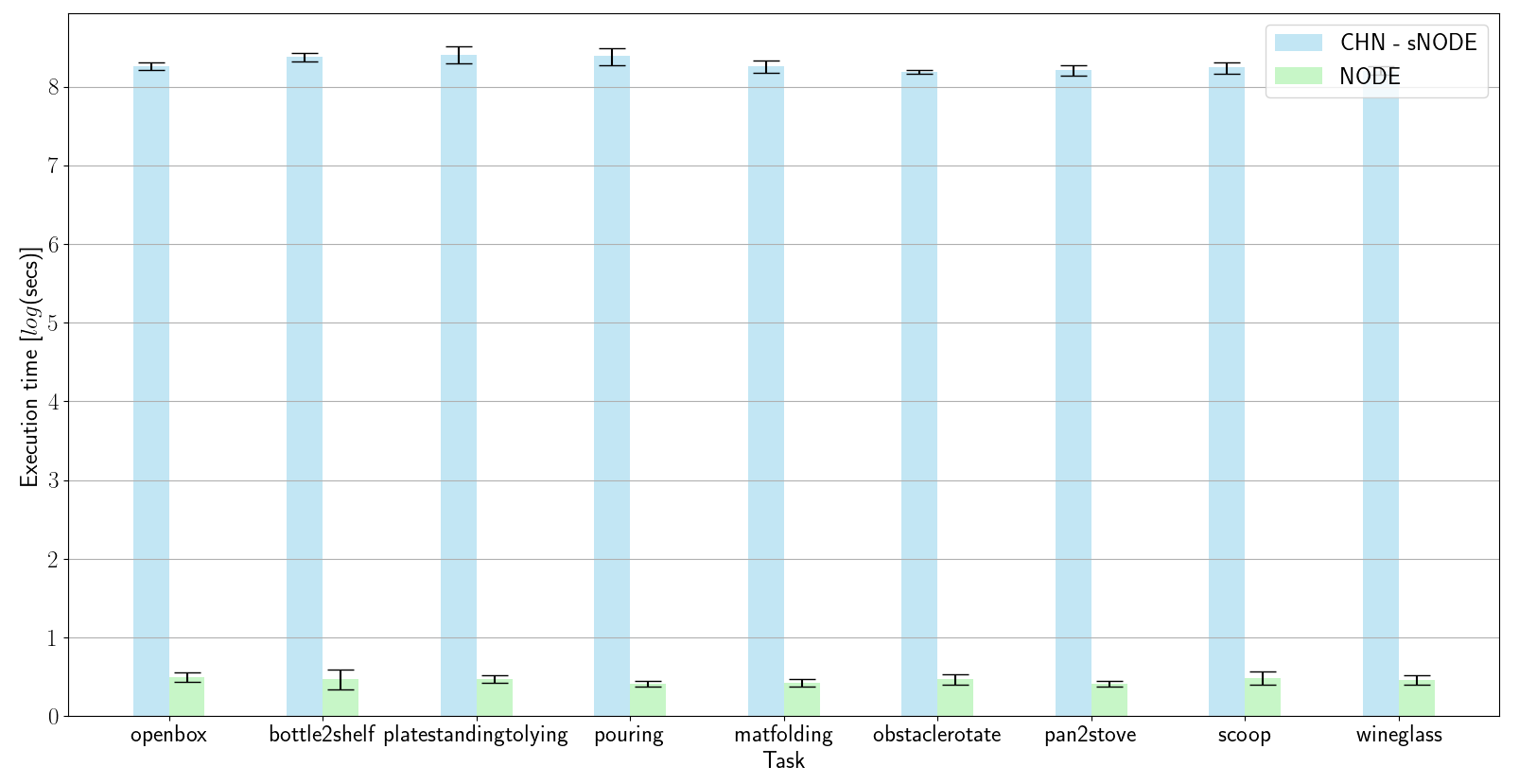}
    \caption{Execution time}
    \label{fig:exec_time_rot}
\end{subfigure}

\caption{Comparison of performance metrics between CHN-sNODE and our approach (NODE) for reproducing full pose trajectories from the real robot dataset in~\cite{clfd_sNODE}.}
\label{fig:rot}
\end{figure*}

\subsubsection{Results}
\label{app:rot_results}

We validate our approach on the real robot dataset presented in~\cite{clfd_sNODE} that contains 9 full pose motions representing different manipulation tasks. In Fig.~\ref{fig:rot}, we compare our approach (NODE) with the CHN-sNODE method proposed in~\cite{clfd_sNODE} across four metrics on the full-pose robotic tasks. NODE achieves comparable DTW distance and Quaternion error, indicating better trajectory alignment in both the position and rotational space. Moreover, NODE significantly reduces training time by over $50\%$ on average in the log scale, and improves execution time by several orders of magnitude, making it far more suitable for real-time applications.


\begin{figure*}[!b]
     \centering
         \begin{subfigure}[b]{0.24\textwidth}
         \centering         \includegraphics[width=\textwidth]{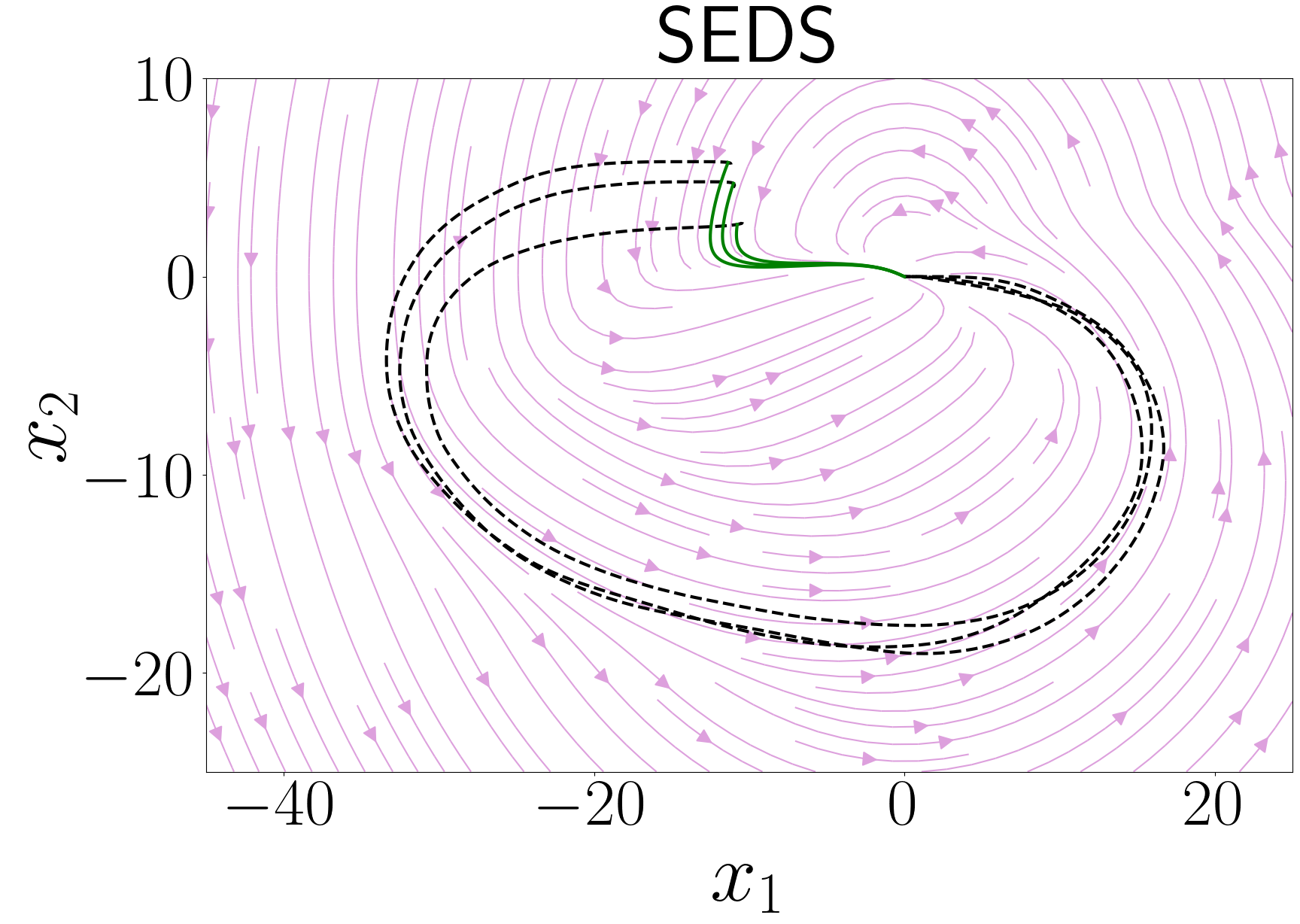}
         \label{fig:DoubleBendedLine_SEDS}
     \end{subfigure}
     \hfill
     \begin{subfigure}[b]{0.24\textwidth}
         \centering         \includegraphics[width=\textwidth]{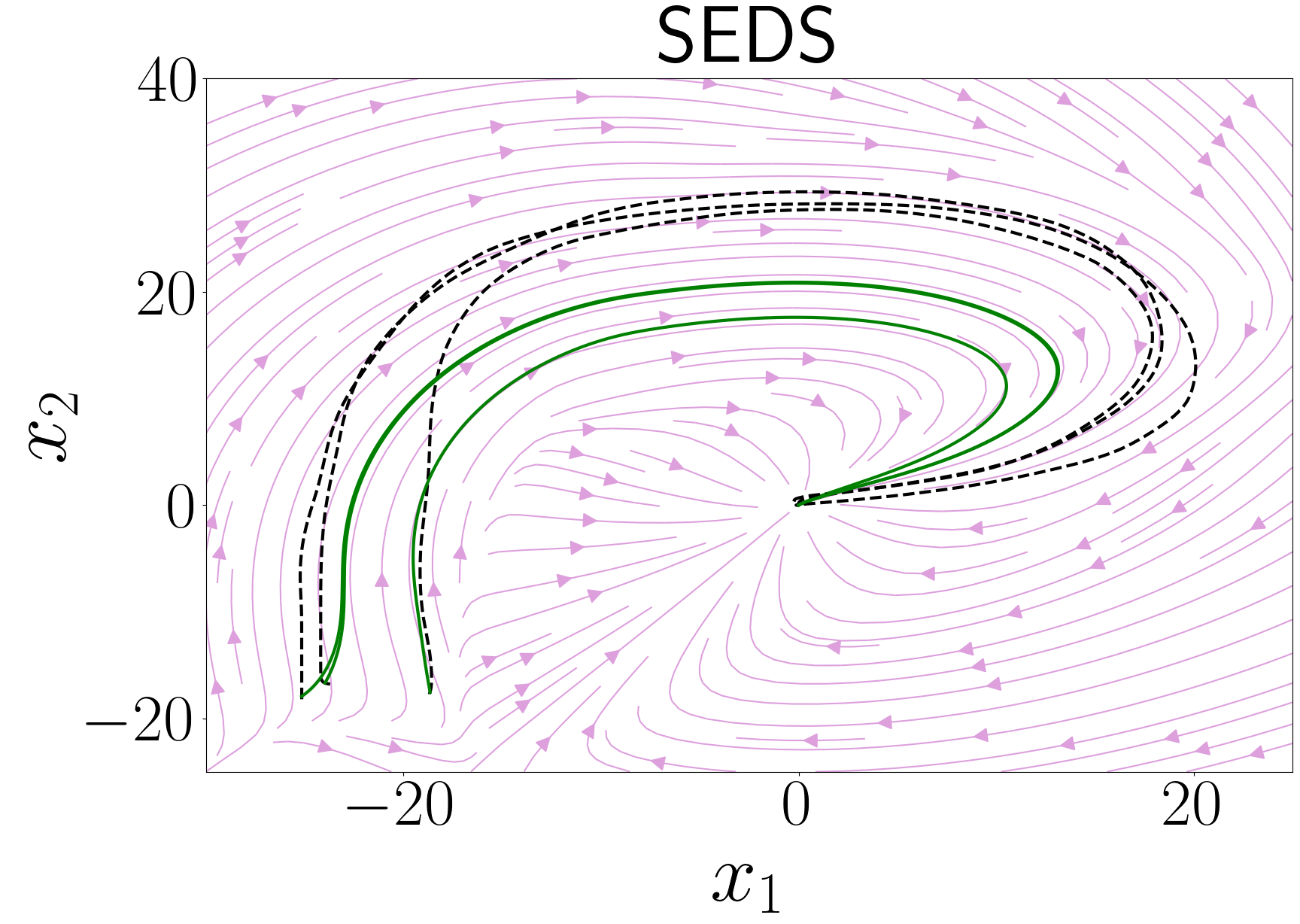}
         \label{fig:PShape_SEDS}
     \end{subfigure}
     \hfill
          \begin{subfigure}[b]{0.24\textwidth}
         \centering         \includegraphics[width=\textwidth]{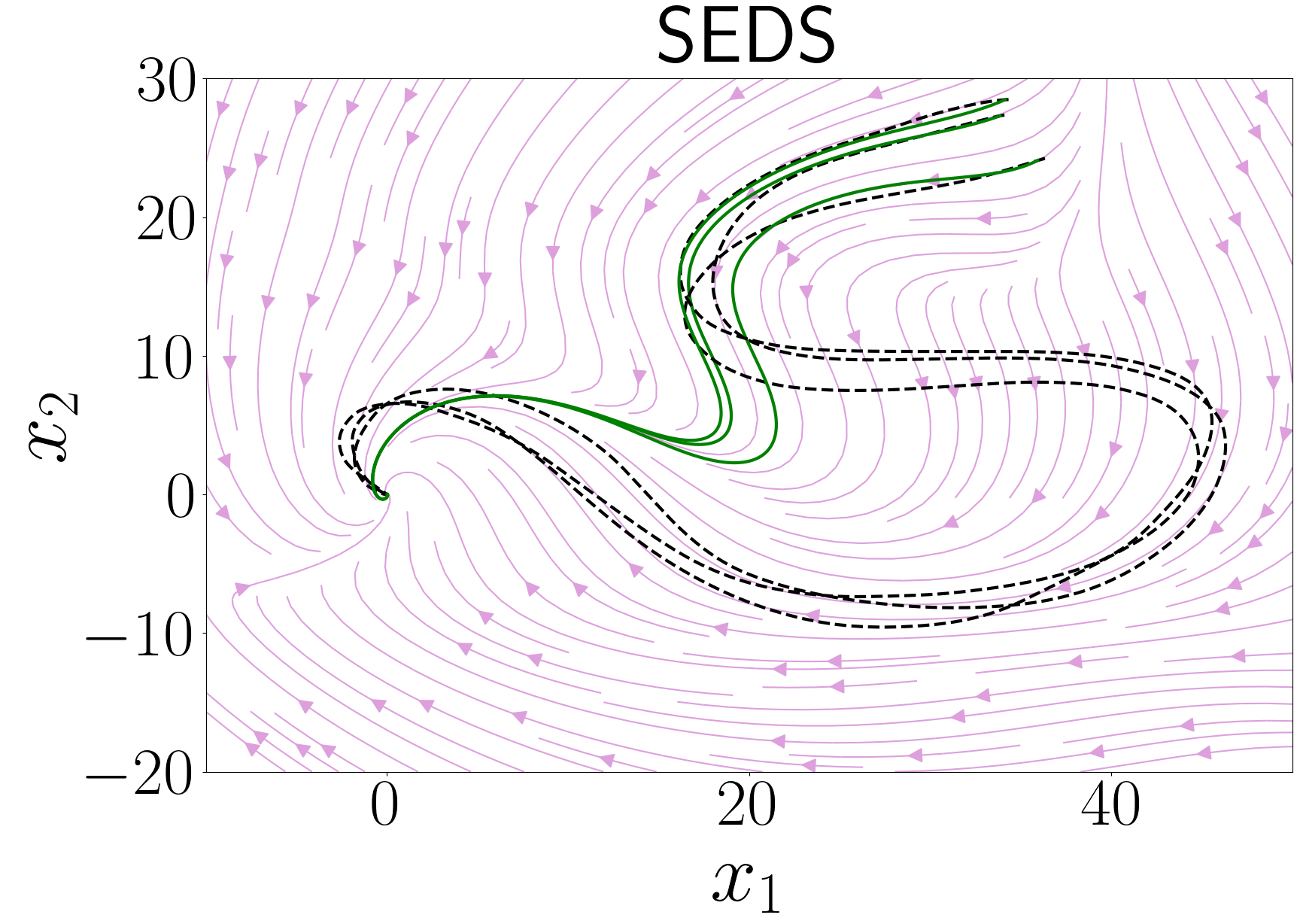}
         \label{fig:Snake_SEDS}
     \end{subfigure}
     \hfill     
     \begin{subfigure}[b]{0.24\textwidth}         \centering        \includegraphics[width=\textwidth]{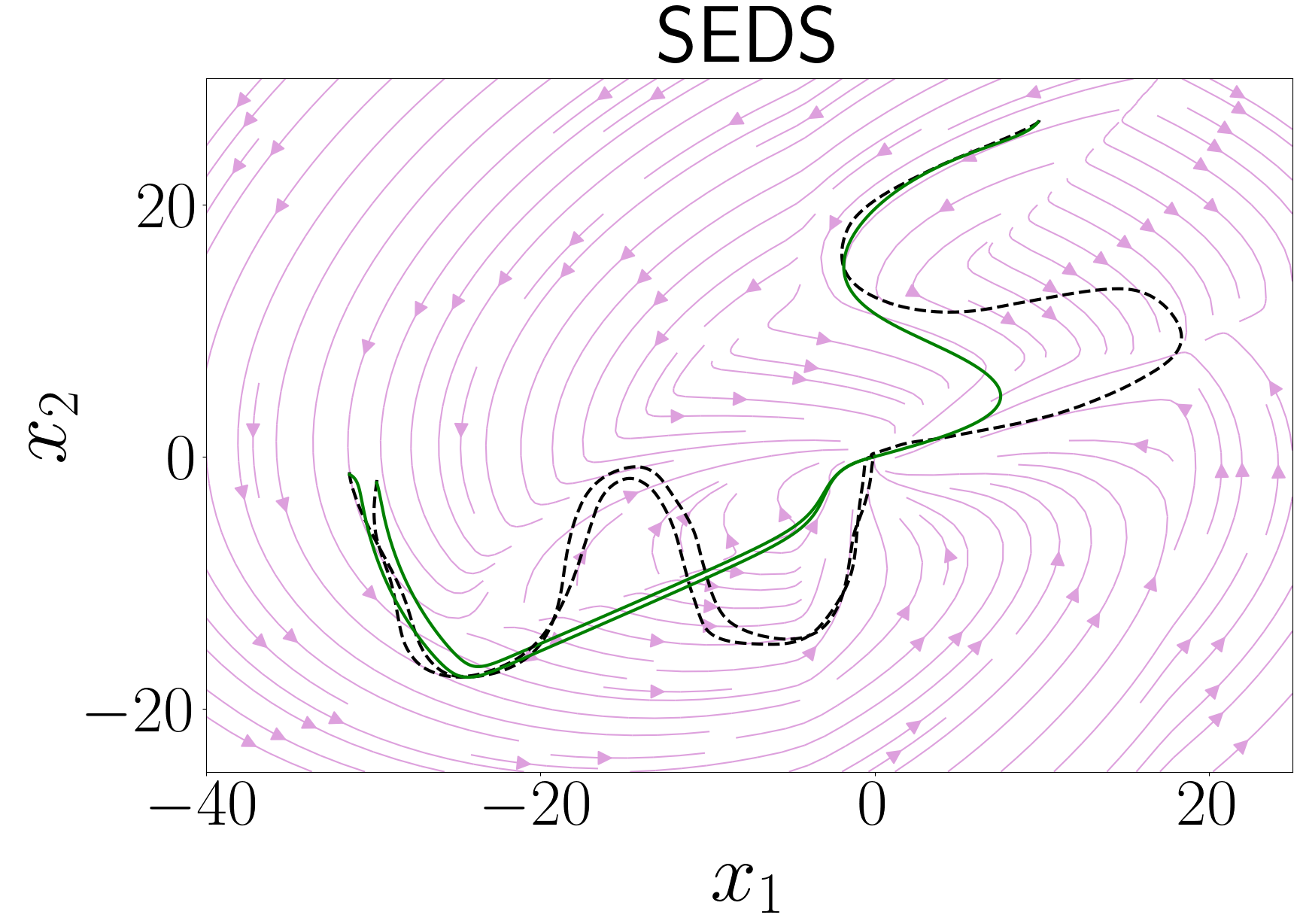}
         \label{fig:MultiModels3_SEDS}
     \end{subfigure}
     \\
          \begin{subfigure}[b]{0.24\textwidth}
         \centering         \includegraphics[width=\textwidth]{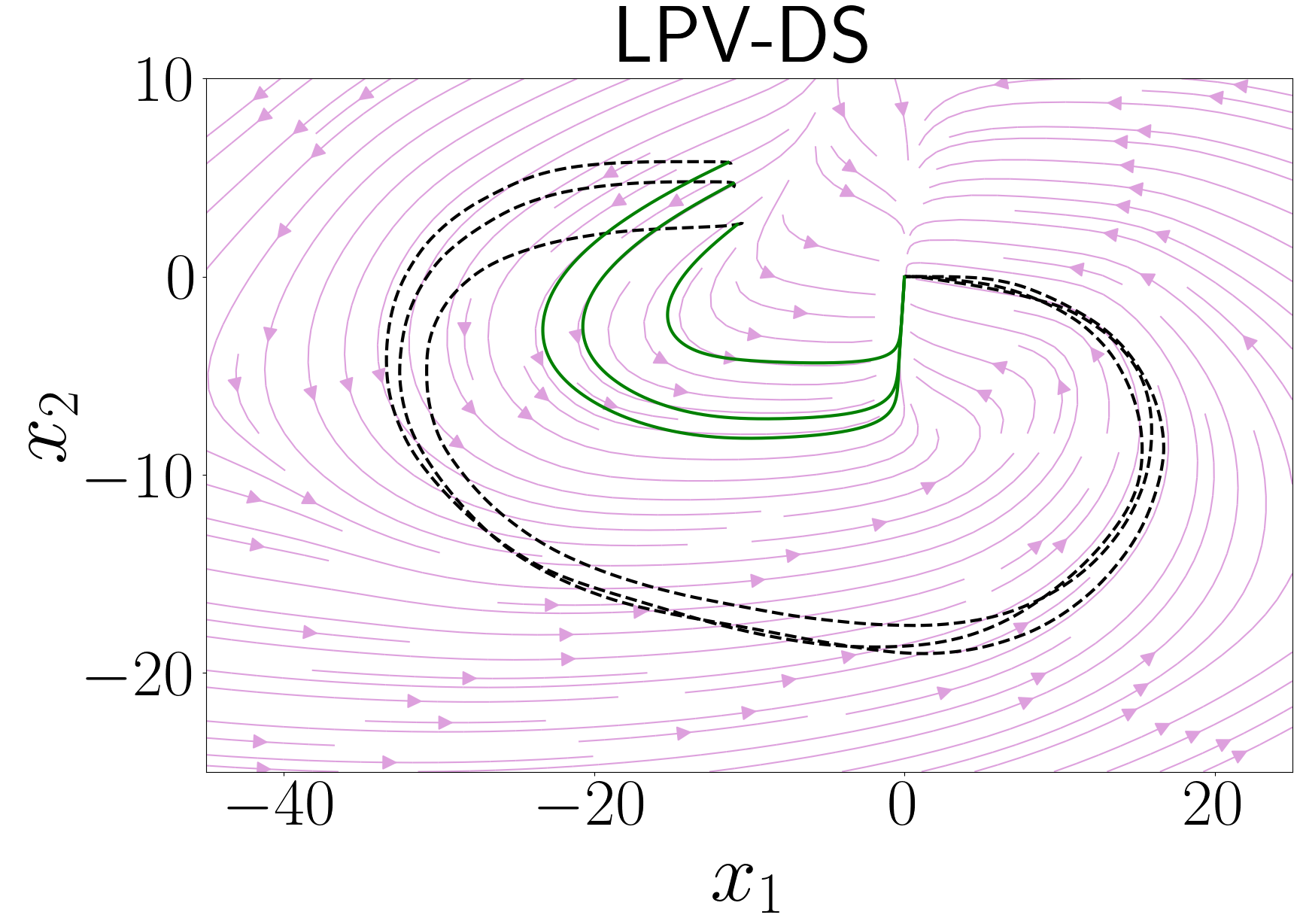}
         \label{fig:DoubleBendedLine_LPVDS}
     \end{subfigure}
     \hfill     
          \begin{subfigure}[b]{0.24\textwidth}
         \centering         \includegraphics[width=\textwidth]{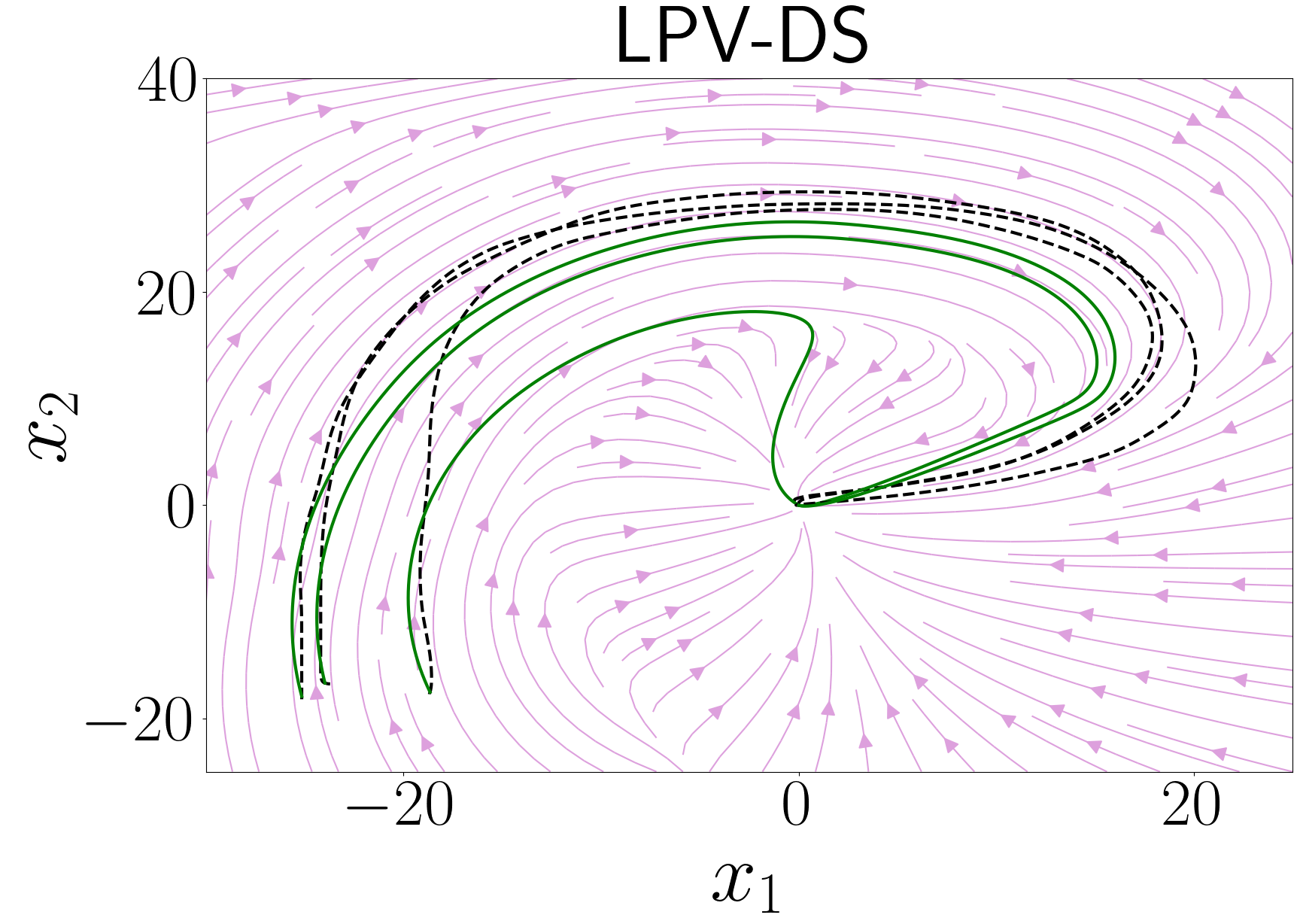}
         \label{fig:PShape_LPVDS}
     \end{subfigure}
     \hfill
     \begin{subfigure}[b]{0.24\textwidth}         \centering        \includegraphics[width=\textwidth]{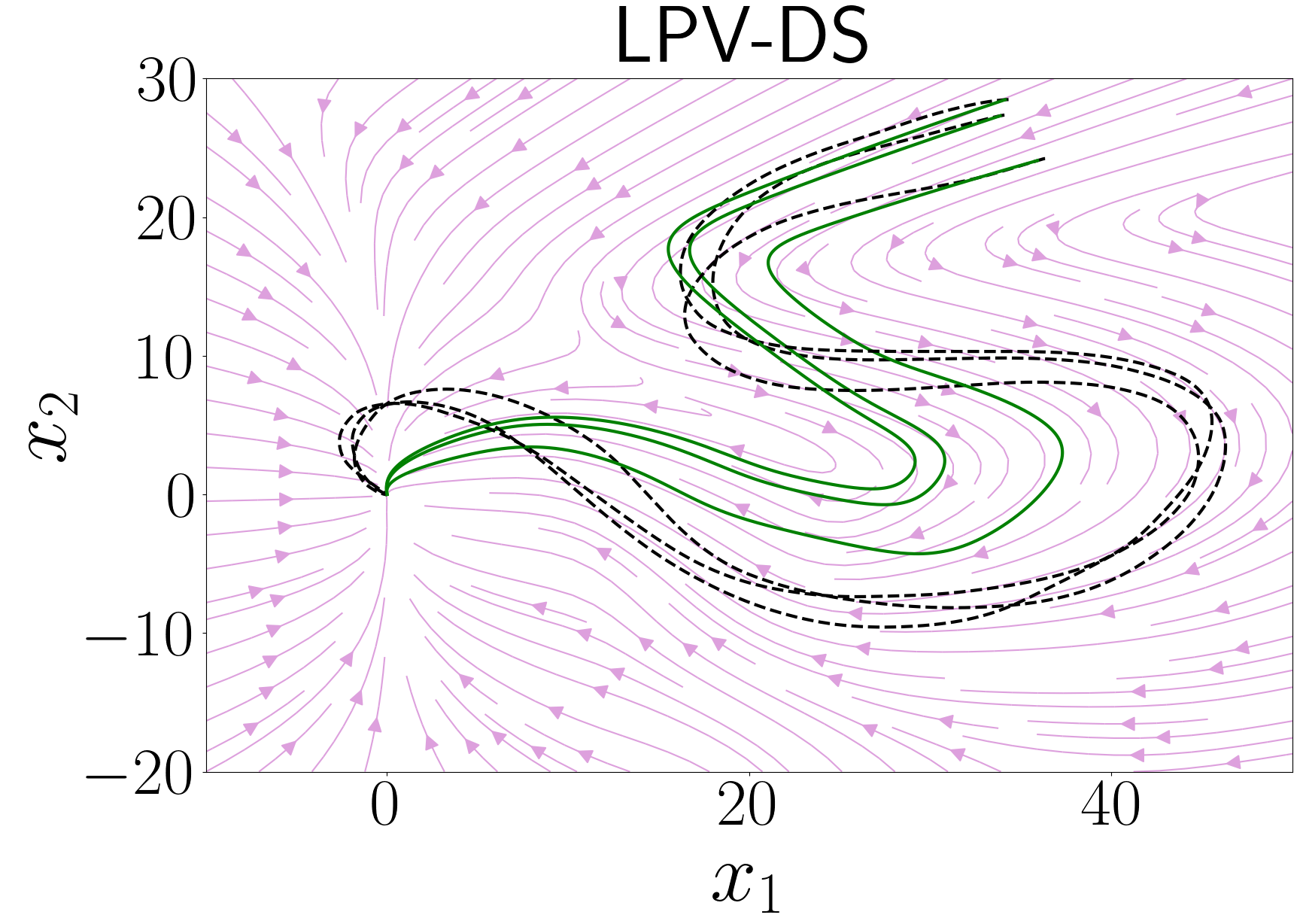}
         \label{fig:Snake_LPVDS}
     \end{subfigure}
     \hfill
          \begin{subfigure}[b]{0.24\textwidth}
         \centering         \includegraphics[width=\textwidth]{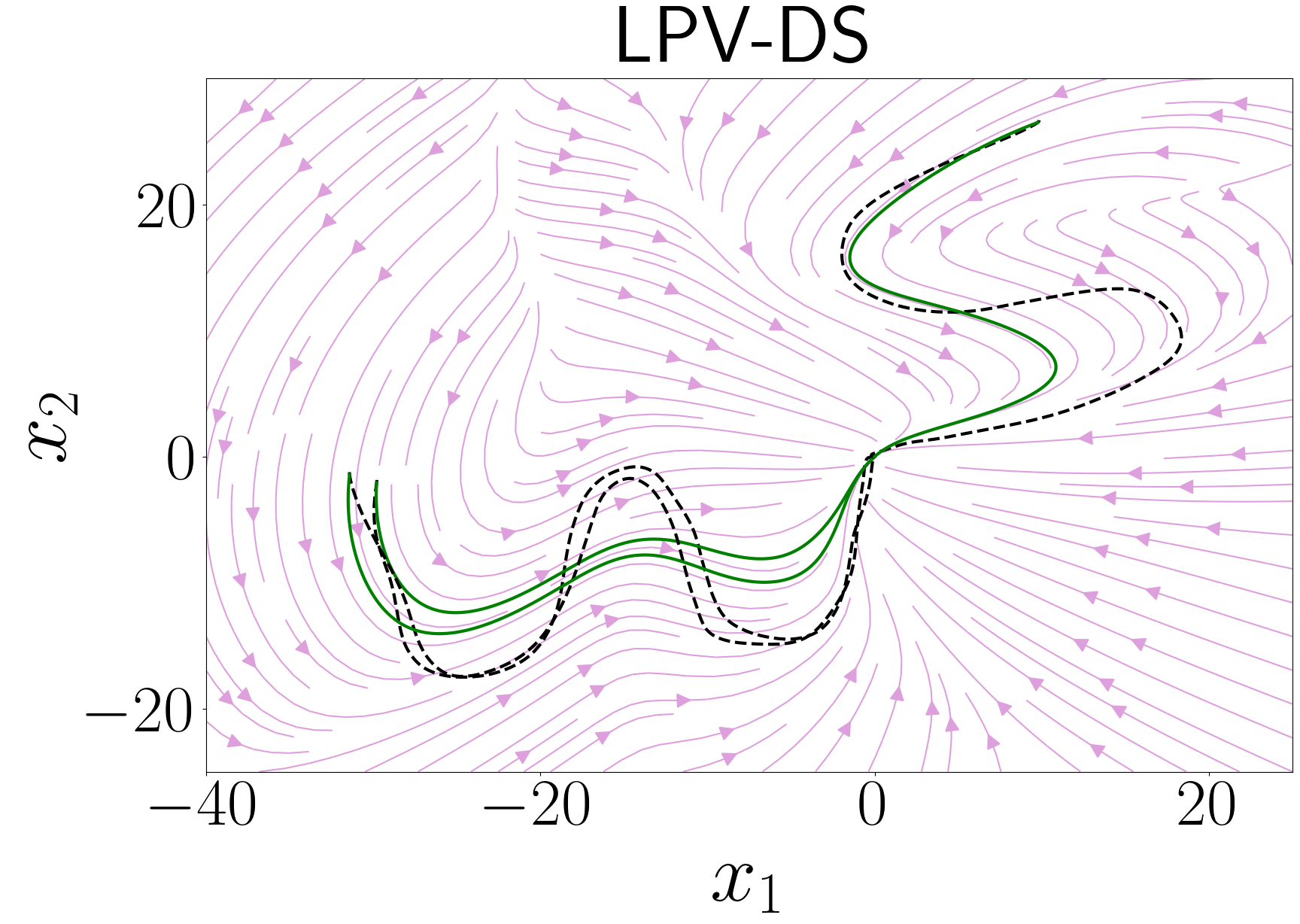}
         \label{fig:MultiModels3_LPVDS}
     \end{subfigure}
     \\
          \begin{subfigure}[b]{0.24\textwidth}
         \centering         \includegraphics[width=\textwidth]{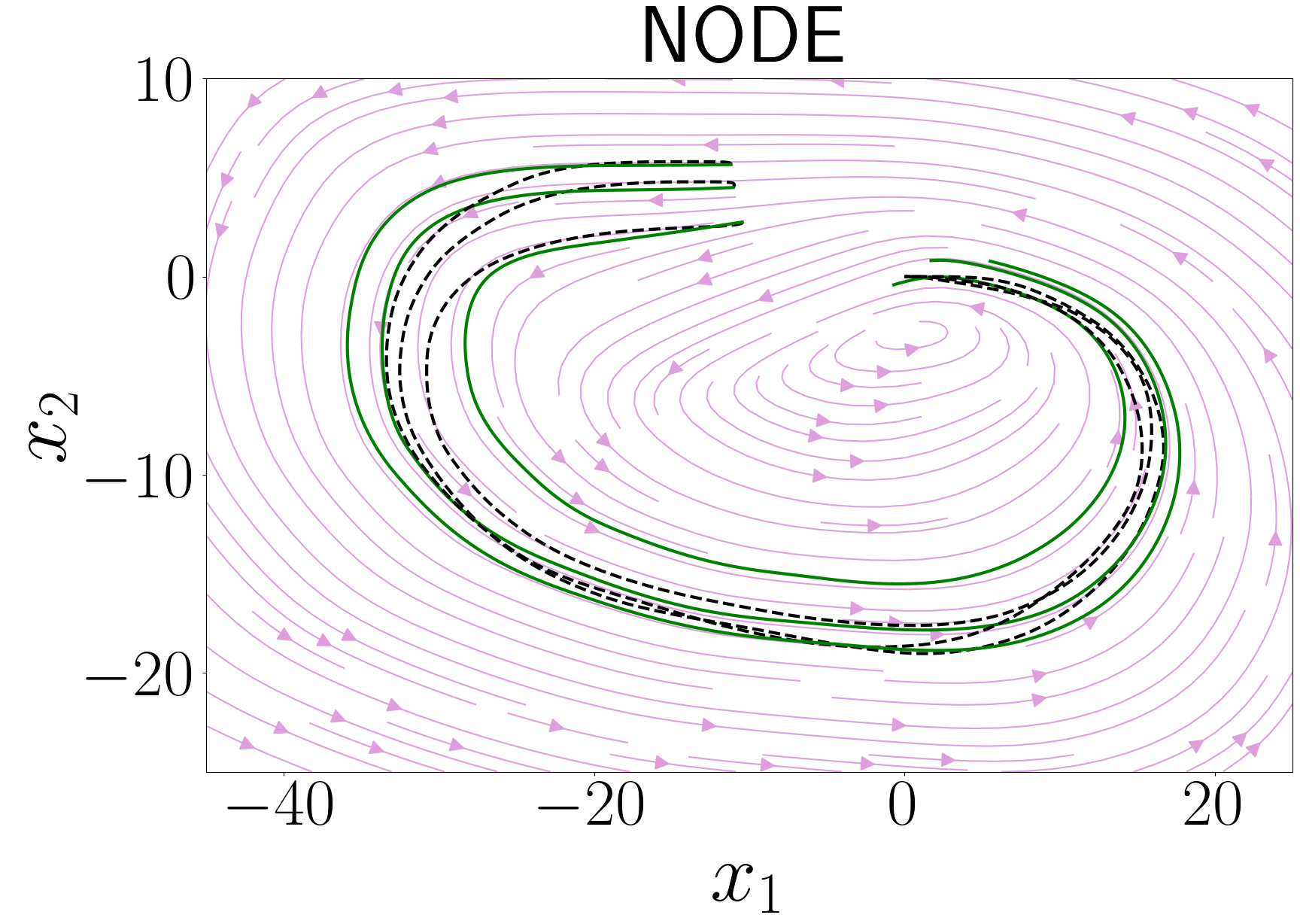}
         \label{fig:DoubleBendedLine_NODE}
     \end{subfigure}
     \hfill     
          \begin{subfigure}[b]{0.24\textwidth}
         \centering         \includegraphics[width=\textwidth]{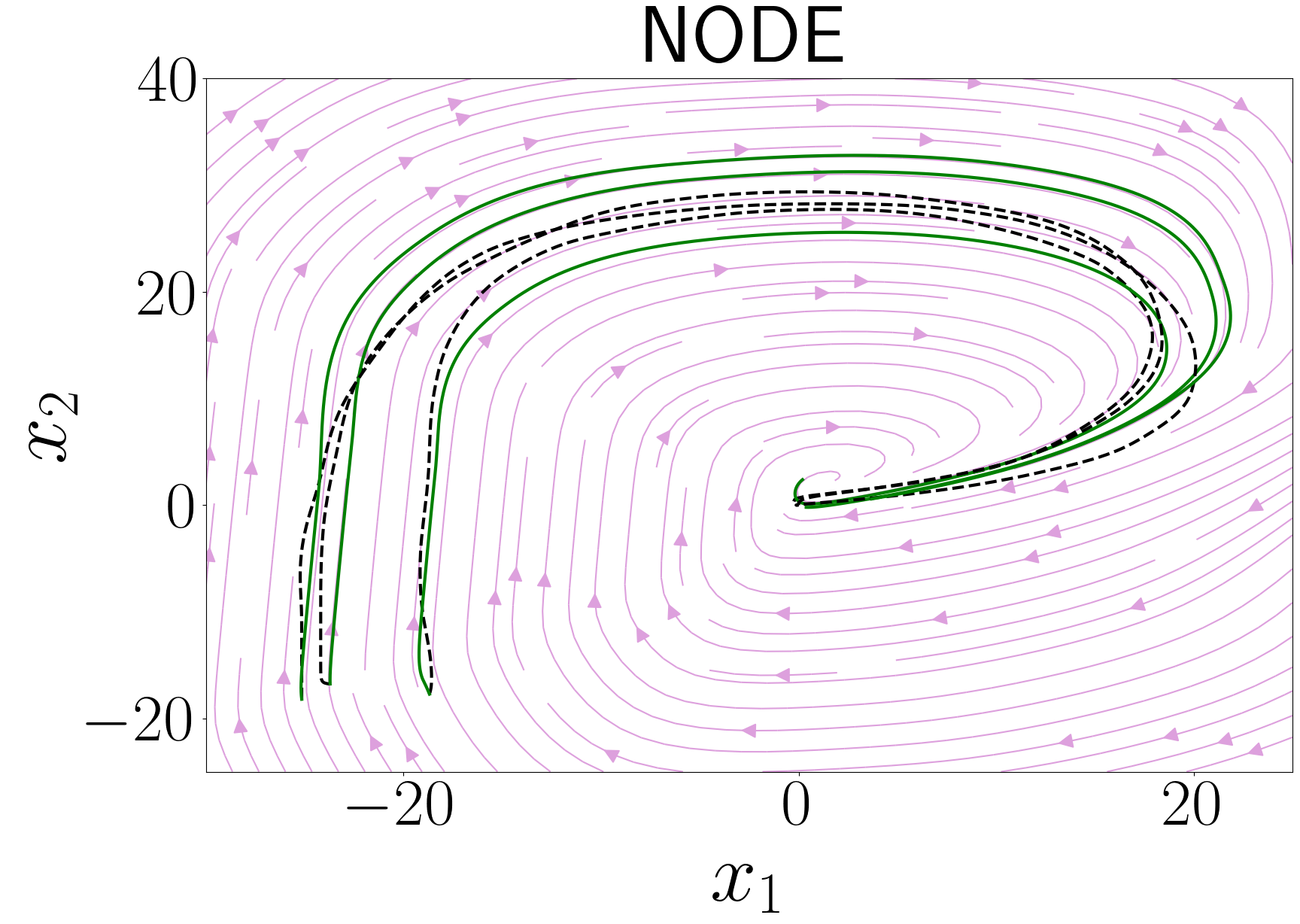}
         \label{fig:PShape_NODE}
     \end{subfigure}
     \hfill
     \begin{subfigure}[b]{0.24\textwidth}         \centering        \includegraphics[width=\textwidth]{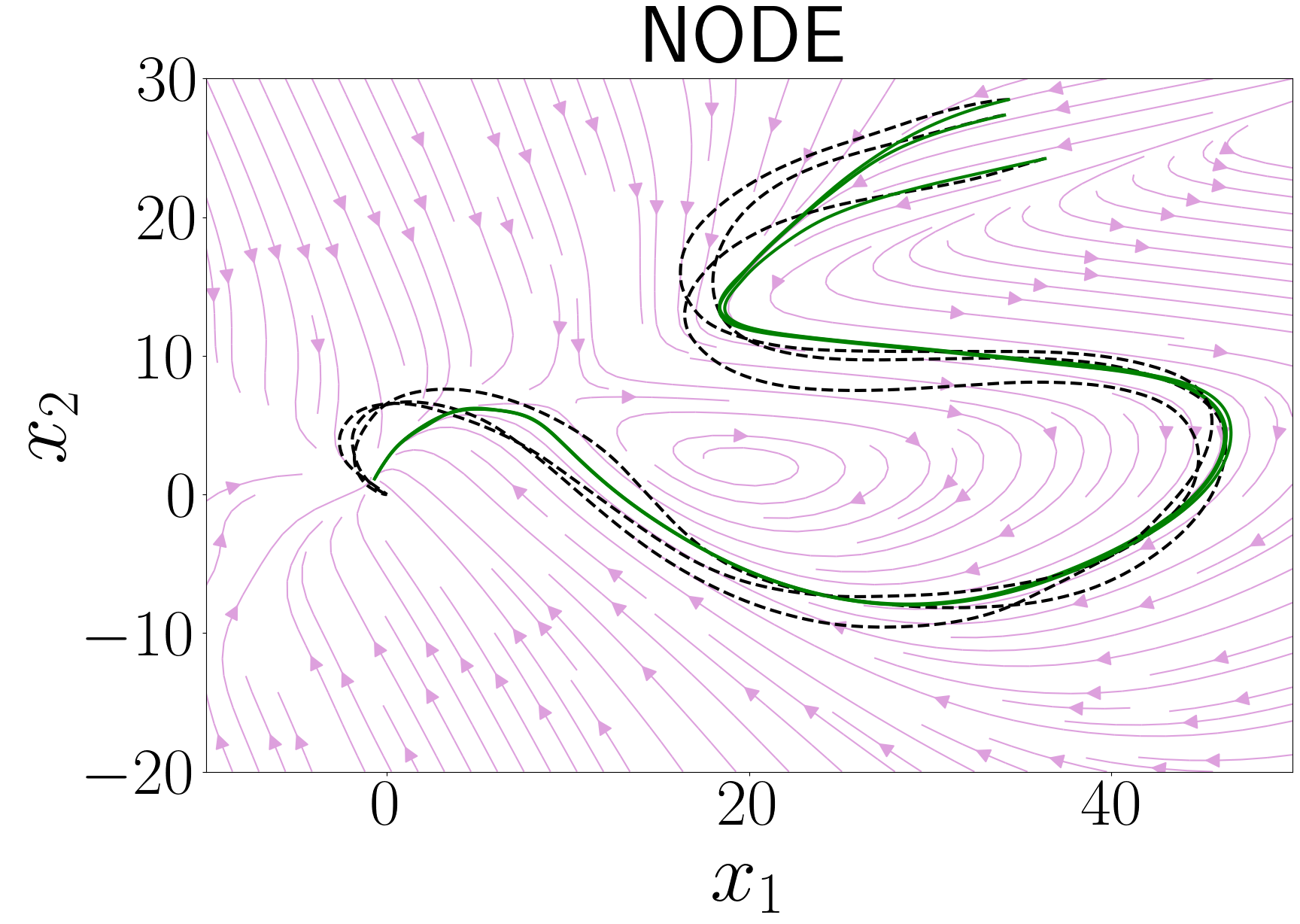}
         \label{fig:Snake_NODE}
     \end{subfigure}
     \hfill
          \begin{subfigure}[b]{0.24\textwidth}
         \centering         \includegraphics[width=\textwidth]{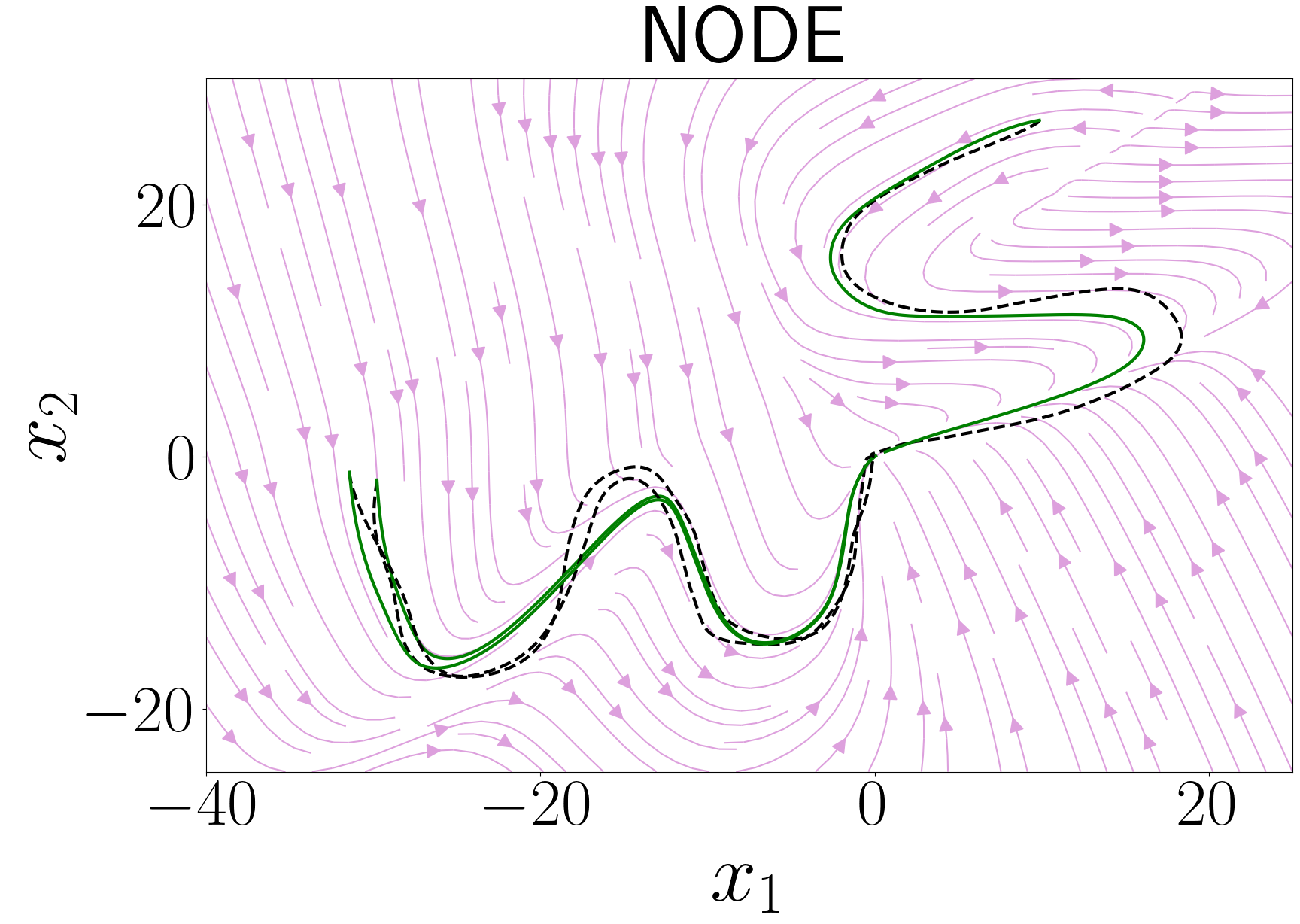}
         \label{fig:MultiModels3_NODE}
     \end{subfigure}
     \\
               \begin{subfigure}[b]{0.24\textwidth}
         \centering         \includegraphics[width=\textwidth]{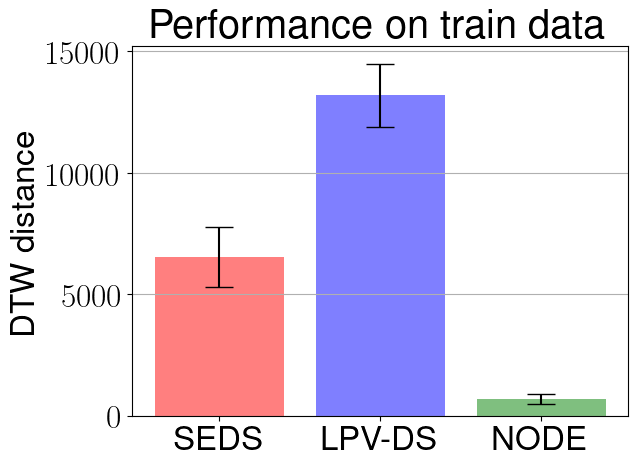}
         \label{fig:DoubleBendedLine_metric_train}
     \end{subfigure}
     \hfill     
          \begin{subfigure}[b]{0.24\textwidth}
         \centering         \includegraphics[width=\textwidth]{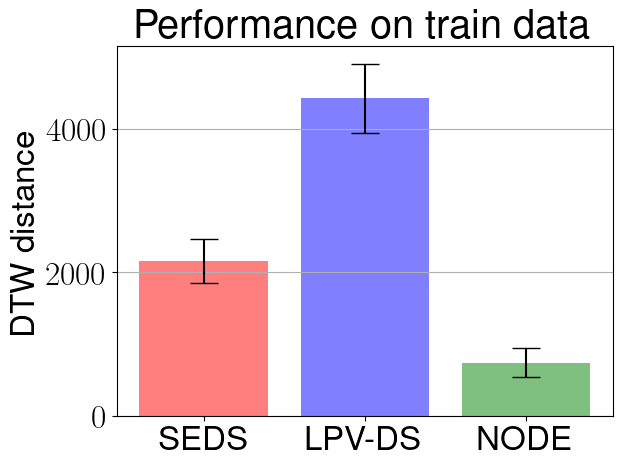}
         \label{fig:PShape_metric_train}
     \end{subfigure}
     \hfill
     \begin{subfigure}[b]{0.24\textwidth}         \centering        \includegraphics[width=\textwidth]{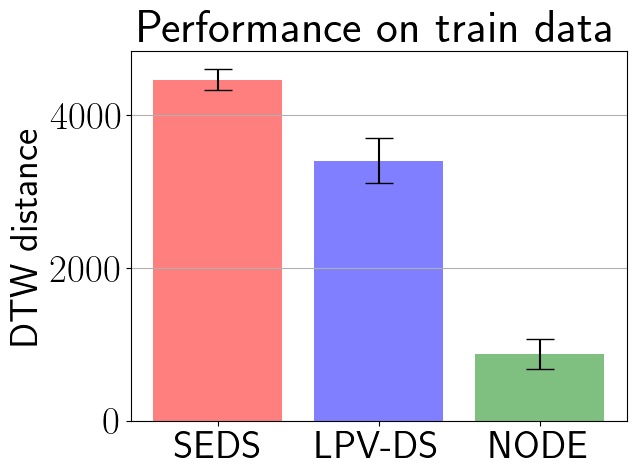}
         \label{fig:Snake_metric_train}
     \end{subfigure}
     \hfill
          \begin{subfigure}[b]{0.24\textwidth}
         \centering         \includegraphics[width=\textwidth]{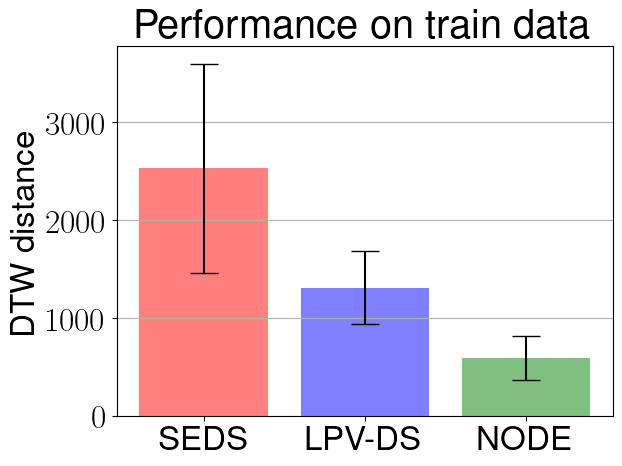}
         \label{fig:MultiModels3_metric_train}
     \end{subfigure}
     \\
                    \begin{subfigure}[b]{0.24\textwidth}
         \centering         \includegraphics[width=\textwidth]{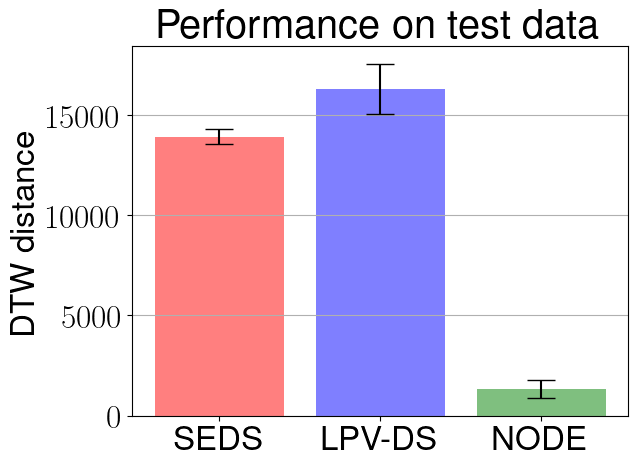}
         \caption{\textit{DoubleBendedLine}}
         \label{fig:DoubleBendedLine_metric_test}
     \end{subfigure}
     \hfill     
          \begin{subfigure}[b]{0.24\textwidth}
         \centering         \includegraphics[width=\textwidth]{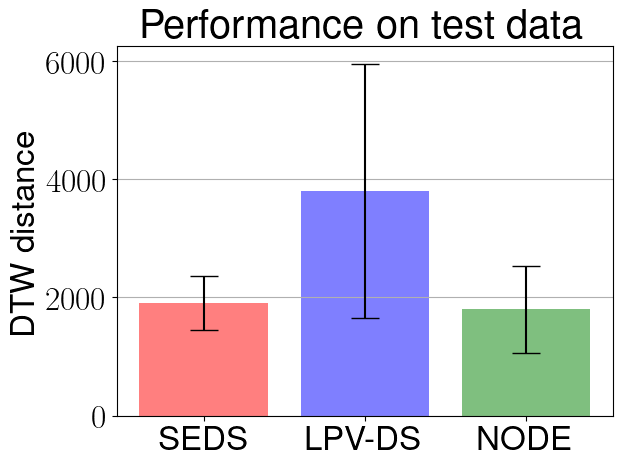}
         \caption{\textit{PShape}}
         \label{fig:PShape_metric_test}
     \end{subfigure}
     \hfill
     \begin{subfigure}[b]{0.24\textwidth}         \centering        \includegraphics[width=\textwidth]{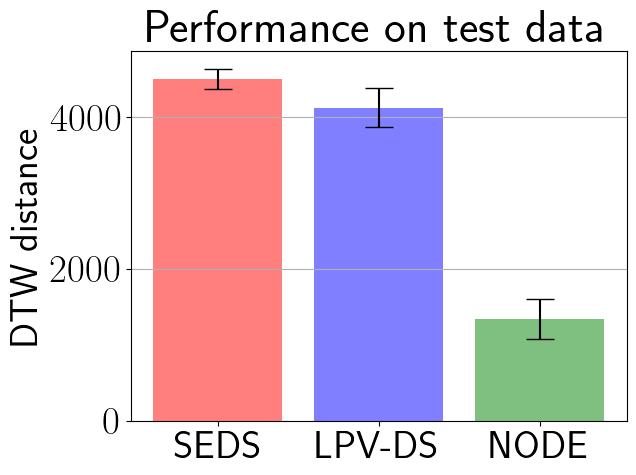}
     \caption{\textit{Snake}}
         \label{fig:Snake_metric_test}
     \end{subfigure}
     \hfill
          \begin{subfigure}[b]{0.24\textwidth}
         \centering         \includegraphics[width=\textwidth]{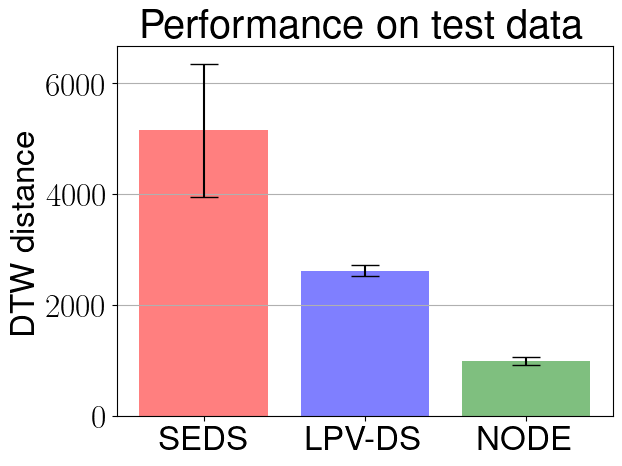}
         \caption{\textit{MultiModels3}}
         \label{fig:MultiModels3_metric_test}
     \end{subfigure}
        \caption{Trajectories predicted from the LASA test data set. The dashed black trajectories are from the data set. The solid green trajectories are predicted by the model. \textbf{1st row:} Prediction from SEDS. \textbf{2nd row:} Predictions from LPV-DS. \textbf{3rd row:} Predictions by our Neural ODE~(NODE) model. \textbf{4th row:} DTW distance comparison on train data. \textbf{5th row:} DTW distance comparison on test data. }
        \label{fig:LASA_comparison}
\end{figure*}

\subsection{Further Validation on Complex Trajectories}
\label{sec:valid_LASA_app}

\begin{figure*}[!htb]
     \centering
         \begin{subfigure}[b]{0.24\textwidth}
         \centering         \includegraphics[width=\textwidth]{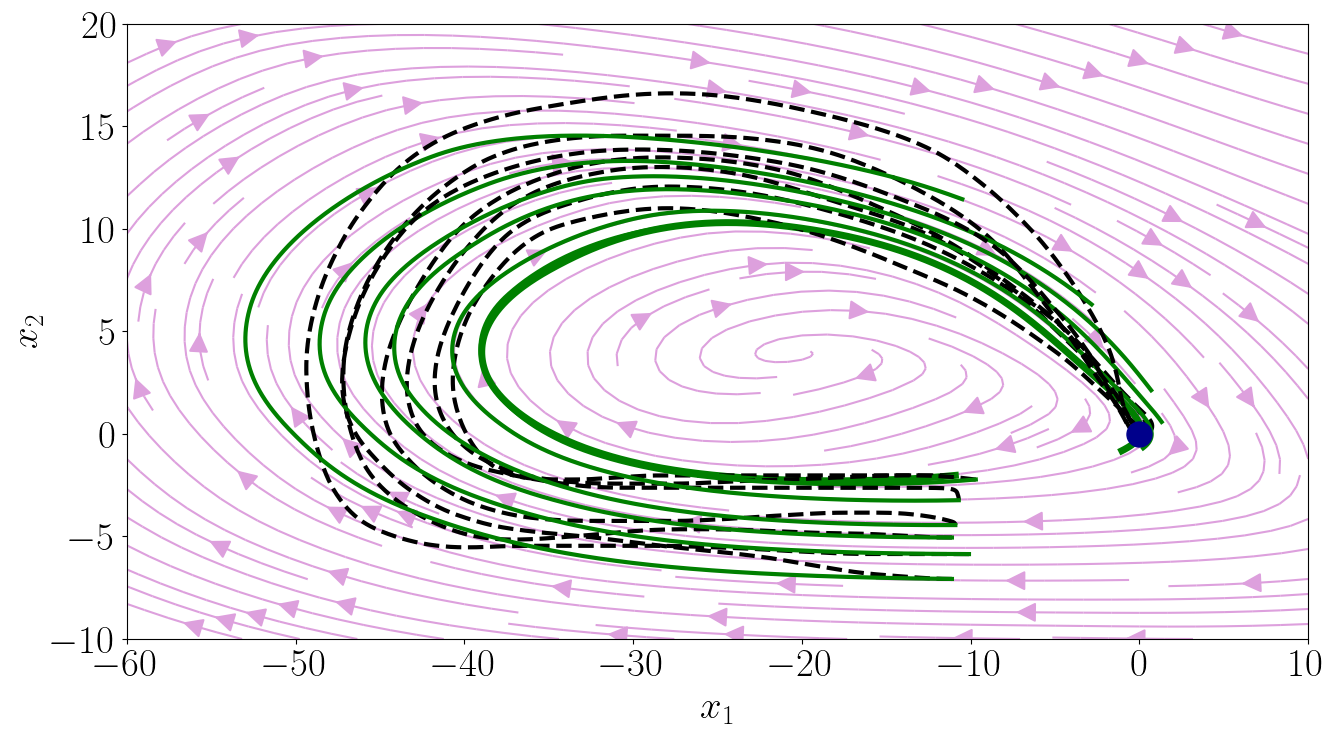}
         \caption{\textit{BendedLine}}
         \label{fig:BendedLine}
     \end{subfigure}
     \hfill
     \begin{subfigure}[b]{0.24\textwidth}
         \centering         \includegraphics[width=\textwidth]{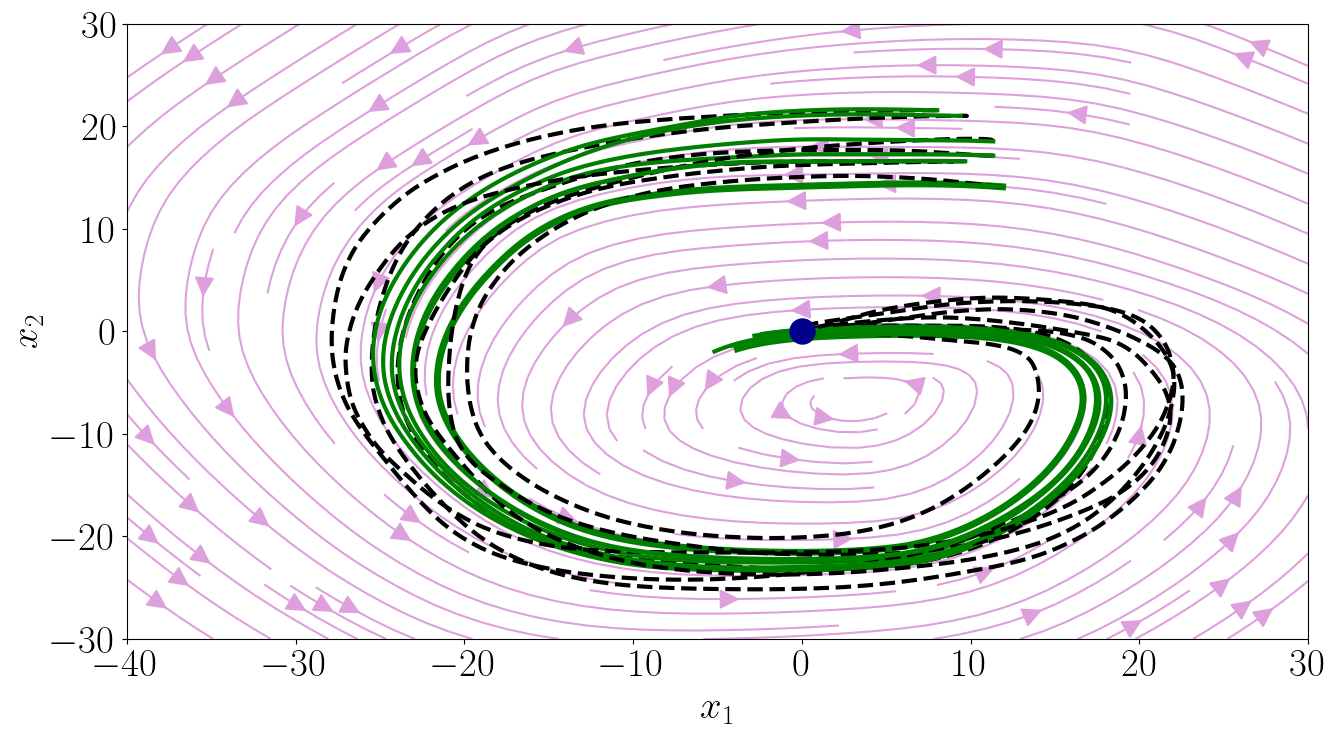}
         \caption{\textit{GShape}}
         \label{fig:GShape}
     \end{subfigure}
     \hfill
          \begin{subfigure}[b]{0.24\textwidth}
         \centering         \includegraphics[width=\textwidth]{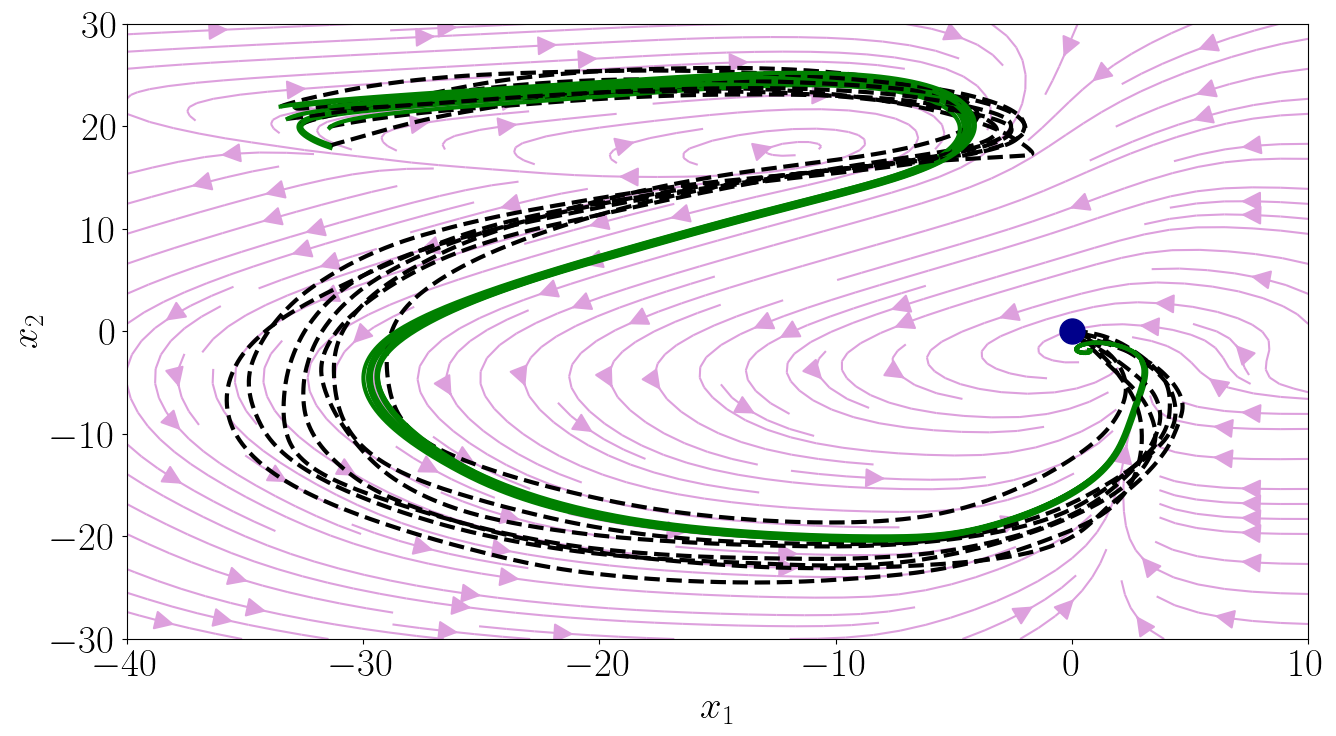}
         \caption{\textit{heee}}
         \label{fig:heee}
     \end{subfigure}
     \hfill     
     \begin{subfigure}[b]{0.24\textwidth}         \centering        \includegraphics[width=\textwidth]{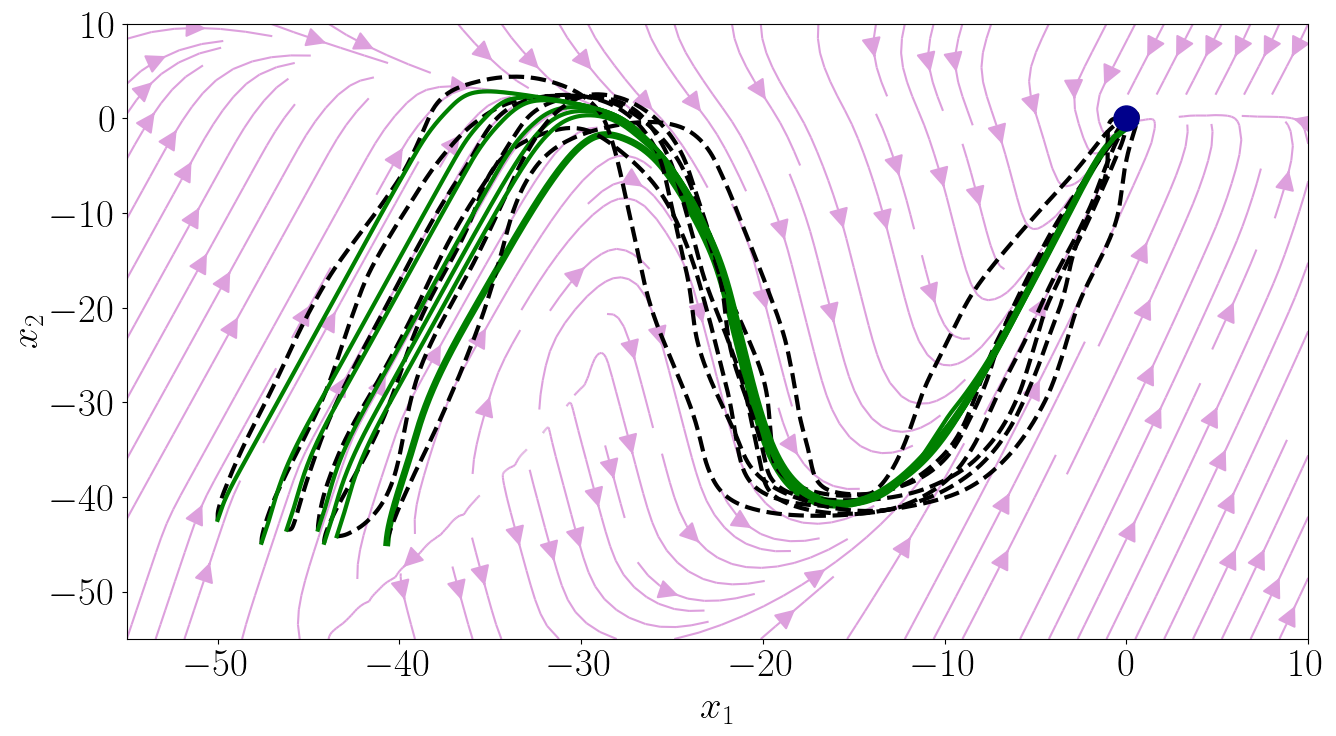}
         \caption{\textit{NShape}}
         \label{fig:NShape}
     \end{subfigure}
     \\
          \begin{subfigure}[b]{0.24\textwidth}
         \centering         \includegraphics[width=\textwidth]{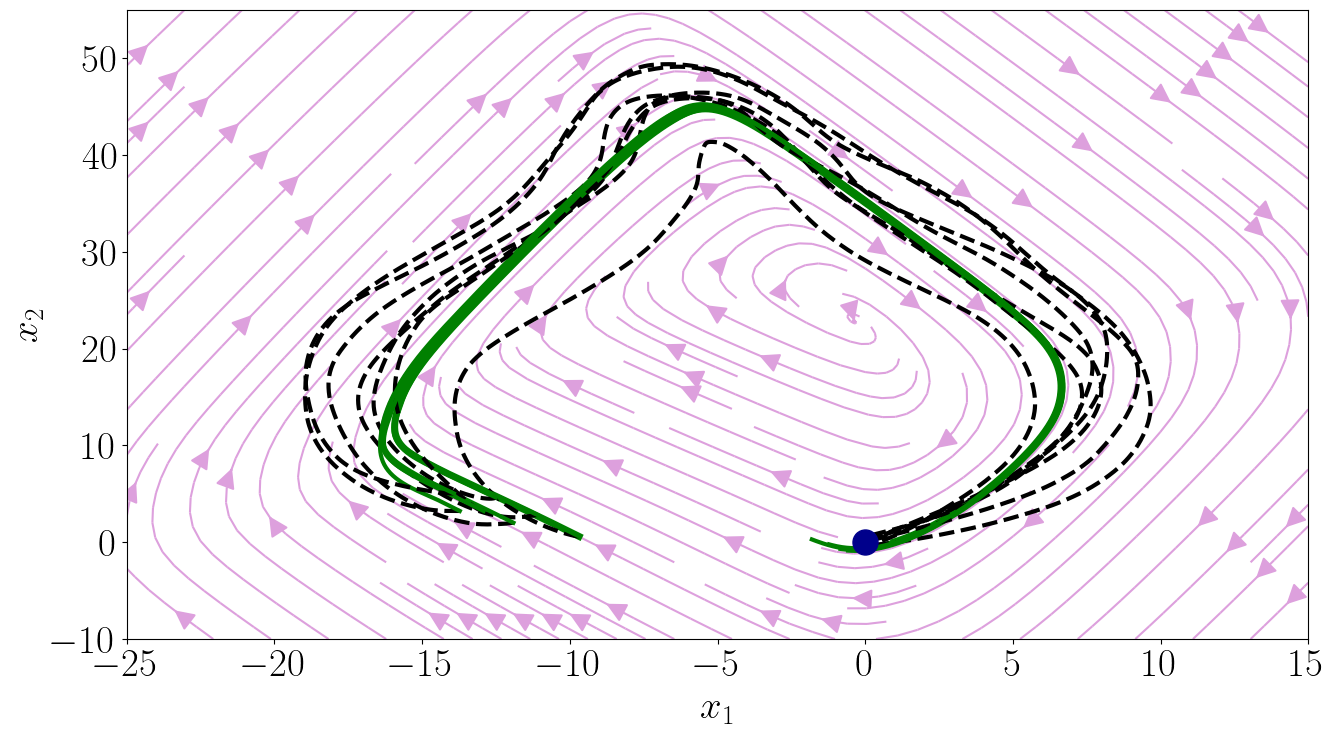}
         \caption{\textit{Leaf-1}}
         \label{fig:Leaf_1}
     \end{subfigure}
     \hfill     
          \begin{subfigure}[b]{0.24\textwidth}
         \centering         \includegraphics[width=\textwidth]{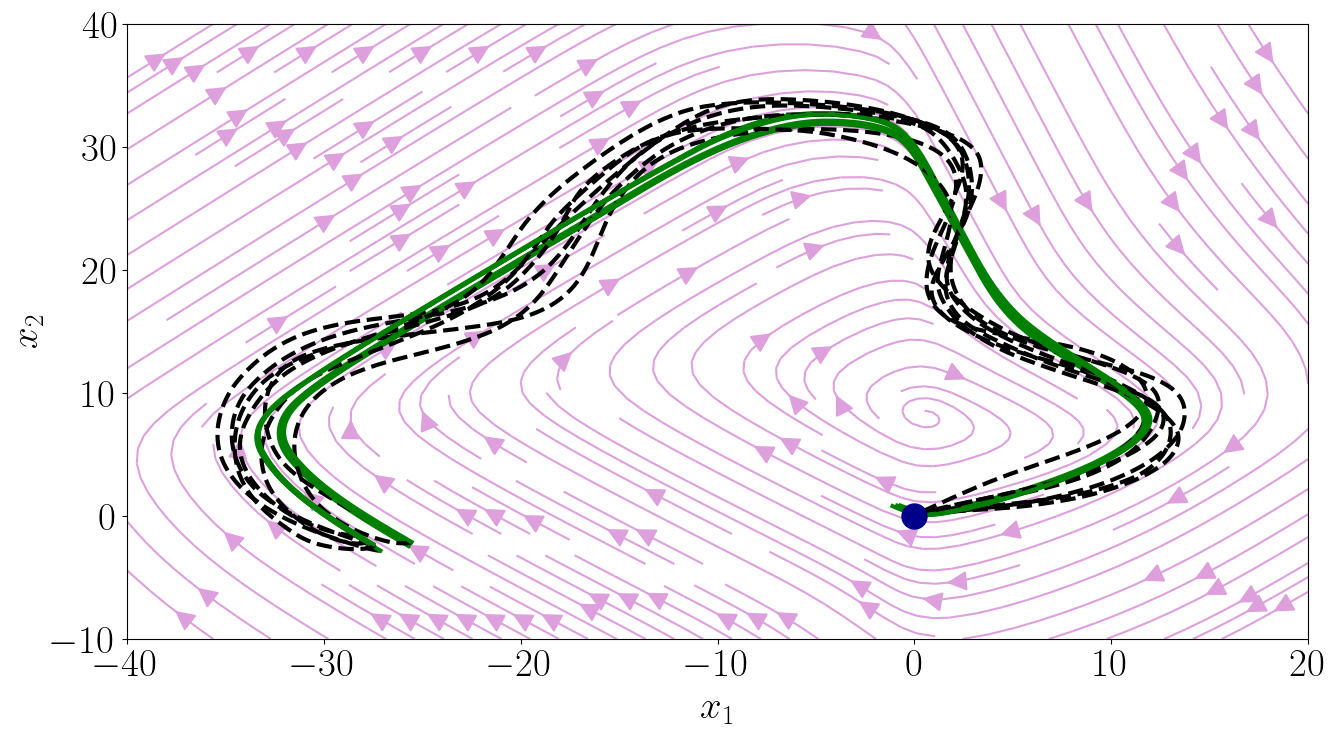}
         \caption{\textit{Leaf-2}}
         \label{fig:Leaf_2}
     \end{subfigure}
     \hfill
          \begin{subfigure}[b]{0.24\textwidth}
         \centering         \includegraphics[width=\textwidth]{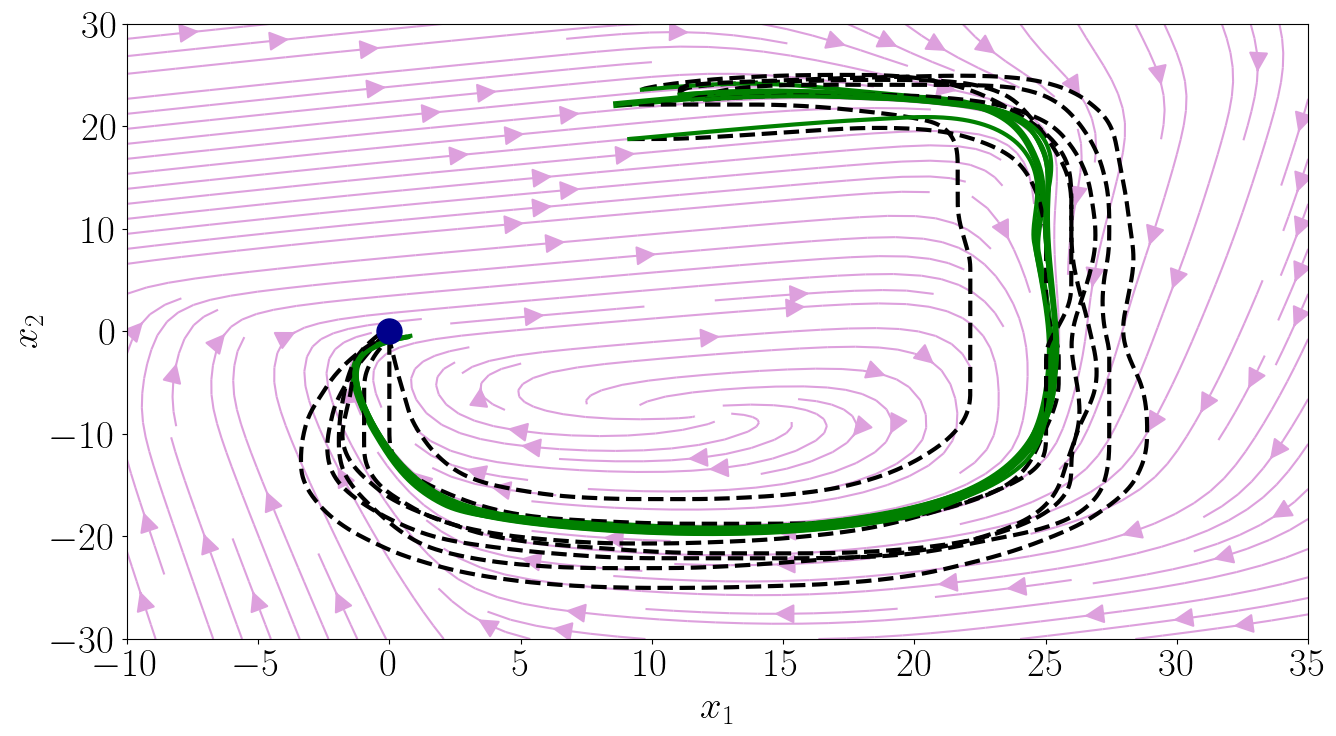}
         \caption{\textit{JShape-2}}
         \label{fig:JShape_2}
     \end{subfigure}
     \hfill
     \begin{subfigure}[b]{0.24\textwidth}         \centering        \includegraphics[width=\textwidth]{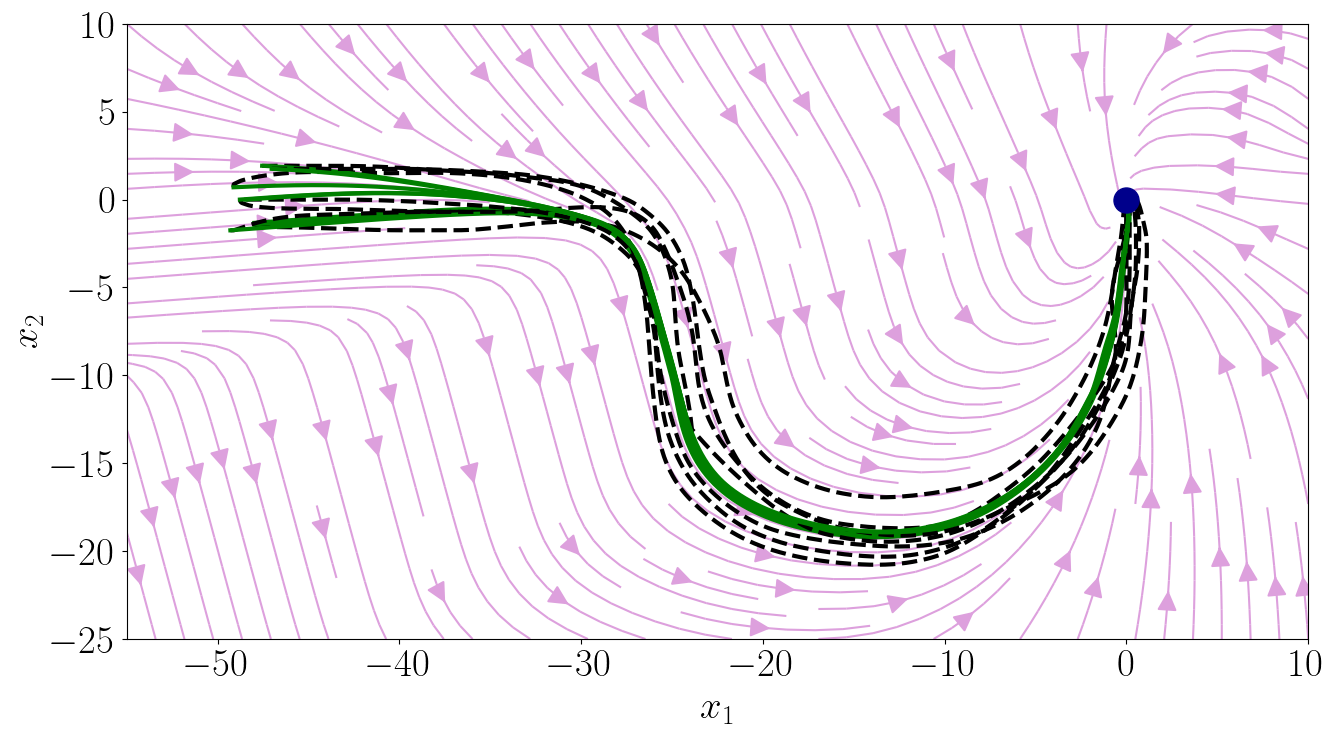}
         \caption{\textit{Spoon}}
         \label{fig:Spoon}
     \end{subfigure}
     \\
         \hfill
    \begin{subfigure}[b]{0.24\textwidth}
         \centering         \includegraphics[width=\textwidth]{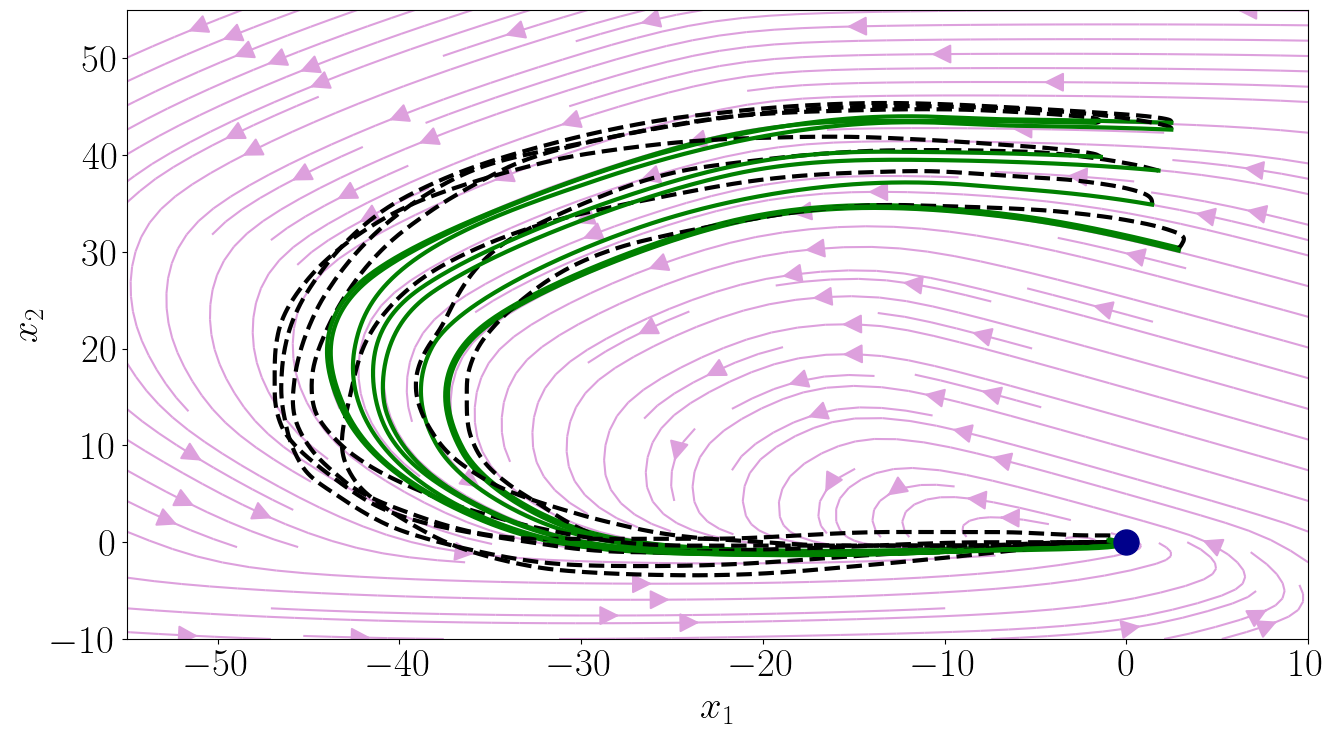}
         \caption{\textit{CShape}}
         \label{fig:CShape}
     \end{subfigure}
         \hfill
    \begin{subfigure}[b]{0.24\textwidth}
         \centering         \includegraphics[width=\textwidth]{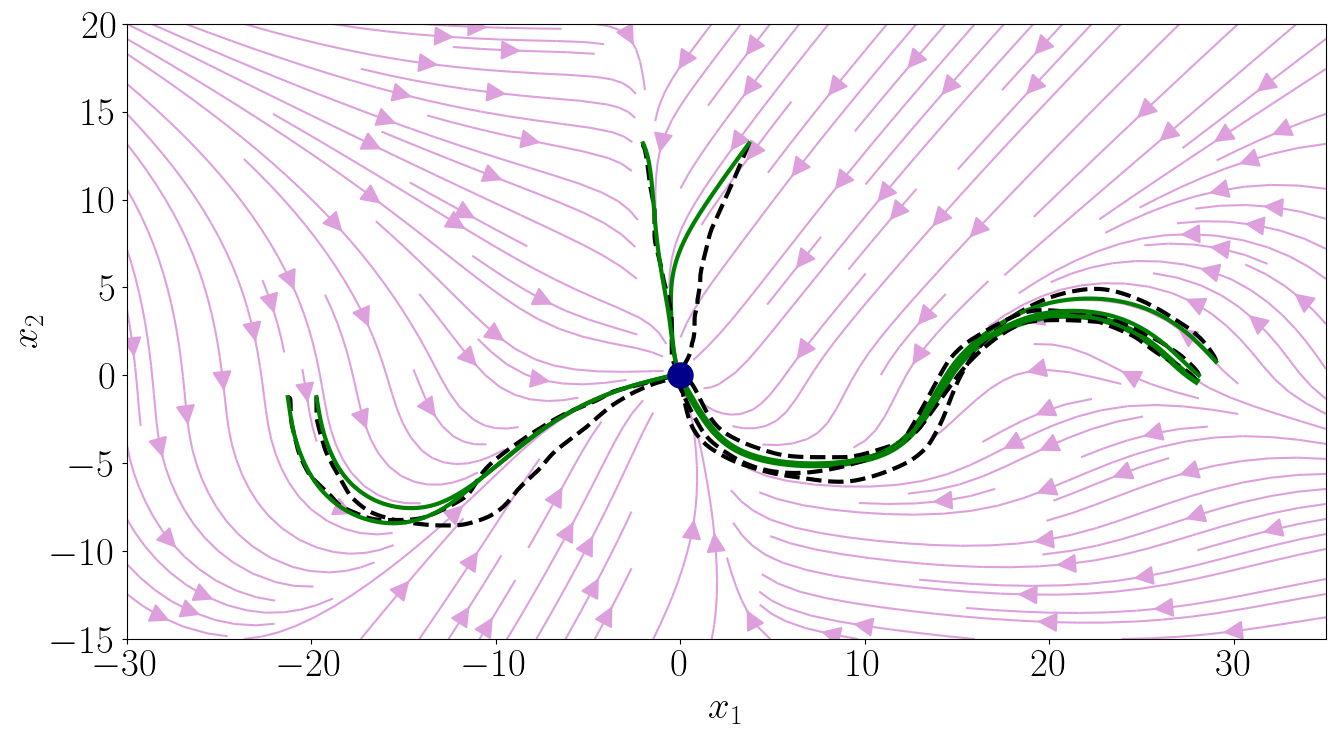}
         \caption{\textit{MultiModels1}}
         \label{fig:Multi_Models_1}
     \end{subfigure}
     \hfill
          \begin{subfigure}[b]{0.24\textwidth}
         \centering         \includegraphics[width=\textwidth]{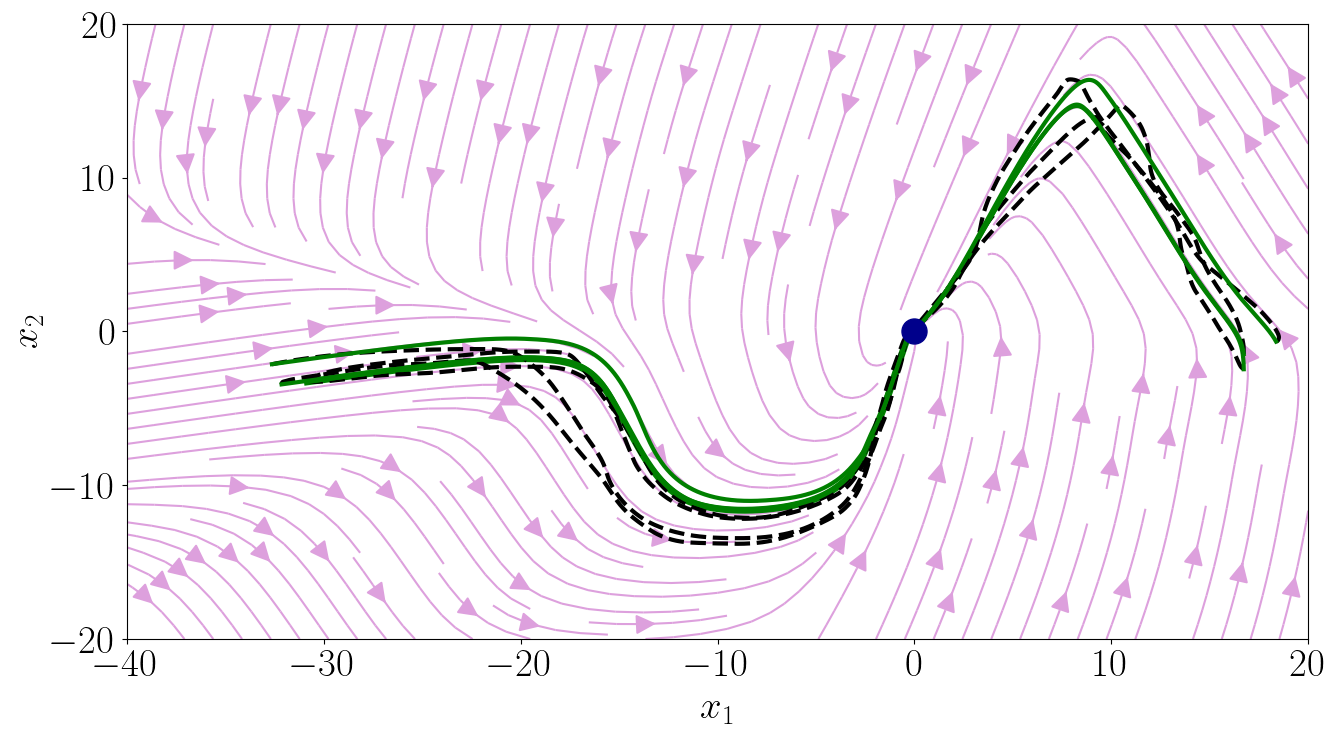}
         \caption{\textit{MultiModels2}}
         \label{fig:Multi_Models_2}
     \end{subfigure}
     \hfill
     \begin{subfigure}[b]{0.24\textwidth}         \centering        \includegraphics[width=\textwidth]{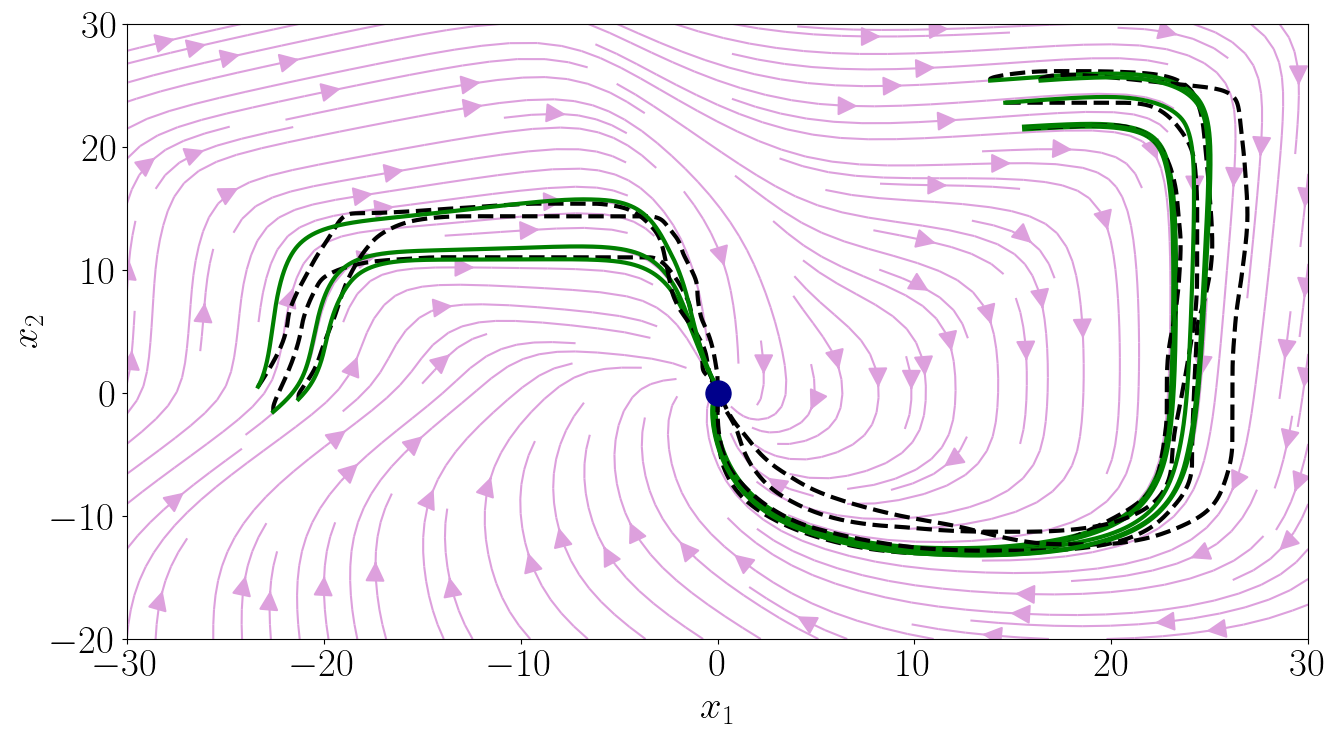}
         \caption{\textit{MultiModels4}}
         \label{fig:Multi_Models_4}
     \end{subfigure}
        \caption{Trajectories predicted by the NODE model without any safety or stability constraints on some non-linear and multi modal trajectories from the LASA handwriting data set. The dashed black trajectories are from the data set. The solid green trajectories are predicted by our model.}
        \label{fig:LASA_field}
\end{figure*}

\begin{figure*}[!h]
\begin{minipage}{0.48\textwidth}
          \begin{subfigure}[b]{0.49\linewidth}
         \centering         \includegraphics[width=\textwidth]{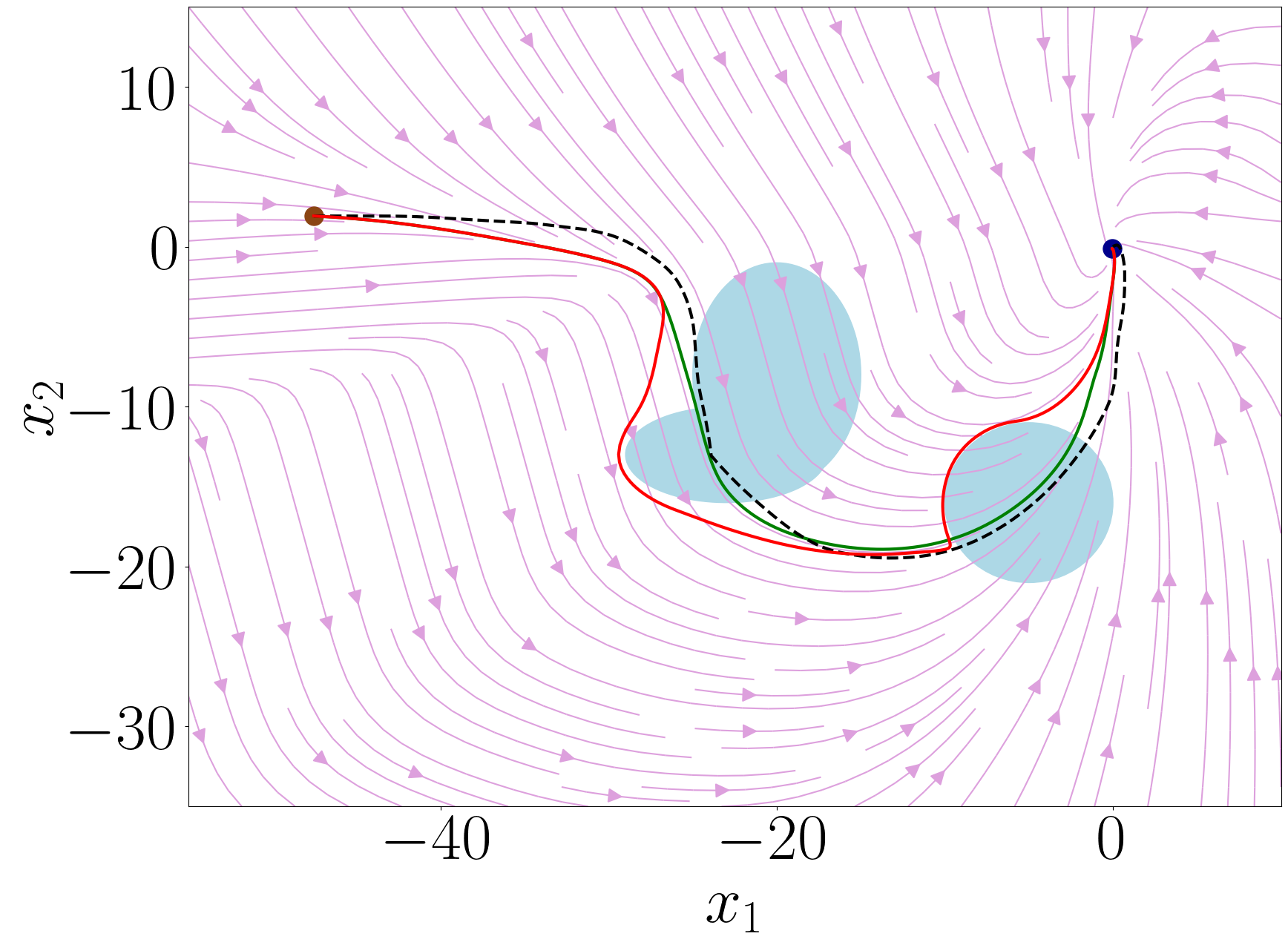}
         \caption{\textit{Spoon} with obstacles.} 
         \label{fig:concave_obst_1}
     \end{subfigure}
     \begin{subfigure}[b]{0.49\linewidth}
         \centering         \includegraphics[width=\textwidth]{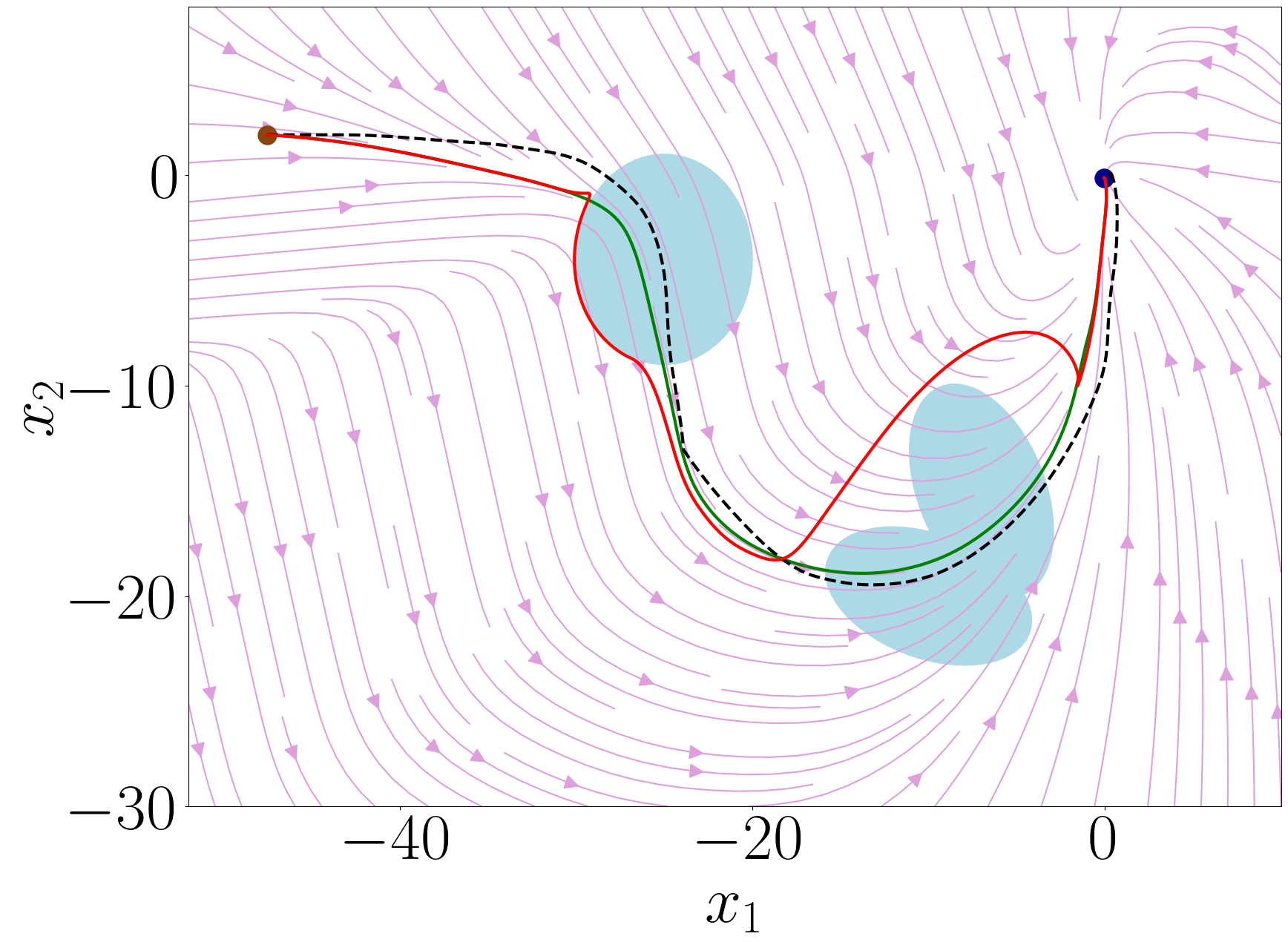}
         \caption{\textit{Spoon} with tilted obstacles}
         \label{fig:concave_obst_2}
     \end{subfigure}
        \caption{Stable and safe motion plan generated by our CBF-CLF NODE approach. Brown dot is the initial point, blue dot is the target point, and the light blue region are the obstacles.}
        \label{fig:LASA_dist_obst}
\end{minipage}\hfill
\begin{minipage}{0.48\textwidth}
     \begin{subfigure}[b]{0.49\linewidth}
         \centering         \includegraphics[width=\textwidth]{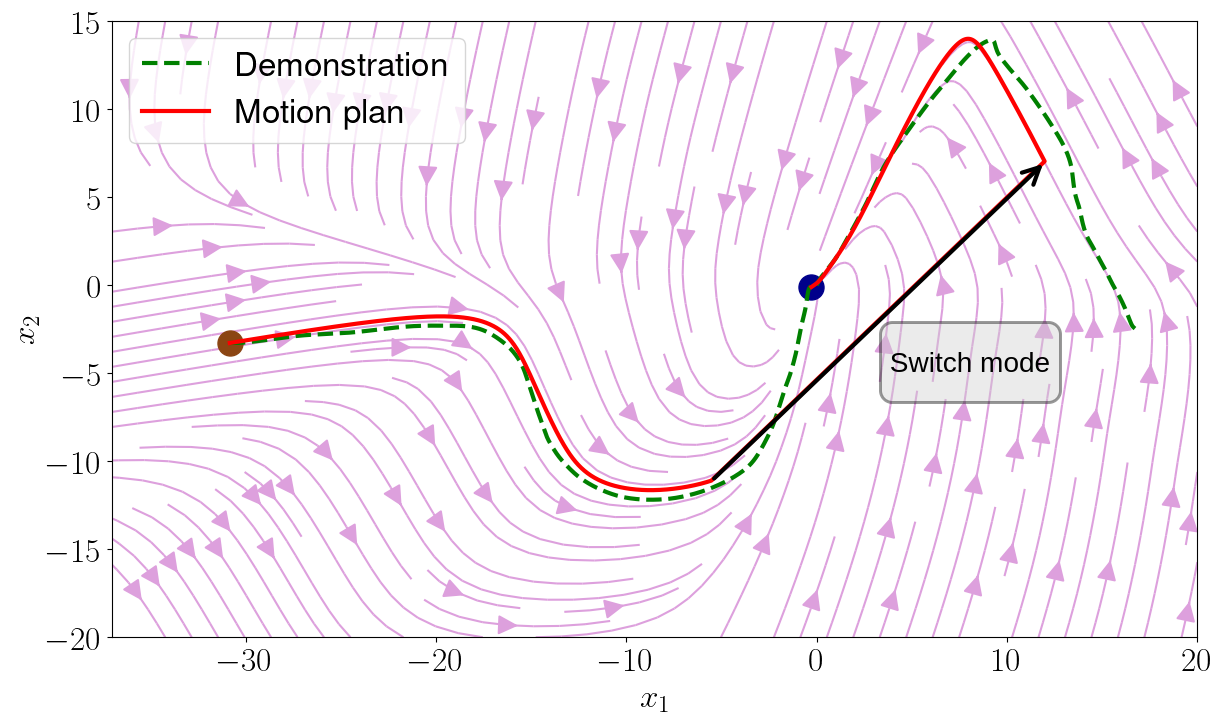}
         \caption{\textit{MultiModels2}}
         \label{fig:MultiModels2}
     \end{subfigure}
     \hfill
     \begin{subfigure}[b]{0.49\linewidth}         \centering        \includegraphics[width=\textwidth]{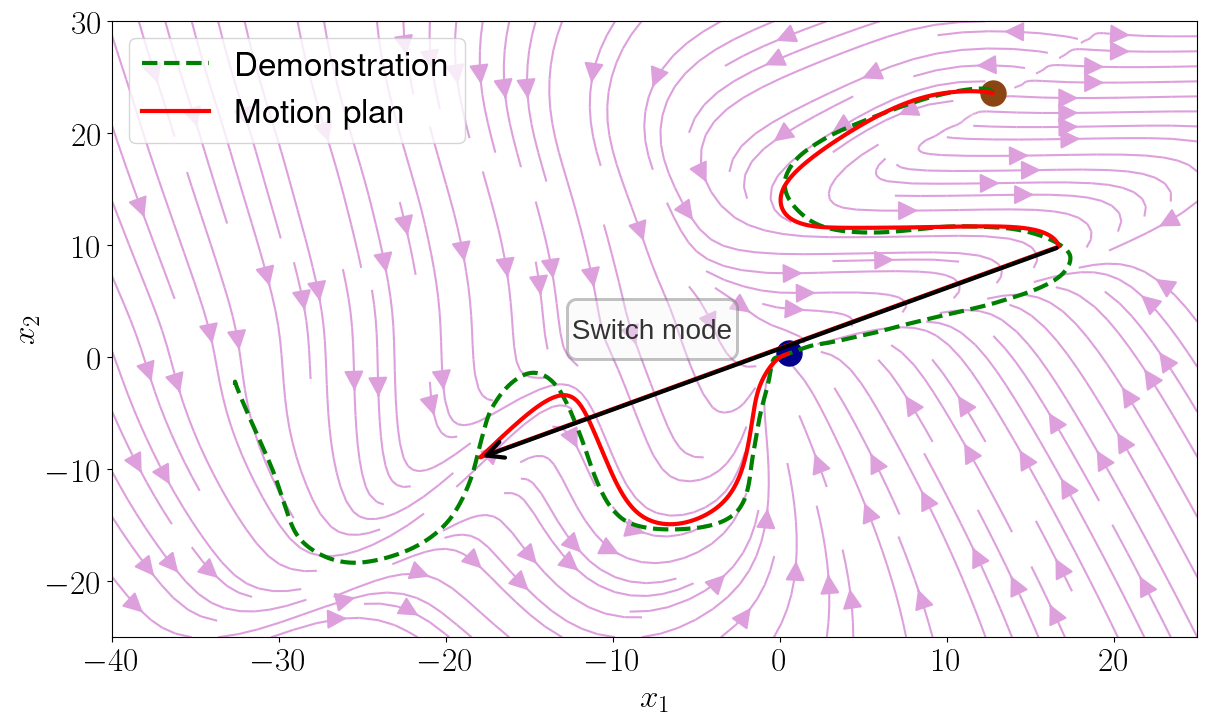}
         \caption{\textit{MultiModels3}}
         \label{fig:MultiModels3}
     \end{subfigure}
        \caption{Motion plan generated by our method when switching between modes. Brown dot is the initial point, and blue dot is the target point.}
        \label{fig:LASA_switch}
        \end{minipage}
\end{figure*}

In Fig.~\ref{fig:LASA_comparison}, we compare the performance of our NODE model on four highly non-linear LASA trajectories where existing approaches: SEDS~\cite{SEDS} and LPV-DS\cite{figueroa2022locally}; either directly go to the target point or do not capture the entire shape. We also present the trajectories predicted by our model on other non-linear and multi modal shapes in Fig.~\ref{fig:LASA_field}. We present obstacle avoidance scenarios in Fig.~\ref{fig:LASA_dist_obst}. We model one obstacle as a circle in Figs.~\ref{fig:concave_obst_1} and~\ref{fig:concave_obst_2}, where, the region outside the blue obstacle is the safe region~$\mathcal{C}$ as defined for the Control Barrier Function (CBF). The safe region is described using the CBF~$B(x) = \|x - c\|_2^2 - r^2$, where $c$ and $r$ is the center and radius of the obstacle, respectively. The other obstacle is concave and the barrier function of the concave obstacle is given by the union of two ellipses as below.
\begin{align}
B(x_1, x_2) = &\left(\frac{(x_1 - x_{c1,1})^2}{a_1^2} + \frac{(x_2 - x_{c1,2})^2}{b_1^2} - 1\right) \\ \nonumber
&\left(\frac{(x_1 - x_{c2,1})^2}{a_2^2} + \frac{(x_2 - x_{c2,2})^2}{b_2^2} - 1\right),
\end{align}
where $x = (x_1, x_2)$ is the Cartesian coordinates defined to be the state space. The parameters of the ellipses in Fig.~\ref{fig:concave_obst_1} are $(x_{c1, 1}, x_{c1, 2}) =(-20, -10), a_1 = 5, b_1= 7,  (x_{c2, 1}, x_{c2, 2}) = (-23, -13), 
a_2 = 6, b_2 = 3$; and the parameters of the ellipses in Fig.~\ref{fig:concave_obst_2} are $(x_{c1, 1}, x_{c1, 2}) =(-10, -22), a_1 = 3.5, b_1= 5.5,  (x_{c2, 1}, x_{c2, 2}) = (-8, -16), 
a_2 = 6, b_2 = 3$. The barrier function $B$ reshapes the vector field of the motion plan using the virtual control input $u(x)$ so that the robot motion always avoids the obstacle defined by the unsafe set: $\{x | B(x)  < 0\}.$ 

In Fig.~\ref{fig:LASA_switch}, we present two scenarios where the robot switches from one mode to another during task execution. We use a linear form for the class $\mathcal{K}$ functions in~\eqref{safety_opt}: $\alpha(x) = k_L x , \gamma(x) = k_B x $; where, ${k_L=4}$ and ${k_B=3}$ are appropriate numerical constants for the CLF and CBF constraint, respectively. We note that we did not have to re-tune $k_L$ and $k_B$ for different LASA shapes. If we increase the value of $k_L$, then the trajectory tracking behavior is more aggressive. Similarly, if we increase the value of $k_B$, the the obstacle avoidance behaviour is less conservative. The trade off for more aggressive maneuvers is the increase in the magnitude of the commanded desired velocity -- $f(x) + u(x)$ -- which might lead to actuation limits in real robot experiments. We used $\lambda = 0.5$ as the penalty term for the CLF-CBF QP. The forward looking horizon used is $N = 3$ for Algorithm~\ref{alg:target}. 

\begin{figure*}[!b]
     \centering
     \begin{subfigure}[b]{0.19\linewidth}
         \centering         \includegraphics[width=\textwidth]{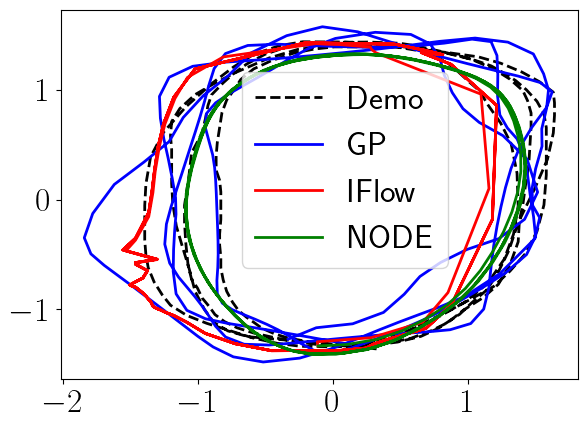}
         \caption{I shape}
         \label{fig:I_shape}
     \end{subfigure}
     \begin{subfigure}[b]{0.19\linewidth}         \centering        \includegraphics[width=\textwidth]{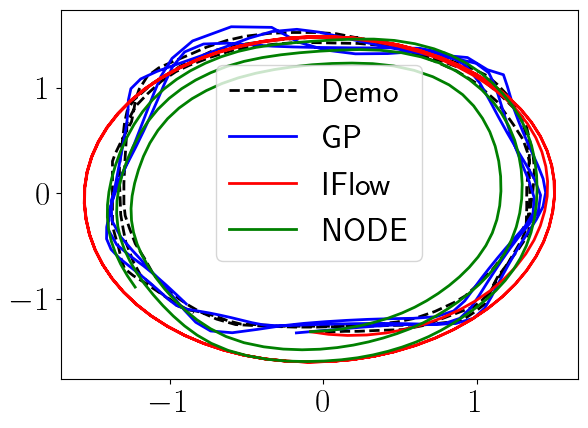}
         \caption{O shape}
         \label{fig:O_shape}
     \end{subfigure}
          \begin{subfigure}[b]{0.19\linewidth}         \centering        \includegraphics[width=\textwidth]{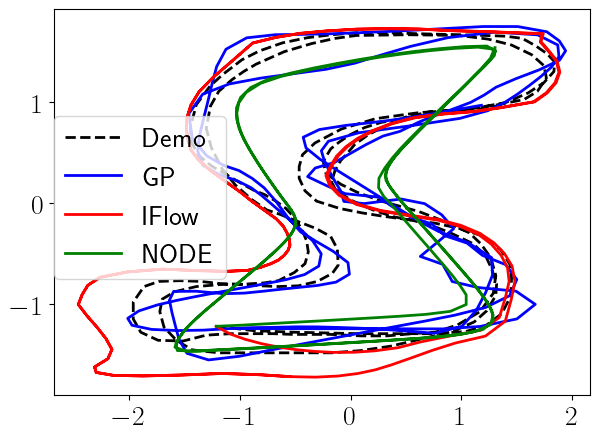}
         \caption{S shape}
         \label{fig:S_shape}
     \end{subfigure}
          \begin{subfigure}[b]{0.19\linewidth}         \centering        \includegraphics[width=\textwidth]{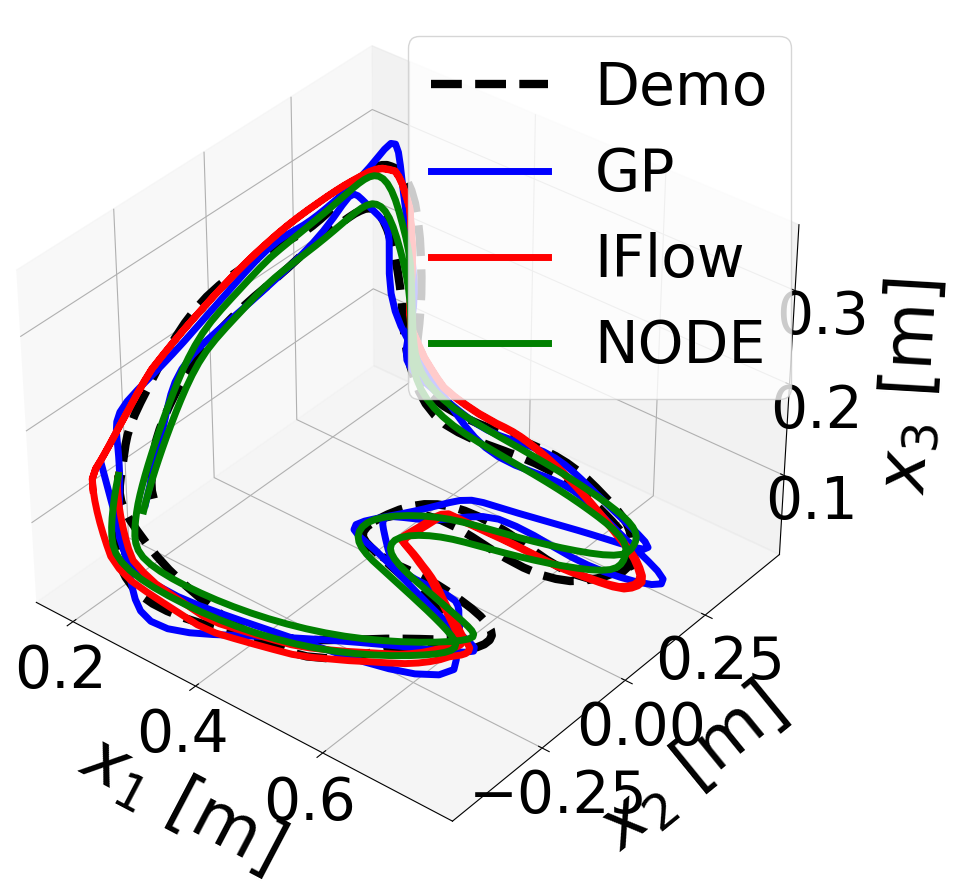}
         \caption{Wiping a whiteboard}
         \label{fig:wiping_whiteboard_app}
     \end{subfigure}
               \begin{subfigure}[b]{0.19\linewidth}         \centering        \includegraphics[width=\textwidth]{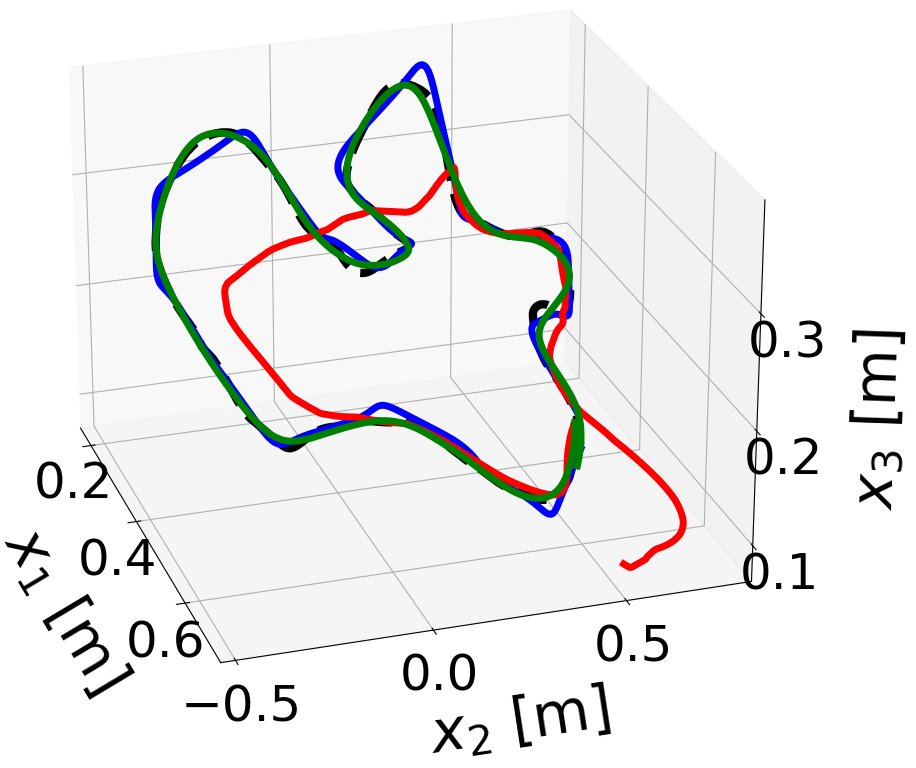}
         \caption{Wiping a mannequin}
         \label{fig:wiping_mannequin_app}
     \end{subfigure}
        \caption{Comparison of trajectory reproductions between IFlow, GP and our approach (NODE) on more periodic trajectories.}
        \label{fig:periodic_app}
\end{figure*}

In Fig.~\ref{fig:periodic_app}, we compare periodic motions predicted by our method (NODE) with Imitation Flow (IFlow)~\cite{imit_norm} and Gaussian Process~(GP)~\cite{LfD_GP} based approach. In addition to the \textbf{R} shape presented in Fig.~\ref{fig:RShape}, we also present trajectory predictions for the \textbf{I, O} and \textbf{S} shapes from~\cite{imit_norm} in Figs.~\ref{fig:I_shape}, \ref{fig:O_shape} and~\ref{fig:S_shape}. We compare the trajectory reproductions for 3D periodic wiping motions in Figs.~\ref{fig:wiping_whiteboard_app} and~\ref{fig:wiping_mannequin_app}, where we collect the training data by kinesthetically teaching the Franka robot the wiping tasks, similar to the spiral wiping motion in Fig.~\ref{fig:IFlow_Spiral}.

\end{document}